\PassOptionsToPackage{dvipsnames,table}{xcolor}
\documentclass[]{fairmeta}

\usepackage{hyperref}       
\usepackage{url}   
\usepackage{amssymb}
\usepackage{booktabs}       
\usepackage{nicefrac}       
\usepackage{microtype}      
\usepackage{xcolor}  
\usepackage{graphicx}
\usepackage{float}
\usepackage{makecell}
\usepackage{wrapfig}
\usepackage{subcaption}
\usepackage{amsmath}
\usepackage{cleveref}
\usepackage[justification=centering]{caption}
\newcommand{\ie}{\mbox{\it{i.e.,\ }}}
\newcommand{\eg}{\mbox{\it{e.g.,\ }}}
\usepackage{multirow}
\usepackage{pgfplots}
\pgfplotsset{compat=1.18}
\usepackage{tikz}
\usepgfplotslibrary{groupplots}
\usepackage{graphicx} 
\usepackage{tabularx}

\definecolor{primaryblue}{RGB}{31, 119, 180}
\definecolor{gridgray}{RGB}{235, 235, 235}
\definecolor{darkgray}{RGB}{50, 50, 50}

\pgfplotsset{
    academic style/.style={
        width=5.5cm,
        height=3.2cm,
        symbolic x coords={1, 2, 4, 8, 16, 32},
        xtick=data,
        tick label style={font=\tiny, color=darkgray},
        title style={font=\bfseries\tiny, yshift=-4pt},
        ylabel style={font=\bfseries\tiny, yshift=-4pt},
        y major grids=true,
        grid style={line width=0.4pt, color=gridgray},
        axis line style={color=darkgray, line width=0.5pt},
        tick style={color=darkgray, line width=0.4pt},
        enlarge x limits=0.08,
    },
    bottom x axis style/.style={
        x axis line style={primaryblue!80!black, line width=1.2pt},
        xticklabel style={
            font=\bfseries\tiny, 
            color=primaryblue!90!black,
            fill=gray!12,
            rounded corners=2pt,
            inner sep=1.5pt,
            yshift=-1pt
        },
        xlabel style={
            font=\bfseries\tiny,
            color=primaryblue!90!black,
            fill=primaryblue!12,
            draw=primaryblue!60,
            line width=0.5pt,
            rounded corners=3pt,
            inner sep=2.5pt,
            yshift=-2pt
        }
    }
}

\title{TTE-Flash: Accelerating Reasoning-based Multimodal Representations via Think-Then-Embed Tokens}

\author[]{Jianpeng Cheng}
\author[]{Xian Wu}
\author[]{Jiangfan Zhang}
\author[]{Wentao Bao}
\author[]{Chaitanya Ahuja}
\author[]{Shlok Kumar Mishra}
\author[]{Xuanming Cui}
\author[]{Hanchao Yu}
\author[]{Yang Gao}
\author[]{Fan Xia}
\author[]{Haixing Dai}
\author[]{Frankie Yuan}
\author[]{Zihao Wang}
\author[]{Xiaobing Chen}
\author[]{Qi Guo}
\author[]{Shaodan Zhai}
\author[]{Aashu Singh}
\author[]{Xiangjun Fan}
\author[]{Jun Xiao}
\affiliation[]{Meta-AI}

\abstract{
Recent research has demonstrated that Universal Multimodal Embedding (UME) benefits significantly from Chain-of-Thought (CoT) reasoning. In this paradigm, a generative model produces explicit reasoning traces for a multimodal query, with the final representation extracted from an <eos> embedding token attending to both the query and the reasoning. Despite its effectiveness, the computational overhead of generating explicit CoT traces is often prohibitive. In this work, we propose replacing explicit CoT with latent think tokens, which are interpreted as latent variables that can produce explicit CoT traces as observed variables. By optimizing think tokens using CoT generation loss and subsequent embedding tokens using contrastive loss, we produce high-performance, reasoning-aware representations at a constant inference cost. Our study investigates two key architectural designs: 1) how think and embeddings tokens should be extracted from the same LLM backbone. 2) how the tokens should be trained as two dependent tasks. We introduce TTE-Flash-2B, a reasoning-aware multimodal representation model that outperforms its explicit-CoT counterpart on the MMEB-v2 benchmark, while producing latent think tokens that are interpretable both textually and visually. Furthermore, zero-shot evaluation across 15 video datasets reveals scaling behavior as the number of think tokens increases and motivates study on adaptive think budget allocation based on task difficulty. 
}

\date{\today}

\begin{document}

\maketitle

\section{Introduction}

Universal Multimodal Embedding (UME) aims to learn instruction-following representations across text and visual modalities. Recent advancements have demonstrated that integrating Chain-of-Thought (CoT) reasoning before representation learning can significantly enhance representation quality, known as the Think-Then-Embed (TTE) framework \citep{cui2025think,lan2025ume}. Because multimodal content and instructions are often information-dense yet abstract, CoT serves to unearth latent information and guide the representation learning process, mirroring its success in complex reasoning and code generation tasks \citep{wei2022chain}. However, despite its efficacy, the practical deployment of TTE models is hindered by the prohibitive computational cost of generating explicit CoT traces in real-time.

The latency bottleneck intrinsic to CoT is not unique to UME; it has been explored extensively through latent reasoning as a means of achieving constant inference speeds in generation tasks \citep{hao2024training,zhu2025scaling,geiping2025scaling}. Within this paradigm, the LLM generates a fixed number of intermediate hidden states—representing continuous-space latent thoughts—between prompt encoding and target decoding. We introduce this approach to TTE, and formulating two  classes of continuous tokens emitted by the LLM backbone: \textbf{think tokens} and \textbf{embed tokens}. Specifically, the model processes multimodal inputs to generate a predefined number of think tokens prior to representation learning via embed tokens, all within a unified backbone. Under this framework, we investigate the following two research questions.

\textbf{Q1. How do we extract think and embed representations from a unified LLM backbone?} We seek a solution that balances high concurrency with representational fidelity. To this end, we evaluate two complementary paradigms: looped architectures~\citep{zhu2025scaling,hao2024training} and register-based mechanisms~\citep{wen2024efficient,darcet2023vision}. Looped architectures, widely adopted for latent reasoning, recursively re-inject the LLM’s continuous outputs as subsequent inputs. While this ensures representational depth, the process is inherently autoregressive and memory-bound. In contrast, registers frequently utilized in Vision Transformers~\citep{darcet2023vision}—serve a pooling role analogous to the <eos> token in UME, where all inputs are processed within a single pre-filling pass. Although this approach is compute-bound and significantly more efficient, registers lack context-dependence, potentially yielding less expressive representations. Our empirical results on MMEB~\citep{meng2025vlm2vec} indicate that for a 8-token model, registers double the throughput of looped methods, but they incur a 3\% relative degradation in retrieval accuracy. However, we show that introducing registers at every transformer layer narrows this gap while maintaining the forward efficiency.

\textbf{Q2. How should think and embed tokens be trained?} A primary challenge in latent reasoning is the effective supervision of intermediate hidden states. Prior research has explored curriculum learning to transition from explicit to implicit CoT \citep{deng2024explicit}, distillation of hidden states from explicit CoT-based teacher \citep{deng2023implicit}, and utilizing final task loss as indirect supervision \citep{yue2025hybrid}. In this work, we propose a simpler yet effective alternative training formalism: we treat think tokens as latent variables that generate explicit CoT as observed variables. As illustrated in Figure \ref{gen_loss}, we teach a pre-trained LLM to emit think tokens which then serve as an information bottleneck for decoding the explicit CoT. Out-of-domain evaluations on MMEB demonstrate that this approach enables think tokens to produce high-fidelity CoT traces that are well-aligned with multimodal instructions and ground truth.

After the reasoning phase, subsequent embed tokens are trained to extract multimodal representations via a standard multi-vector retrieval framework using contrastive learning \citep{santhanam2022colbertv2,faysse2024colpali}. We investigate the interplay between thinking and embedding objectives.  Preliminary experiments reveal that overlapping think and embed tokens degrades performance across both tasks. This indicates that while reasoning and representation learning are grounded to the same multimodal input, they require dedicated parameter specialization. Consequently, we isolate think and embed tokens into distinct sequence positions while using a shared backbone with decoupled register parameters, better aligning with the sequential "Think-Then-Embed" workflow.

The outcome of this research is TTE-Flash-2B, a reasoning-aware UME model designed for constant-time inference via ``Think-Then-Embed'' tokens. On the MMEB-v2 benchmark, TTE-Flash-2B outperforms the baseline reasoning-based UMEs that uses explicit CoT, while being 70x more efficient. We show that the latent think tokens admit both textual interpretation, via decoding into CoT, and visual interpretation, via an attached image generation head.
Furthermore, zero-shot evaluation across 15 video datasets reveals task-specific scaling behavior as the number of think tokens increases. This motivates further study into adaptive thinking budget allocation based on task difficulty.

\begin{figure}[!t]
    \centering
    \makebox[\textwidth][c]{
        \begin{subfigure}[b]{0.5\textwidth}
            \centering
            \includegraphics[width=\textwidth]{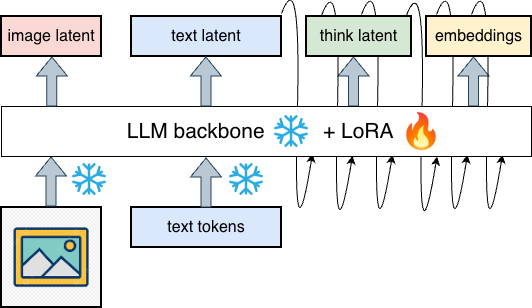}
            \caption{Looped architecture auto-regressively generates the next hidden state for reasoning and embedding.}
            \label{1a}
        \end{subfigure}
        \hspace{2em} 
        \begin{subfigure}[b]{0.5\textwidth}
            \centering
            \includegraphics[width=\textwidth]{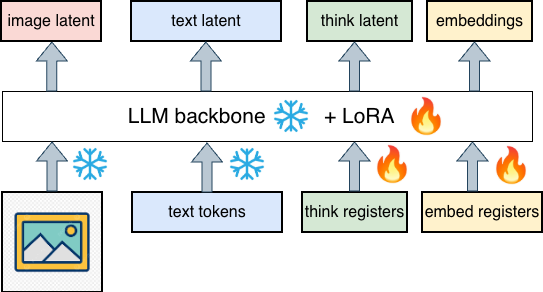}
            \caption{Registers are special tokens encoded together with the prompt as a single pre-filling pass.}
            \label{1b}
        \end{subfigure}
    }
\caption{A comparison of looped and register-based approaches for Think-Then-Embed tokens. \label{comparison}}
\vspace{-4mm}
\end{figure}

\section{Related Work}

\textbf{Universal Multimodal Embedding.}
Universal Multimodal Embedding (UME) leverages a unified MLLM backbone to map multimodal inputs into a shared space \citep{lin2024mm,gu2026unime,gu2025breaking}. Unlike CLIP-style dual-tower architectures  that rely on disjoint encoders and late-stage alignment \citep{radford2021learning,zhai2023sigmoid}, UME employs LLM-centric fusion to facilitate the generation of rich, reasoning-aware representations \citep{jiang2024e5,zhang2024gme,lin2024mm}.  To enhance instruction-following representations, reasoning-enhanced frameworks like TTE \citep{cui2025think}, UME-R1 \citep{lan2025ume} and MMEmb-R1 \citep{wang2026mmemb} integrate explicit CoT in UME before representation learning. Concurrent to our work, PLUME \citep{he2026plume} introduces latent reasoning in UME to achieve an accuracy-efficiency trade-off on MMEB-v2 \citep{meng2025vlm2vec}. Unlike PLUME’s looped architecture, TTE-Flash leverages parallel register tokens to enable efficient reasoning within a single prefill pass. We provide a direct experimental comparison between parallel register and sequential loop approaches.

\textbf{Latent Reasoning.}
Latent reasoning reframes CoT as recursive hidden-state computation. Instead of decoding reasoning steps to explicit tokens, the model generates intermediate hidden states as latent thought vectors. For example, Coconut \citep{hao2024training} introduced latent reasoning by recursively generating last hidden state and feeding it back to the model. CoLaR \citep{tan2025think} compresses multiple CoT tokens into one latent representation. LaDiR \citep{kang2025ladir} applies diffusion for latent reasoning. 
\citet{goyal2023think} introduces special pause tokens before generation.
To train latent reasoning states, common strategies include curriculum learning to migrate explicit CoT to latent reasoning \citep{hao2024training,deng2024explicit}, distillation from explicit CoT \citep{deng2023implicit,shen2025codi}, using end-to-end loss to indirectly supervise latent state or with reinforcement learning \citep{yue2025hybrid}. Different from these training strategies, we consider thought vectors as the latent information bottleneck that generates explicit CoT within the same pre-trained LLM backbone.

\textbf{Joint Contrastive-Generative Models.}
The integration of dual contrastive-generative objectives is common in foundation models for unified understanding and generation. Early efforts, such as CoCA \citep{yu2022coca}, jointly pre-trains image-text encoder-decoders using contrastive and captioning losses. Similarly, BLIP \citep{li2022blip} optimizes unimodal encoders and a text decoder through a combination of contrastive, captioning, and image-text matching objectives. More recently, InternVL \citep{chen2024internvl} has demonstrated the efficacy of conducting both vision-language contrastive and generative training within a shared LLM backbone. While our work also leverages dual objectives within a unified LLM, we focus on the  ``Think-Then-Embed'' task, utilizing dedicated token heads for  reasoning and  representation learning.

\section{TTE-Flash}
In this section, we detail the TTE-Flash architecture, focusing on how the think and embed tokens are interacted with a unified LLM, trained and used.

\subsection{Backbone}
As shown in Figure \ref{comparison}, we adopt a unified LLM backbone for multimodal content encoding, thinking and subsequent representation learning, considering everything as tokens. The LLM attention is fully causal in favor of simplicity: thinking steps attend to the raw multimodal input encoding; embedding steps attend to both multimodal input and thinking steps, same as the explicit TTE design \citep{cui2025think}. Within this paradigm, we consider two ways to represent the think and embed tokens, and extracting corresponding hidden states as representations.

\textbf{Loop}. As shown in the Figure \ref{1a}, looped architectures recursively feed the final prompt hidden state as the next input to the LLM backbone. The method requires $N+M$ auto-regressive steps (where $N$ is the total count of think tokens, and $M$ the embed tokens) to generate all representations, making it memory-bounded. 

\textbf{Register}. As shown in the Figure \ref{1b}, registers are learnable special tokens appended to the end of the input sequence. With a total of $N+M$ registers shared among different inputs, the architecture enables the model to extract all thinking and embedding states within a single pre-filling stage. The method is compute-bounded.

The two approaches share the same training recipe, and  differ only in how the $N+M$ tokens are extracted from the backbone.
In theory, using register tokens saves the latency of $N+M$ decoding steps compared to the looped architectures. In the looped approach, the GPU must repeatedly load the prompt KV cache $N+M$ times to generate each embedding sequentially. In the register approach, the GPU computes and loads all KV exactly once. Such dramatic efficiency gains are critical for scaling embedding applications like retrieval. In Section~\ref{arc_ablation}, we will compare loop and register-based approaches regarding performance and latency. Overall, TTE-Flash focuses on a register-based  design due to its superior efficiency. 


\begin{figure}[t!]
     \centering
     \includegraphics[width=0.8\textwidth]{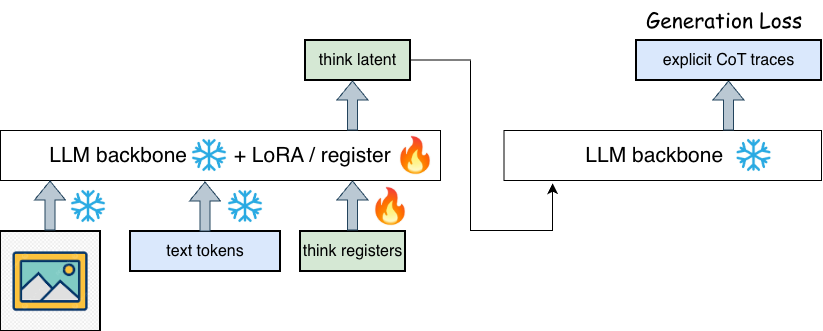}
     \caption{Think tokens incur explicit CoT generation loss.}
     \label{gen_loss}
\end{figure}

\begin{figure}[t!]
     \centering
         \includegraphics[width=\textwidth]{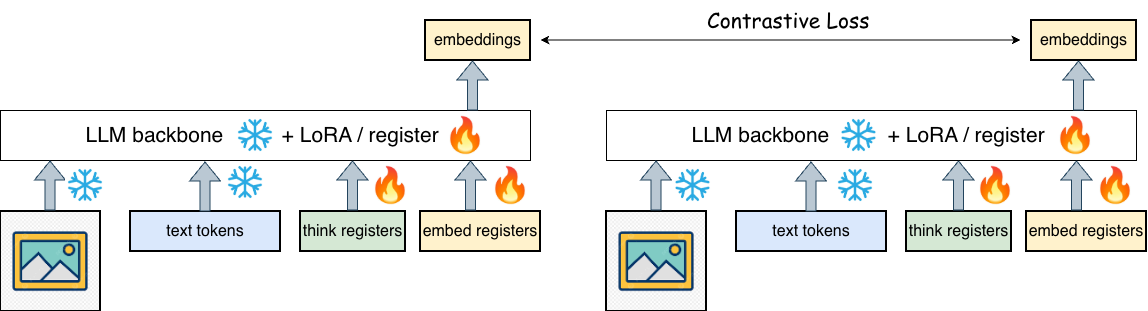}
     \caption{Embed tokens incur standard contrastive loss.}
     \label{cl_loss}
\end{figure}

\subsection{Training}
\textbf{Think Loss.}
 We define the following conditional generative process for CoT traces given multimodal input $x$: 
 \begin{align} 
 \mathbf{H_{thi}} &= f_\phi(x) \label{eq:encoder} \\ \text{CoT}_l &\sim p_\theta(\cdot \mid \text{CoT}_{<l},\, \mathbf{H_{thi}}) \label{eq:decoder} 
 \end{align} 
 Here $f_\phi$ denotes the LLM backbone with LoRA and think registers (or loops), producing $N$ thought vectors $\mathbf{H_{thi}} \in \mathbb{R}^{N \times d}$. The decoder $p_\theta$ is the same LLM backbone, kept frozen and without LoRA, preserving its pre-trained generation capability. Since $\mathbf{H_{thi}}$ is a deterministic function of $x$, the loss 
 \begin{equation}
 \mathcal{L}_{\text{gen}} = -\sum_{l=1}^{L} \log p_\theta(\text{CoT}_l \mid \text{CoT}_{<l}, \mathbf{H_{thi}}) \end{equation} 
 is the sole signal shaping the latent representation (Figure \ref{gen_loss}). The fixed think budget $N \ll L$ forces $\mathbf{H_{thi}}$ to act as an information bottleneck: it must compress the salient content of $x$ into $N$ vectors sufficient for constructing an $L$-token reasoning trace. Unlike variational approaches that regularize the latent space via a KL term against a fixed prior~\citep{kingma2013auto}, our bottleneck is implicitly regularized by the token budget $N$: with $N = 1$ to 32 vectors encoding information that would otherwise require an average of $L \approx 300$ discrete tokens in the eval dataset, the model must learn a maximally compressed yet generative representation.


\textbf{Embed Loss.}
As illustrated in the Figure \ref{cl_loss}, embed tokens (registers or loops) extract single or multiple embeddings conditioned on the original input and latent thought vectors. We impose standard contrastive loss upon each query embeddings $\mathbf{H_{emb}^q}$ and target embeddings $\mathbf{H_{emb}^t}$
\begin{equation}
    L_{\text{cl}} = -\log \frac{\text{exp}(\phi(\mathbf{H_{emb}^q}, \mathbf{H_{emb}^t}) / \tau )}{\sum_{t' \in \text{Batch}} \text{exp}( \phi( \mathbf{H_{emb}^q}, \mathbf{H_{emb}^{t'}}) / \tau)}
\end{equation}
where $\mathbf{H_{emb}^{t'}}$ denotes the in-batch negatives and $\tau$ the temperature. To compare embedding matrices $\mathbf{H_{emb}^q}$ and $\mathbf{H_{emb}^t}$, the scoring function $\phi$ is defined as the sum of pairwise vector similarities:
\begin{equation}
    \phi(\mathbf{H_{emb}^q} \mathbf{H_{emb}^t}) = \sum_{i=1}^{M} \text{sim}(\mathbf{H_{emb, i}^q}, \mathbf{H_{emb, i}^t})
\end{equation}
where $M$ represents the total number of embed tokens. When $M=1$, the scoring function reduces to the standard single-vector retrieval. We opt for a simple sum of pairwise similarities rather than the widely used ``sum of maximum'' \citep{santhanam2022colbertv2,faysse2024colpali} because the backbone LLM utilizes causal attention to generate dependent vectors. Consequently, pairwise similarity better preserves the positional correspondence between each register pair, which will be experimented in Section~\ref{arc_ablation}.

\subsection{Inference}
For embedding extraction, the fine-tuned backbone runs a single pre-filling pass with think and embed tokens. To visualize or interpret the think tokens, the think latents are passed through the pre-trained backbone for decoding.

\section{Experiments \label{exp}}
We conduct experiments using MMEB-V1 \citep{jiang2024vlm2vec} and MMEB-V2 \citep{meng2025vlm2vec} datasets. The contrastive training setup follows the same as TTE \citep{cui2025think}, following the per dataset weight used in VLM2Vec \citep{jiang2024vlm2vec}. To additionally supervise think tokens via CoT generation loss, we adopt the TTE CoT annotation \citep{cui2025think} and LLaVA-CoT dataset \citep{xu2025llava}. For comprehensive details on the training procedures, evaluation protocols, and the various tasks included in the dataset, please refer to the Appendix~\ref{sec:exp_details}. We first present ablation studies using cleaner MMEB-V1 (image only) setup to decide the best architecture and training recipe for TTE-flash. Then we report the full MMEB-V2 (image, video and visdoc) results and visualize the  think tokens.

\begin{figure*}[t]
    \centering
    \begin{minipage}{0.24\textwidth}
        \centering
        \includegraphics[width=\linewidth]{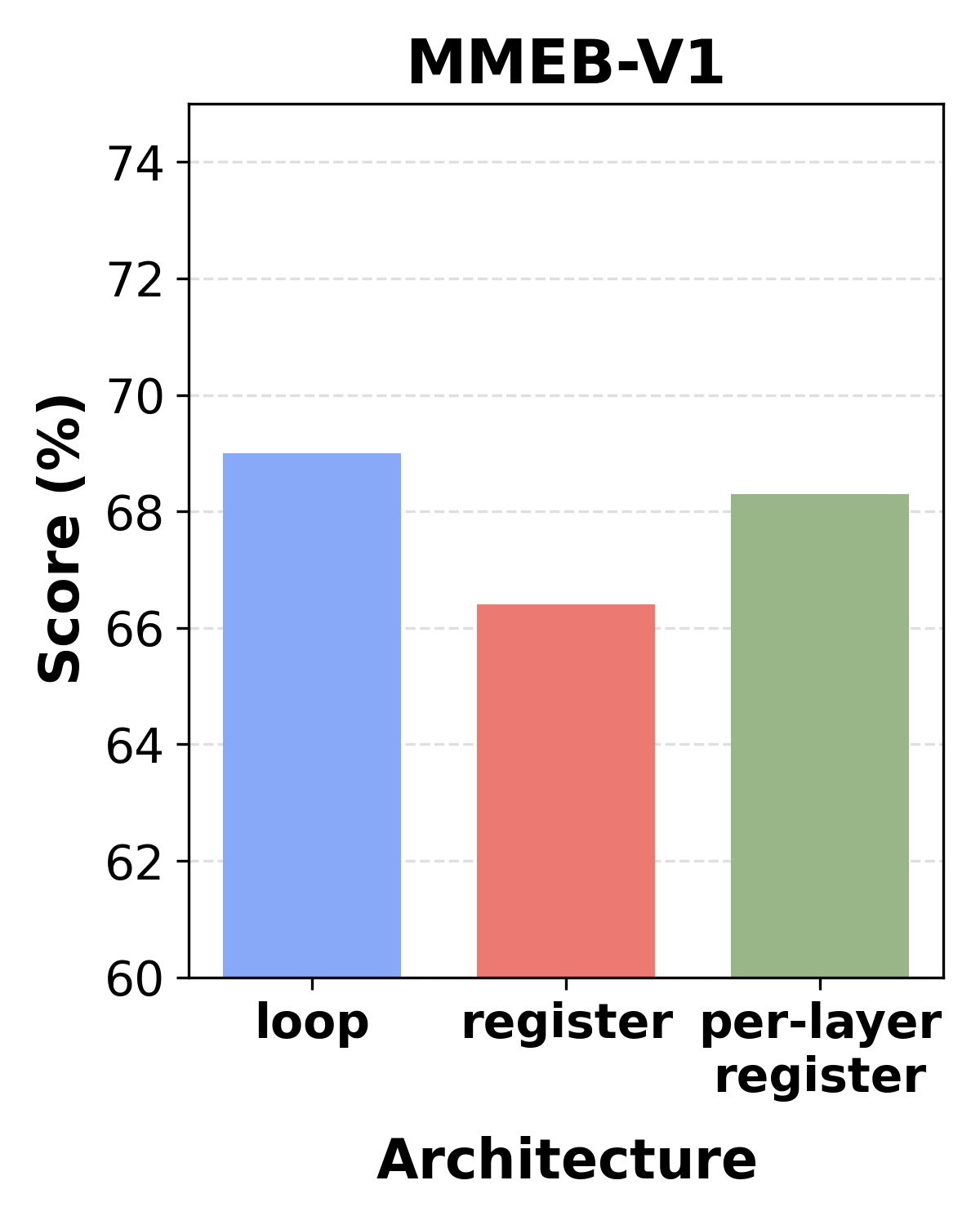}
        \captionsetup{font=small}
        \caption{Loop outperforms register. Per-layer register reduces the gap. \label{model_comp}}
    \end{minipage}
    \hfill
    \begin{minipage}{0.24\textwidth}
        \centering
        \includegraphics[width=\linewidth]{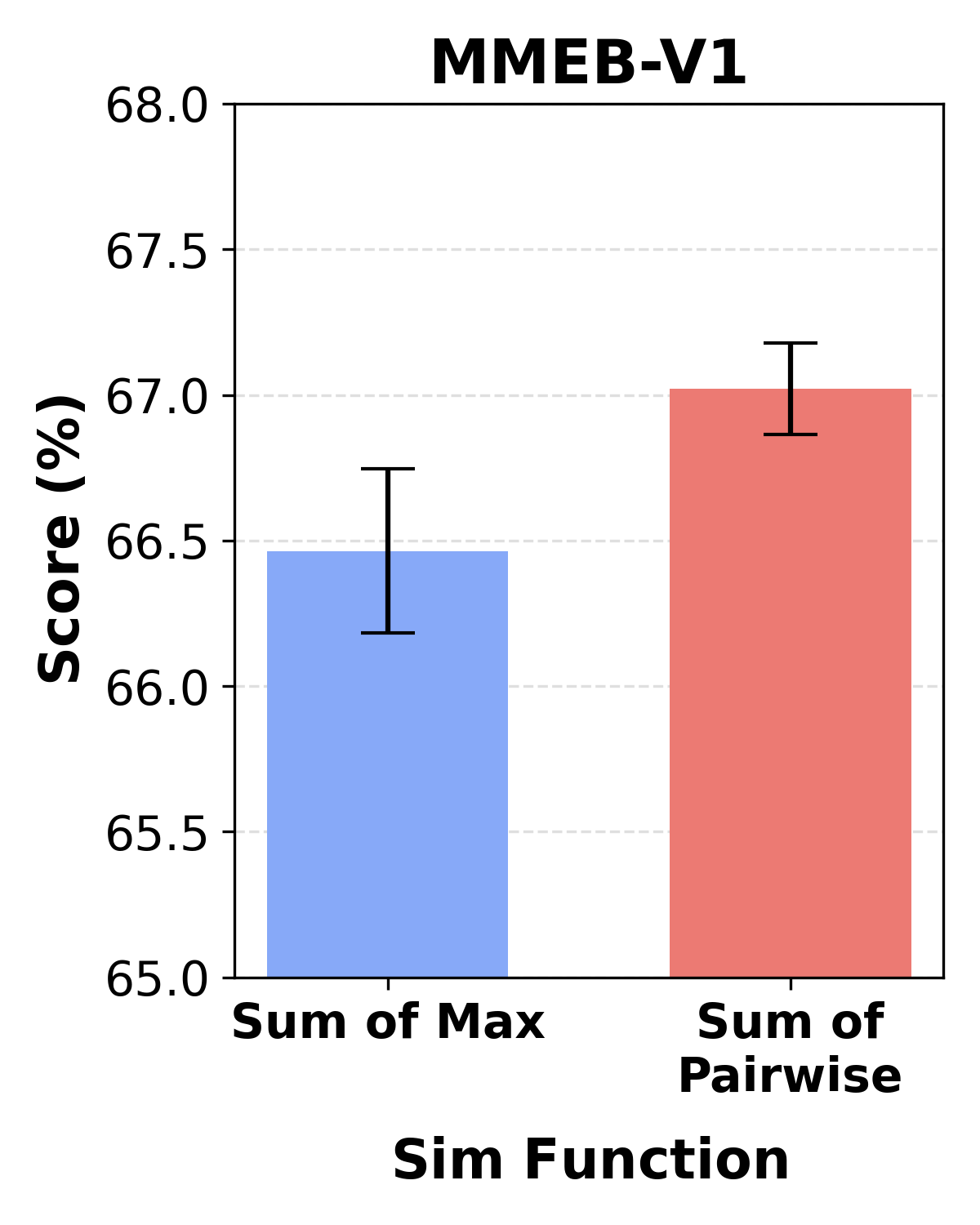}
        \captionsetup{font=small}
        \caption{Sum of pairwise similarity outperforms sum of max. \label{sim_comp}}
    \end{minipage}
    \hfill
    \begin{minipage}{0.48\textwidth}
        \centering
        \includegraphics[width=\linewidth]{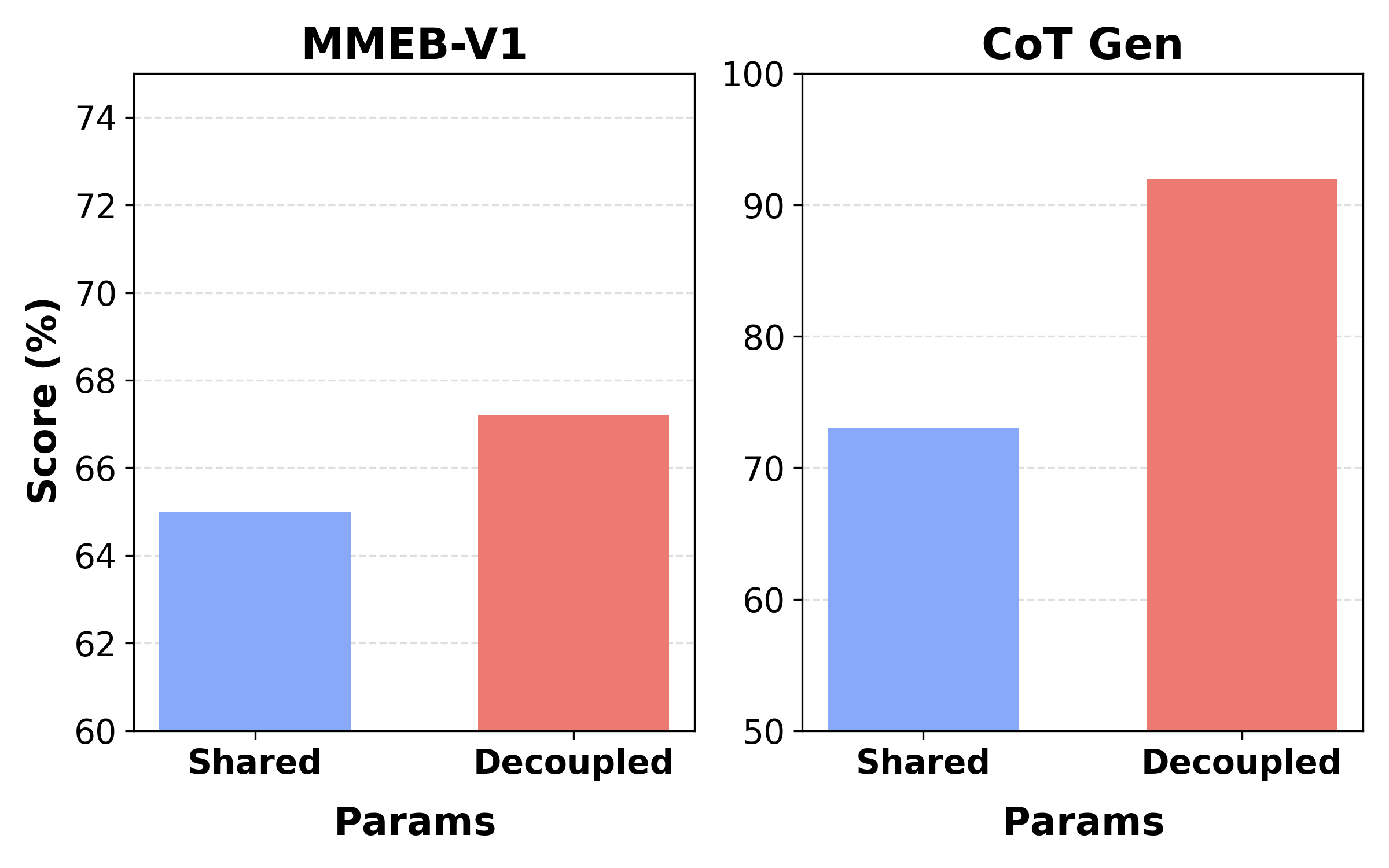}
        \captionsetup{font=small}
        \caption{Think and embed token decoupling outperforms sharing, measured by MMEB-V1 and CoT generation (similarity with ground truth). \label{param_comp}}
    \end{minipage}
\end{figure*}

\subsection{Architectural Design Ablations}
\label{arc_ablation}
\textbf{Comparison between Loops and Registers} Figure \ref{model_comp} compares register-based and loop-based approaches on MMEB-V1. The results are averaged across 20 checkpoints using total token budget 8. We observe that the loop-based approach consistently outperforms registers. Following PLUME \citep{he2026plume}, we also measure the inference efficiency of the two methods by sampling 500 examples per modality as one evaluation set, and repeat with 5 independently draws to report the average efficiency metrics. Table \ref{qps} shows the comparison where registers achieve a significant reduction of latency and increase of throughput, compared to loops. We observe that increasing the number of registers has a negligible impact on latency.

\begin{table}[ht]
    \centering
    \small
    \begin{tabular}{lcccccc}
        \toprule
        \textbf{Metric} & \makecell{TTE-flash\\4-loop} & \makecell{TTE-flash\\8-loop} & \makecell{TTE-flash\\4-registers} & \makecell{TTE-flash\\8-registers} &  \makecell{PLUME\\ \citep{he2026plume}} & \makecell{TTE$s$~\citep{cui2025think}\\explicit cot} \\
        \midrule
        Reasoning Tokens &4&8 &4 &8 &8 & 300 \\
        Latency(ms/sample) & 149& 179& 70&71 &298 & 5k \\
        Throughput(sample/s) & 6.7 & 5.6 & 14.2  & 14.1 &3.3  & 0.2\\
        \bottomrule
    \end{tabular}
    \vspace{4pt}
    \caption{Latency and throughput analysis of loop-based vs. register-based methods. \label{qps}}
\end{table}

Given restricted computational constraints, we focus on the register-based architecture. To enhance its performance, we introduce per-transformer-layer register tokens, added directly to each layer's hidden state. Since register tokens are position specific, the approach provides the model with additional expert parameters for think and embed tasks respectively. This modification changes the register token size from (N,) to (N, L) where L is the total transformer layers. In total, the added parameters is 55K for 1 token and up to 1.7M for 32 tokens. The performance gains of per-layer registers are illustrated in Figure \ref{model_comp}.

\textbf{Impact of Similarity Function}
To compute the similarity between query and target embedding matrices, we compare two similarity functions: 1) sum of maximum similarities between each query vector and the target set vectors \citep{santhanam2022colbertv2}, and 2) sum of all pairwise similarities between each query-target vector in the same position. We hypothesize that the pairwise approach is superior because our embeddings are position-aware and causally dependent; unlike sum of max, sum of pairwise explicitly preserves this position information. Figure \ref{sim_comp} compares the performance of both functions across MMEB-V1 using total token budget 8 over 20 runs. While the two functions achieve comparable performance, the pairwise approach exhibits a higher mean and lower variance.

\textbf{Impact of Think and Embed Token Decoupling} 
We evaluate an ablation study using 8 tokens where think and embed tokens are shared, forcing the model to learn a unified representation head for both reasoning and retrieval tasks. Figure \ref{param_comp} compares the performance of this shared configuration against the decoupled approach on both MMEB-v1 retrieval (embedding evaluation) and CoT generation (thinking evaluation). For CoT generation, we compare the average cosine similarity between the decoded CoT from think tokens and the ground truth CoT. The results demonstrate that token decoupling yields better performance for both thinking and embedding evaluations, suggesting that task-specific representations remain more effective than unified ones. 

\begin{figure*}[t]
\centering

\definecolor{brandblue}{RGB}{37, 99, 235}
\definecolor{branddark}{RGB}{30, 41, 59}
\definecolor{gridline}{RGB}{241, 245, 249}
\definecolor{axisline}{RGB}{203, 213, 225}

\pgfplotsset{
    prof plot style/.style={
        width=\linewidth,
        height=3.8cm,
        xtick=data,
        tick label style={font=\tiny\bfseries, color=branddark},
        title style={font=\bfseries\small, color=branddark, yshift=-3pt},
        xlabel style={font=\bfseries\small, color=branddark, yshift=2pt},
        ylabel style={font=\bfseries\small, color=branddark, yshift=-2pt},
        ymajorgrids=true,
        grid style={line width=0.5pt, color=gridline},
        axis line style={color=axisline, line width=0.6pt},
        tick style={color=axisline, line width=0.5pt},
        enlarge x limits=0.1,
        nodes near coords,
        every node near coord/.append style={
            font=\tiny\bfseries,
            color=branddark!90,
            yshift=2pt,
            /pgf/number format/fixed,
            /pgf/number format/precision=3,
            /pgf/number format/zerofill
        }
    },
    prof line style/.style={
        color=brandblue,
        line width=1.3pt,
        mark=*,
        mark size=2pt,
        mark options={fill=white, draw=brandblue, line width=1.2pt}
    }
}

\begin{minipage}[t]{0.32\textwidth}
\centering
\begin{tikzpicture}
\begin{axis}[
    prof plot style,
    title={MMEB-v1 (\#Emb = 1)},
    xlabel={\#Think tokens},
    ylabel={Score},
    symbolic x coords={0, 2, 4, 8, 16, 32},
    ymin=0.635, ymax=0.683,
    ytick={0.64, 0.65, 0.66, 0.67, 0.68},
    yticklabel style={/pgf/number format/fixed, /pgf/number format/precision=2, /pgf/number format/zerofill}
]
\addplot[prof line style] coordinates {
    (0, 0.643) 
    (2, 0.656) 
    (4, 0.664) 
    (8, 0.668) 
    (16, 0.670) 
    (32, 0.670)
};
\end{axis}
\end{tikzpicture}
\caption{Positive scaling pattern with the number of think tokens on MMEB-v1.}
\label{fig:scaling_think}
\end{minipage}%
\hfill
\begin{minipage}[t]{0.32\textwidth}
\centering
\begin{tikzpicture}
\begin{axis}[
    prof plot style,
    title={MMEB-v1 (\#Think = 0)},
    xlabel={\#Embed tokens},
    ylabel={Score},
    symbolic x coords={1, 2, 4, 8, 16, 32},
    ymin=0.632, ymax=0.672,
    ytick={0.63, 0.64, 0.65, 0.66, 0.67},
    yticklabel style={/pgf/number format/fixed, /pgf/number format/precision=2, /pgf/number format/zerofill}
]
\addplot[prof line style] coordinates {
    (1, 0.638) 
    (2, 0.642) 
    (4, 0.650) 
    (8, 0.657) 
    (16, 0.659) 
    (32, 0.661)
};
\end{axis}
\end{tikzpicture}
\caption{Positive scaling pattern with the number of embed tokens when \#think=0.}
\label{fig:scaling_embed_think0}
\end{minipage}%
\hfill
\begin{minipage}[t]{0.32\textwidth}
\centering
\begin{tikzpicture}
\begin{axis}[
    prof plot style,
    title={MMEB-v1 (\#Think = 8)},
    xlabel={\#Embed tokens},
    ylabel={Score},
    symbolic x coords={1, 2, 4, 8, 16, 32},
    ymin=0.660, ymax=0.675,
    ytick={0.660, 0.663, 0.666, 0.669, 0.672},
    yticklabel style={/pgf/number format/fixed, /pgf/number format/precision=3, /pgf/number format/zerofill}
]
\addplot[prof line style] coordinates {
    (1, 0.668) 
    (2, 0.665) 
    (4, 0.667) 
    (8, 0.668) 
    (16, 0.667) 
    (32, 0.668)
};
\end{axis}
\end{tikzpicture}
\caption{Neutral scaling pattern with the number of embed tokens when \#think=8.}
\label{fig:scaling_embed_think8}
\end{minipage}

\end{figure*}

\textbf{Does More Think Tokens Help?}
To identify the optimal training recipe, we conducted an ablation study on the number of think tokens using the MMEB-V1, while fixing the number of embedding tokens to 1. 
As shown in Figure \ref{fig:video_think_tokens}, we generally observe a positive correlation between performance and the number of think tokens. 
To further reveal the scaling benefits of think tokens, we conducted a think-intensive evaluation task by applying the image-trained checkpoint to MMEB-V2 video tasks in a zero-shot setting, hypothesizing that zero-shot tasks require more thinking. Interestingly we observed that the image-trained checkpoint can handle video tasks with the aid of TTE tokens. The figure highlights the video subsets that benefited most from this scaling. Notably, benchmarks requiring complex reasoning—such as ActivityNet-QA \citep{yu2019activitynet}, NextQA \citep{xiao2021next}, and VideoMME \citep{fu2025video}—exhibited more robust upward trends, supporting the hypothesis that temporal and logical reasoning demand a higher computational budget through extended think tokens. Additionally, the different scaling curve suggests that the optimal think budget is inherently task-dependent, which motivates our further experiment in Section \ref{adaptive}.

\textbf{Does More Embedding Tokens Help?} 
Next we ablate how the retrieval performance scale with an increase of embedding tokens. Since the embed task is dependent on the think task, we need to fix the number of of think tokens in this ablation study. We first studied \#think=0 in Figure \ref{fig:scaling_embed_think0} and then \#think=8 in Figure \ref{fig:scaling_embed_think8} on MMEB-V1. The result in Figure \ref{fig:scaling_embed_think0} reveals that the migration from single vector retrieval to multi-vector retrieval does provide incremental gains and the result also serves as the no-think baseline. 
However, as we apply \#think=8  in
Figure \ref{fig:scaling_embed_think8}, we observe diminishing  gains of adding more embedding tokens, in which case the performance of multi-vector retrieval is close to single-vector retrieval. We also confirmed this finding on ViDoRe retrieval tasks \citep{faysse2024colpali} where the inputs are longer. 

\begin{figure*}[t]
\centering
\begin{tikzpicture}
\begin{groupplot}[
    group style={
        group size=3 by 5,
        horizontal sep=0.6cm,
        vertical sep=0.6cm,
        xticklabels at=edge bottom,
        x descriptions at=edge bottom,
        ylabels at=edge left,
    },
    width=0.30\textwidth,
    height=2.3cm,
    tick label style={font=\tiny},
    label style={font=\scriptsize},
    ylabel style={font=\scriptsize, yshift=-0.1cm},
    title style={font=\scriptsize\bfseries, yshift=-1.2ex},
    xmode=log,
    log basis x={2},
    xtick={1, 2, 4, 8, 16, 32},
    xticklabels={1, 2, 4, 8, 16, 32},
    grid=major,
    grid style={line width=0.2pt, color=gray!25},
    axis line style={gray!50},
    tick style={gray!50}
]

\nextgroupplot[title={VATEX (T2V)}, ylabel={Score}]
\addplot[thick, primaryblue, mark=*, mark size=1pt] coordinates {(1, 41.3) (2, 46.8) (4, 50.4) (8, 53.6) (16, 44.3) (32, 48.6)};

\nextgroupplot[title={DiDeMo (T2V)}]
\addplot[thick, primaryblue, mark=*, mark size=1pt] coordinates {(1, 21.0) (2, 33.0) (4, 34.0) (8, 39.0) (16, 34.0) (32, 36.0)};

\nextgroupplot[title={MSVD (T2V)}]
\addplot[thick, primaryblue, mark=*, mark size=1pt] coordinates {(1, 66.0) (2, 73.0) (4, 74.0) (8, 78.0) (16, 75.0) (32, 78.0)};

\nextgroupplot[title={MSR-VTT (T2V)}, ylabel={Score}]
\addplot[thick, primaryblue, mark=*, mark size=1pt] coordinates {(1, 32.4) (2, 31.1) (4, 35.0) (8, 33.5) (16, 40.1) (32, 37.1)};

\nextgroupplot[title={Breakfast}]
\addplot[thick, primaryblue, mark=*, mark size=1pt] coordinates {(1, 31.0) (2, 25.0) (4, 42.0) (8, 40.0) (16, 42.0) (32, 41.0)};

\nextgroupplot[title={UCF-101}]
\addplot[thick, primaryblue, mark=*, mark size=1pt] coordinates {(1, 71.8) (2, 69.2) (4, 71.8) (8, 72.4) (16, 72.8) (32, 73.5)};

\nextgroupplot[title={Kinetics-700}, ylabel={Score}]
\addplot[thick, primaryblue, mark=*, mark size=1pt] coordinates {(1, 41.2) (2, 47.5) (4, 52.0) (8, 53.8) (16, 56.2) (32, 54.5)};

\nextgroupplot[title={SSv2}]
\addplot[thick, primaryblue, mark=*, mark size=1pt] coordinates {(1, 33.0) (2, 23.0) (4, 37.0) (8, 43.0) (16, 46.0) (32, 47.0)};

\nextgroupplot[title={EgoSchema}]
\addplot[thick, primaryblue, mark=*, mark size=1pt] coordinates {(1, 49.0) (2, 49.2) (4, 58.0) (8, 56.0) (16, 59.0) (32, 58.7)};

\nextgroupplot[title={ActivityNet-QA}, ylabel={Score}]
\addplot[thick, primaryblue, mark=*, mark size=1pt] coordinates {(1, 52.5) (2, 56.8) (4, 68.5) (8, 55.8) (16, 59.6) (32, 65.2)};

\nextgroupplot[title={NExT-QA}]
\addplot[thick, primaryblue, mark=*, mark size=1pt] coordinates {(1, 56.0) (2, 57.5) (4, 70.0) (8, 71.0) (16, 78.0) (32, 76.0)};

\nextgroupplot[title={Video-MME}]
\addplot[thick, primaryblue, mark=*, mark size=1pt] coordinates {(1, 48.0) (2, 51.0) (4, 58.0) (8, 69.0) (16, 66.0) (32, 71.0)};

\nextgroupplot[title={MVBench}, ylabel={Score}, xlabel={\# Think Tokens}]
\addplot[thick, primaryblue, mark=*, mark size=1pt] coordinates {(1, 47.6) (2, 45.2) (4, 49.8) (8, 47.0) (16, 49.2) (32, 47.8)};

\nextgroupplot[title={QVHighlights}, xlabel={\# Think Tokens}]
\addplot[thick, primaryblue, mark=*, mark size=1pt] coordinates {(1, 65.0) (2, 65.0) (4, 65.0) (8, 68.0) (16, 72.0) (32, 75.0)};

\nextgroupplot[title={Charades-STA}, xlabel={\# Think Tokens}]
\addplot[thick, primaryblue, mark=*, mark size=1pt] coordinates {(1, 27.2) (2, 27.8) (4, 27.8) (8, 29.8) (16, 29.4) (32, 29.8)};

\end{groupplot}
\end{tikzpicture}
\caption{Model performance across varying numbers of think tokens.}
\label{fig:video_think_tokens}
\end{figure*}

\subsection{Full Results}
Based on our ablation results, we adopt 32 think tokens and 1 embedding token for TTE-flash, based on per-layer registers. The trainable parameters include the LoRA adapters shared between think and embed, and the registers which are decoupled for the two tasks.
Table \ref{full} summarizes results on MMEB-V2, revealing three primary findings: 1) TTE-flash 2B outperforms TTE-V1 2B variants using explicit CoT; 2) TTE-flash 2B surpasses the 7B VLM2vec baseline, demonstrating that scaling reasoning steps offers an alternative than scaling parameters; and 3) TTE-flash exceeds latent-reasoning-based PLUME (2B), showing the advantage of our simplified training recipe for latent reasoning. Notably, TTE-flash is 70x more efficient than TTE-V1 by utilizing a single pre-filling stage (Table \ref{qps}).

\begin{table}[h]
\centering
\small
\setlength{\tabcolsep}{3pt} 
\resizebox{\textwidth}{!}{
\begin{tabular}{l *{15}{c} c}
\toprule
\multirow{2}{*}{\textbf{Model}} & \multicolumn{5}{c}{\textbf{Image}} & \multicolumn{5}{c}{\textbf{Video}} & \multicolumn{5}{c}{\textbf{VisDoc}} & \multirow{2}{*}{\textbf{All}} \\
\cmidrule(lr){2-6} \cmidrule(lr){7-11} \cmidrule(lr){12-16}
& \textbf{CLS} & \textbf{QA} & \textbf{RET} & \textbf{GD} & \textbf{All} & \textbf{CLS} & \textbf{QA} & \textbf{RET} & \textbf{MRET} & \textbf{All} & \textbf{VDR1} & \textbf{VDR2} & \textbf{VR} & \textbf{OOD} & \textbf{All} & \\
\midrule
\rowcolor{gray!15} \multicolumn{17}{c}{\textbf{2B Baselines}} \\
VLM2Vec-V2~\citep{meng2025vlm2vec} & 62.9 & 56.3 & 69.5 & 77.3 & 64.9 & 39.3 & 34.3 & 28.8 & 38.5 & 34.9 & 75.5 & 44.9 & 79.4 & 39.4 & 65.4 & 58.0 \\
ColPali~\citep{faysse2024colpali} & 40.3& 11.5& 48.1& 40.3& 34.9& 26.7& 37.8& 21.6& 25.5& 28.2 &83.6 &52.0& 81.1& 43.1& 71.0 &44.4\\
GME~\citep{zhang2024gme} & 54.4& 29.9& 66.9& 55.5& 51.9& 34.9& 42.0& 25.6& 32.4& 33.9 &86.1& 54.0& 82.5& 43.1 &72.7 &54.1\\
LamRA~\citep{liu2025lamra} & 59.2&  26.5&  70.0&  62.7&  54.1&  39.3&  42.6 & 24.3 & 34.6&  35.2&  22.0 & 11.5&  37.4 & 21.0&  23.9&  40.4\\
DUME~\citep{zhang2025bridging} & 59.3& 55.0 &66.3 &78.0 &62.5 &37.7 &46.6& 17.1 &30.0 &33.2& 67.6& 43.3& 47.1& 33.8& 52.8& 52.7\\

\rowcolor{gray!15} \multicolumn{17}{c}{\textbf{7B Models}} \\

VLM2Vec-V2~\citep{meng2025vlm2vec} & 65.7 & 61.5 & 70.0 & 85.2 & 68.1 & 45.9 & 33.9 & 27.6 & 39.3 & 36.4 & 78.8 & 52.6 & 82.7 & 42.1 & 69.3 & 61.2 \\

\rowcolor{gray!15} \multicolumn{17}{c}{\textbf{2B Models with Explicit-CoT}} \\
TTE$s$~\citep{cui2025think} & 67.9 & 66.6 & 70.2 & 84.1 & 70.1 & 47.3 & 49.1 & 34.4 & 33.2 & 32.1 & 77.5 & 53.2 & 83.2 & 41.1 & 68.8 & 63.1 \\
UME-R1~\citep{lan2025ume} & 64.8 & 62.8 & 67.6 & 77.2 & 66.6 & 44.3 & 51.2 & 32.9 & 39.7 & 42.2 & 72.4 & 46.2 & 79.2 & 37.2 & 63.9 & 60.1  \\

\rowcolor{gray!15} \multicolumn{17}{c}{\textbf{2B Models with Latent Reasoning}} \\
PLUME~\citep{he2026plume} & 66.5 & 59.2 & 67.6 & 79.7 & 66.3 & 45.0 & 52.3 & 33.5 & 46.7 & 44.1 & 72.1 & 49.8 & 78.1 & 57.4 & 67.5 & 61.6 \\
\rowcolor{gray!15} \textbf{TTE-Flash} & \textbf{67.1} & \textbf{61.7} & \textbf{70.7} & \textbf{81.3} & \textbf{68.3} & \textbf{54.4} & \textbf{51.5} & \textbf{45.4} & \textbf{51.5} & \textbf{50.6} & \textbf{75.6} & \textbf{52.7} & \textbf{83.8} & \textbf{41.1} & \textbf{68.1} & \textbf{64.1} \\

\bottomrule
\end{tabular}
}
\vspace{5pt}
\caption{Model performance on MMEB-V2 benchmark. \label{full}}
\end{table}

\vspace{-2mm}

\begin{figure}[t!]
     \centering
     \includegraphics[width=0.8\textwidth]{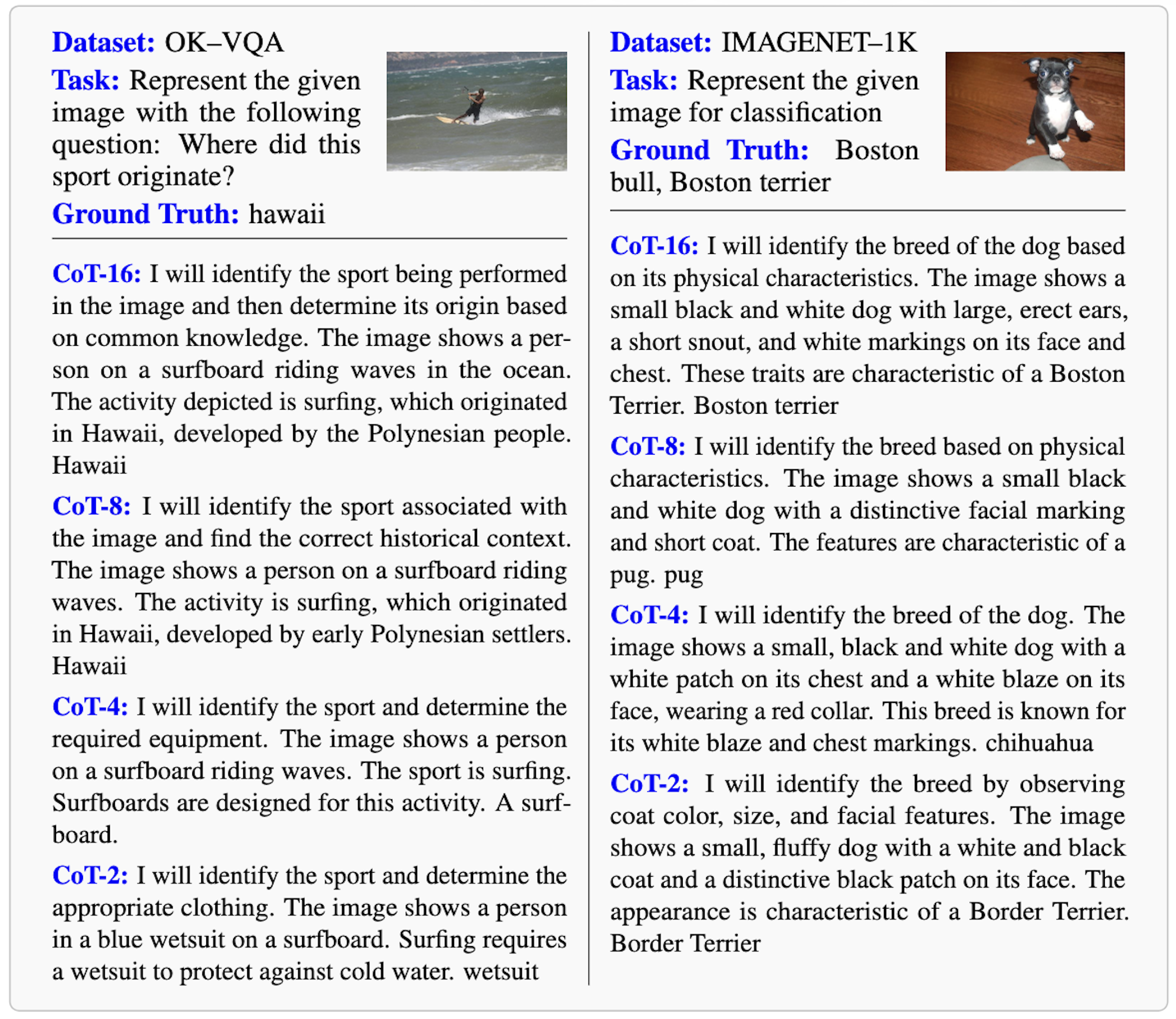}
     \caption{CoT traces decoded from model trained with different number of think tokens (2, 4, 8, 16).}
     \label{cot}
\end{figure}

\subsection{TTE-Flash Analysis}
\subsubsection{Test-time Analysis: Visualization of Think Tokens}
An advantage of TTE-Flash is that its think tokens are mappable to explict CoT for interpretation. 
We visualize how effectively the think tokens encode the explicit CoT in Figure \ref{cot}. The figure shows an example of the decoding output, comparing model trained with 2, 4, 8, and 16 think tokens. In general, increasing the number of think tokens yields higher-quality CoT traces. Additional examples of the decoded CoT across various MMEB tasks are provided in the Appendix \ref{more_examples}.

Beyond decoding think tokens into text, we investigate whether they encode sufficient visual semantics for image generation, following \citet{yu2024representation}'s finding that semantically rich representation encoders co-occur with strong generative capabilities.
Our text-to-image decoder is a flow-matching diffusion transformer conditioned on the think-token embeddings of TTE-Flash, trained to reconstruct DINOv2 \citep{oquab2023dinov2} latents of target images (training details in Appendix \ref{sec:visual-decoding-method}).
We train the decoder on the text-to-image (MSCOCO, VisualNews) and image+text-to-image (CIRR, NIGHTS) splits of MMEB. Qualitative results in Appendix \ref{image_vis} show that think tokens capture a semantically rich representation that visualizes the model's internal reasoning: the decoder yields caption-grounded images for text-to-image tasks and accurately composes reference images with textual modifications for composed retrieval.

\subsubsection{Test-time Analysis: Importance of Think Tokens}
\label{sec:latent_interventions}

To evaluate whether intermediate think states are essential at test time, we perform a post-hoc intervention analysis on the trained TTE-Flash model. We modify only the think-token latent states during embedding extraction across 4 test-time variants: the unmodified \textit{Baseline}, \textit{Masking} (zeroing think latent states at ratios of 25\%, 50\%, 75\%, and 100\% prior to embedding attention), \textit{Shuffling} (permuting input think registers across test instances), and \textit{Randomization} (injecting additive Gaussian noise into all think latent states).

As shown in Table~\ref{tab:think_interventions}, disrupting learned think states causes a steep drop in retrieval performance. Permuting think registers or adding Gaussian noise severely degrades model functionality. Under masking, performance decays monotonically with the masking ratio, though complete zeroing retains marginally higher capability than shuffling or noise. These results indicate that intermediate latent states are functionally essential for constructing the final embedding.

\begin{table}[h!]
\centering
\small
\begin{tabular}{lc}
\toprule
\textbf{Variant} & \textbf{MMEB-V1 Score} \\
\midrule
Original (Baseline) & \textbf{68.30} \\
Mask 25\% Think State & 21.27 \\
Mask 50\% Think State & 12.60 \\
Mask 75\% Think State & 11.46 \\
Mask 100\% Think State & 11.35 \\
Randomize Think State & 4.70 \\
Shuffle Think Registers & 2.19 \\
\bottomrule
\end{tabular}
\caption{MMEB-V1 score under test-time think-state disruptions.}
\label{tab:think_interventions}
\end{table}

\subsubsection{Training Analysis: Impact of CoT Form}
\label{sec:supervision_ablation}
CoT reasoning exhibits high entropy and inherently lacks a unique ground-truth trajectory. To study sensitivity of TTE-Flash training to the CoT form, we vary the think-token generation target while keeping model architecture and losses fixed. We evaluate 4 variants: \textbf{Full CoT} (original baseline trace), \textbf{Shortened CoT} (concise trace summarized from the query), \textbf{Noisy CoT} (randomly swapped reasoning sentences across training samples), and \textbf{Caption as CoT} (task-specific caption supervision for image-to-text retrieval).

\begin{table}[h!]
\centering
\small
\begin{tabular}{lc}
\toprule
\textbf{Variant} & \textbf{MMEB-v1 Score} \\
\midrule
Full CoT & \textbf{68.3} \\
Shortened CoT & 67.9 \\
Noisy CoT & 59.0 \\
\bottomrule
\end{tabular}
\caption{Performance across CoT variants on MMEB-v1.}
\label{tab:cot_variants}
\end{table}

As shown in Table~\ref{tab:cot_variants}, Full CoT achieves the top score, but Shortened CoT performs comparably well (67.9 vs.\ 68.3), demonstrating that long verbatim reasoning traces are not required for training. Noisy CoT degrades performance significantly (59.0), with the largest drops occurring on abstract retrieval tasks like VQA where mismatched reasoning introduces incorrect answers. This indicates that semantic correctness is critical, while exact linguistic structure is secondary.

Table~\ref{tab:caption_cot} benchmarks caption-style supervision on MSCOCO I2T retrieval, which yields results near Full CoT (70.4 vs.\ 70.8) as captions act as condensed reasoning. However, caption-style targets cannot be universally applied across MMEB, as many retrieval targets are multimodal (e.g., CIRR, MomentSeeker). 

\begin{table}[h!]
\centering
\small
\begin{tabular}{lc}
\toprule
\textbf{Variant} & \textbf{MSCOCO I2T Score} \\
\midrule
Full CoT & \textbf{70.8} \\
Caption as CoT & 70.4 \\
\bottomrule
\end{tabular}
\caption{Caption-style supervision on MSCOCO Image-to-Text retrieval.}
\label{tab:caption_cot}
\end{table}

We conclude that think tokens effectively capture task-relevant information from CoT supervision. The learning mechanism relies primarily on the semantic quality of the supervision signal rather than its length or surface phrasing.

\subsubsection{Training Analysis: Adaptive Think Budget Allocation \label{adaptive}}
In Figure \ref{fig:video_think_tokens} we observe that different tasks require varying computational overhead for think tokens. Therefore we study an adaptive approach that allows the TTE-flash model to dynamically allocate thinking budget based on the multimodal input. The core mechanism is to train the model to predict a first-K activation mask for think tokens.
We frame budget allocation as a categorical sampling task, where the $i$-th class represents the decision to activate the first $i$ tokens. The budget logits are generated by a two-layer MLP conditioned on the multimodal input. To maintain end-to-end differentiability during training, we employ the Gumbel-Softmax trick \citep{jang2016categorical} with hard sampling. After sampling, only first-K tokens participate in the forward pass. In addition to standard thinking and embedding losses, we incorporate a loss term to penalize K, encouraging less token activations. Technical implementation details are provided in the Appendix \ref{sec:adaptive-think-budget}.

\begin{table}[h!]
\centering
\caption{Mean think token activations per sample across evaluation datasets.}
\label{tab:think_tokens_separated}
\small
\begin{tabular}{llc}
\toprule
\textbf{Category} & \textbf{Dataset} & \textbf{Mean Think Tokens} \\
\midrule
Visual Question Answering & GQA & 13.7 \\
                          & OK-VQA & 13.0 \\
                          & TextVQA & 12.7 \\
\midrule
Retrieval-Oriented        & MSCOCO & 7.1 \\
                          & RefCOCO & 7.0 \\
                          & Wiki-SS-NQ & 6.0 \\
\midrule
Cross-Modal Retrieval     & VisualNews (I2T) & 17.8 \\
    & VisualNews (T2I) & 8.6 \\
                          & MSCOCO (I2T) & 15.5 \\
                          & MSCOCO (T2I) & 11.2 \\
                          
\midrule
Temporal Video Reasoning  & SSv2 & 16.5 \\
                          & MSVD (T2V) & 15.5 \\
                          & EgoSchemaQA & 14.1 \\
\midrule
Video Action Recognition  & UCF101 & 7.6 \\
                          & Breakfast & 7.0 \\
\bottomrule
\end{tabular}
\end{table}

As shown in Table~\ref{tab:think_tokens_separated}, the model dynamically scales compute with task complexity: complex VQA and temporal video reasoning tasks receive higher average token budgets than  retrieval or action recognition. This adaptivity extends to cross-modal tasks, where asymmetric image-to-text retrieval demands significantly more thinking tokens than text-to-image generation. Although adaptive thinking operates at an efficient 50\% token activation rate, capping compute at 32 tokens slightly trails a fixed 32-token baseline across image (66.0 vs. 68.3), video (49.4 vs. 50.6), and visdoc (65.5 vs. 68.1) results. This gap highlights key optimization challenges in learning adaptive allocation policies, which we leave for future exploration.

\section{Conclusion}
In this work, we introduce TTE-flash, an efficient ``Think-Then-Embed'' model utilizing latent think tokens mappable to explicit CoT traces. We favor a register-token-based architecture that generates all think and embed representations within a single, causal pre-filling pass.  TTE-flash outperforms both TTE-2B variants using explicit CoT and the larger VLM2vec 7B baseline. Zero-shot video evaluations across 15 subsets confirm the positive scaling patterns of think tokens. Finally, motivated by ablation studies that different tasks require varying thinking depths, we conduct a pilot study on teaching the model to adaptively select its think budget based on the input task requirements, marking the next direction of this work.

\bibliographystyle{assets/plainnat}
\bibliography{ref}


\appendix
\newpage
\section{Adaptive Think Budget Allocation}
\label{sec:adaptive-think-budget}

We introduce an adaptive think budget mechanism that learns how many think register tokens to allocate
for each input. Easy inputs (\eg short
classification queries) are ideally routed to a shorter reasoning prefix,
while harder inputs (\eg compositional VQA) are routed to use more think tokens.

\subsection{Budget Predictor}
Given input embeddings $\mathbf{X}$ and a maximum of $N$ think registers, we introduce a lightweight predictor that maps a pooled representation of the input to budget logits $\boldsymbol{\ell}_i$ over  $N$ candidate positions:
\begin{align}
\bar{\mathbf{x}}_i &=  \mathrm{Mean\_Pool}(\mathbf{X_i})
\\
\boldsymbol{\ell}_i &=  \mathbf{W}_2  * \mathrm{GELU}(\mathbf{W}_1\, \bar{\mathbf{x}}_i)
   \;
\end{align}

We sample a one-hot budget vector $\mathbf{z}_i \in \{0,1\}^{N}$ from
$\boldsymbol{\ell}_i$ using the straight-through Gumbel-Softmax
estimator with temperature $\tau$:
\begin{equation}
\mathbf{z}_i = \mathrm{GumbelSoftmax}(\boldsymbol{\ell}_i;\, \tau,\,
\text{hard}=\text{True}).
\end{equation}
The position of the single value $1$ in $\mathbf{z}_i$ encodes the chosen
budget. We convert it to a \emph{keep-mask}
$\mathbf{m}_i \in \{0,1\}^{N}$ via a reverse cumulative sum:
\begin{equation}
m_{i,k} \;=\; \sum_{j=k}^{N} z_{i,j},
\qquad k = 1,\dots,N.
\end{equation}
Because $\mathbf{z}_i$ is one-hot, $\mathbf{m}_i$ is a contiguous prefix of value $1$s of length
$m_i = \sum_k m_{i,k} = \arg\max_k z_{i,k}$.
The last value $1$ in $\mathbf{m}_i$ marks the embedding position; preceding
value $1$s mark the active think tokens. Gradients flow through
$\mathbf{m}_i$ via the Gumbel reparameterization, allowing
end-to-end training.
The mask is applied multiplicatively to the input embeddings.

\subsection{Budget Regularization}
To discourage trivial all-on solutions we add an $\ell_1$ penalty on
the average number of active think tokens:
\begin{equation}
\mathcal{L}_{\text{budget}} \;=\;
\lambda_{\text{b}} \cdot
\frac{1}{B} \sum_{i=1}^{B} m_i ,
\end{equation}
where $B$ is the total number of activated tokens and 
$\lambda_{\text{b}}$ is a small weight (0.01). The total
training objective combines the contrastive loss
$\mathcal{L}_{\text{con}}$, the generation loss
$\mathcal{L}_{\text{gen}}$, and the budget regularizer:
\begin{equation}
\mathcal{L} \;=\;
\mathcal{L}_{\text{con}}
\;+\; \mathcal{L}_{\text{gen}}
\;+\; \mathcal{L}_{\text{budget}}.
\end{equation}
The contrastive and generation losses naturally reward turning think
tokens on when they help; the regularizer pushes them
off when they do not, yielding an equilibrium in which budget
allocation adapts to per-sample difficulty.

\newpage
\section{Experiments Details}
\label{sec:exp_details}

\subsection{Datasets Details}

MMEBv2 \citep{meng2025vlm2vec} is a comprehensive evaluation benchmark for multimodal embedding models, comprising 9 meta-tasks and 78 individual tasks that span four modalities: text, image, video, and document. The benchmark is organized into the following meta-task categories:
\begin{enumerate}
    \item \textbf{Image Classification} (10 tasks: VOC2007, N24News, SUN397, ObjectNet, Country211, Place365, ImageNet-1K, HatefulMemes, ImageNet-A, ImageNet-R),
    \item \textbf{Visual Question Answering} (10 tasks: OK-VQA, A-OKVQA, DocVQA, InfoVQA, ChartQA, Visual7W, ScienceQA, GQA, TextVQA, VizWiz),
    \item \textbf{Image-level Retrieval} (12 tasks: MSCOCO\_I2T, MSCOCO\_T2I, VisDial, CIRR, VisualNews\_I2T, VisualNews\_T2I, NIGHTS, WebQA, EDIS, OVEN, Wiki-SS-NQ, FashionIQ),
    \item \textbf{Visual Grounding} (4 tasks: MSCOCO, RefCOCO, RefCOCO-Matching, Visual7W-Pointing),
    \item \textbf{Visual Document Retrieval} (24 tasks: ViDoRe (10), ViDoRe-V2 (4), VisRAG (6), ViDoSeek (2), MMLongBench-Doc (2)),
    \item \textbf{Video Classification} (5 tasks: UCF101, HMDB51, Kinetics-700, Breakfast, Something-Something V2),
    \item \textbf{Video Question Answering} (5 tasks: Video-MME, MVBench, NExT-QA, EgoSchema, ActivityNetQA),
    \item \textbf{Video-level Retrieval} (5 tasks: MSR-VTT, MSVD, DiDeMo, VATEX, YouCook2), and
    \item \textbf{Moment Retrieval} (3 tasks: QVHighlights, Charades-STA, MomentSeeker).
\end{enumerate}
All tasks are unified under a retrieval-based evaluation framework, where the model embeds a query (which may be text, image, video, or a combination thereof) and retrieves the correct target from a candidate pool, enabling consistent and comparable evaluation across diverse multimodal understanding capabilities. We follow the training/evaluation split defined in MMEBv2~\citep{meng2025vlm2vec}.

\subsection{Training and Evaluation Details}
\begin{table}[h]
    \centering
    \begin{tabular}{ll}
    \toprule
    \textbf{Config} & \textbf{Value} \\
    \midrule
    Optimizer & Adam \\
    Base model & Qwen3-VL-2B-Instruct \\
    Base learning rate & 2e-4 \\
    Weight decay & 0.1 \\
    Optimizer momentum & $\beta_1, \beta_2 = 0.9, 0.999$ \\
    Lora rank & 32 \\
    Lora alpha & 32 \\
    Batch size & 512 \\
    Total epochs & 1 \\
    Max sequence length & 2048 \\
    \bottomrule
    \end{tabular}
    \vspace{5pt}
    \caption{\textbf{Configuration of training setup on MMEB-V2.}}
    \label{tab:exp:training-hyper-parameters}
\end{table}

We train our model based on \textit{Qwen3-VL-2B-Instruct} using the MMEB-V2 training set; the complete set of training hyper-parameters is detailed in Table~\ref{tab:exp:training-hyper-parameters}. The evaluation metric for each individual task is Hit@1 (\ie accuracy at rank 1). The overall benchmark score is computed as a hierarchical weighted average \citep{meng2025vlm2vec}.
\newpage
\section{Visual Decoding: Methodology}
\label{sec:visual-decoding-method}

We describe the methodology for decoding TTE-Flash thinking token representations into images. Following \citet{yu2024representation}, who demonstrate that strong representation encoders with semantically rich embeddings co-occur with strong generative capabilities, we train a latent diffusion model conditioned on the frozen TTE-Flash backbone to generate images from free-form vision-language inputs.

\subsection{Overview}

The visual decoding pipeline consists of two stages:
\begin{enumerate}
    \item \textbf{RAE (Reconstruction Autoencoder):} A frozen DINOv2-B encoder \citep{oquab2023dinov2} maps target images to a semantic latent space $\mathbb{R}^{768 \times 16 \times 16}$, and a trained ViT-XL decoder reconstructs images from these latents. The encoder is never fine-tuned; only the decoder is trained.
    \item \textbf{Latent Diffusion:} A DiT$^{\text{DH}}$-XL diffusion transformer \citep{peebles2023scalable} is trained to generate DINOv2 latents conditioned on TTE-Flash thinking token representations. At inference, the generated latents are decoded into images via the RAE's ViT-XL decoder.
\end{enumerate}

The full pipeline for generating an image from a vision-language query is:
\begin{equation*}
    \text{Query (text/image)} \xrightarrow{\text{TTE-Flash}} \mathbf{H}_{\text{reg}} \xrightarrow{\text{Perceiver}} \mathbf{c} \in \mathbb{R}^{768 \times 16 \times 16} \xrightarrow{\text{DiT ODE}} \hat{\mathbf{z}} \xrightarrow{\text{RAE Decoder}} \hat{\mathbf{x}}
\end{equation*}




\subsection{Conditioning via TTE-Flash Representations}

We condition the diffusion model on the thinking token outputs of TTE-Flash. For the purposes of the visualization experiments we use a fixed length of 8 tokens produced by the frozen TTE-Flash backbone of 2048-dimensional embeddings for a given vision-language query.

\paragraph{Perceiver Resampler.} To map the TTE-Flash output to a fixed length of the spatial conditioning signal, we use a Perceiver resampler. A set of 256 learned latent queries $\mathbf{Q} \in \mathbb{R}^{256 \times 768}$ cross-attend to the projected TTE-Flash features via multi-head attention (8 heads):
\begin{equation*}
    \mathbf{c} = \text{LayerNorm}\!\left(\mathbf{Q} + \text{CrossAttn}(\mathbf{Q},\, \text{Linear}(\mathbf{H}_{\text{think}}),\, \text{Linear}(\mathbf{H}_{\text{think}}))\right)
\end{equation*}
where the linear projection maps from 2048 to 768 dimensions. The output is reshaped to $\mathbf{c} \in \mathbb{R}^{768 \times 16 \times 16}$, matching the DiT's spatial latent dimensions.

\subsection{DiT$^{\text{DH}}$-XL: Diffusion Transformer}

We use a dual-depth Diffusion Transformer (DiT$^{\text{DH}}$-XL) with a DDT (Denoising Diffusion Transformer) head architecture:

\paragraph{Encoder.} 28 transformer layers (hidden size 1152, 16 heads) process the noisy latent $\mathbf{x}_t$, patchified and combined with 2D rotary positional embeddings. Each layer uses Adaptive LayerNorm (AdaLN) modulated by the sum of timestep and conditioning embeddings, with SwiGLU feed-forward networks and RMSNorm.

\paragraph{DDT Decoder Head.} 2 transformer layers (hidden size 2048, 16 heads) receive the encoder output as \emph{per-token} conditioning (not a single global vector), enabling spatially-aware denoising. The original noisy input is re-embedded into the decoder space and conditioned on the thinking tokens via per-token scale, shift, and gate operations.

\paragraph{Flow Matching.} We train with a linear interpolation path: $\mathbf{x}_t = (1-t)\,\mathbf{x}_1 + t\,\mathbf{x}_0$, where $\mathbf{x}_1$ is the data (DINOv2 latent) and $\mathbf{x}_0 \sim \mathcal{N}(\mathbf{0}, \mathbf{I})$. The model predicts the velocity $\mathbf{v}(\mathbf{x}_t, t)$ with MSE loss against the target velocity $\mathbf{u}_t = \mathbf{x}_0 - \mathbf{x}_1$. We sample timesteps from a logit-normal distribution $\text{logit}(t) \sim \mathcal{N}(0, 1)$ with a dimension-dependent shift factor $s = \sqrt{d/4096}$ where $d = 768 \times 16 \times 16$.

\paragraph{Sampling.} We use Euler ODE integration with 50 steps and classifier-free guidance (CFG) with scale 4.0. During training, 10\% caption dropout enables CFG at inference.

\paragraph{Training the Diffusion Transformer}: Trained for 200 epochs on the MMEB training set (243K samples spanning MSCOCO, VisualNews, CIRR, and NIGHTS) with an effective batch size of 1024, AdamW optimizer (lr $= 10^{-4} \to 10^{-5}$ linear decay, 20-epoch warmup), EMA decay 0.9995, and gradient clipping at 1.0.

\subsection{Qualitative Results}

Tables in Appendix \ref{image_vis} show representative successful and failure cases across three MMEB tasks. For \textbf{text-to-image} tasks (MSCOCO, VisualNews), the model receives only a text caption and generates an image grounded to the described content. For \textbf{composed retrieval} tasks (CIRR), the model receives both a reference image and a textual modification instruction, and generates an image that composes both signals. Successful cases demonstrate that the TTE-Flash thinking tokens encode rich visual semantics beyond what is needed for retrieval alone. Failure cases reveal limitations in fine-grained spatial reasoning and faithfulness to complex compositional instructions.

\section{More CoT Examples Generated from Think Tokens \label{more_examples}}
\begin{table}[h!]
\centering
\small
\begin{tabular}{@{}p{0.47\textwidth}@{\hspace{0.04\textwidth}}p{0.47\textwidth}@{}}
\toprule
\multicolumn{2}{@{}l@{}}{\textbf{Dataset: IMAGENET--1K}} \\
\midrule
{\centering\includegraphics[width=0.40\linewidth]{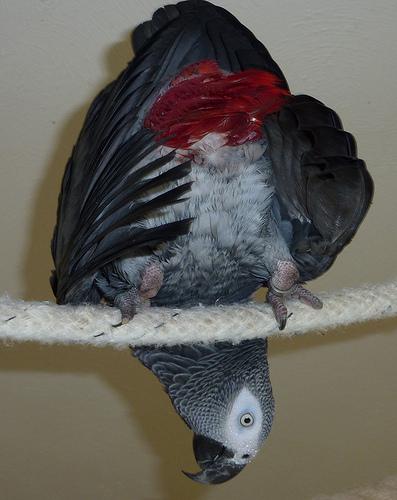}\par}
\textbf{\textcolor{blue}{Prompt:}} Represent the given image for classification \newline
\textbf{\textcolor{blue}{Retrieval Target:}} African grey, African gray, Psittacus erithacus
&
{\centering\includegraphics[width=0.40\linewidth]{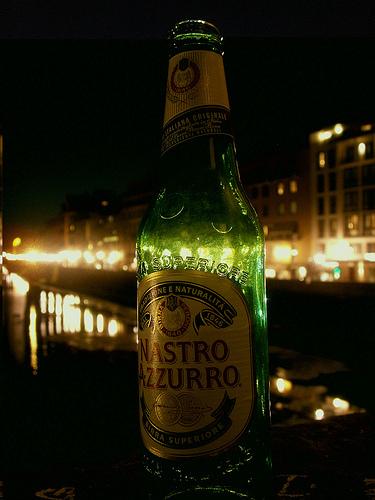}\par}
\textbf{\textcolor{blue}{Prompt:}} Represent the given image for classification \newline
\textbf{\textcolor{blue}{Retrieval Target:}} beer bottle
\\
\midrule
\textbf{\textcolor{blue}{CoT-16 Generation:}}
I will identify the bird in the image and determine its species based on its physical characteristics.  The image shows a large, grey parrot with a white head and chest, perched on a rope. The bird has a prominent crest on its head and a dark grey body.  The bird's plumage, particularly the combination of grey feathers on the back and wings with a white head and chest, is characteristic of an African grey parrot (Psittacus erithacus). This species is known for its distinctive appearance and is often associated with the term "African grey."  African grey
&
\textbf{\textcolor{blue}{CoT-16 Generation:}}
I will identify the main object in the image and determine its classification based on its appearance and context.  A green glass bottle with a label is prominently displayed in the foreground. The label features text and a logo, and the bottle has a cork stopper. The background shows a cityscape at night with illuminated buildings.  The object is clearly a bottle, specifically a glass bottle with a label and a cork stopper. This type of container is commonly used for beverages.  bottle
\\[4pt]
\textbf{\textcolor{blue}{CoT-8 Generation:}}
I will identify the bird in the image and match it to the provided classification.  The image shows a black macaw perched on a branch. The bird has a distinctive red crest, white face, and a long tail.  The bird's features, particularly the red crest and white face, are characteristic of a male black macaw.  black macaw  The bird in the image has a black body, a red crest, and a white face, which are key features of a male black macaw.
&
\textbf{\textcolor{blue}{CoT-8 Generation:}}
I will identify the main object in the image and describe its characteristics to determine the correct classification.  The image shows a green glass bottle with a label that reads "Bier" (German for beer) and a logo. The bottle is placed on a wooden surface, likely a table, in front of a lit lamp.  The object is clearly a bottle, specifically one containing a beverage. The label identifies it as "Bier," which is beer. Therefore, the correct classification for this object is "bottle."  bottle
\\[4pt]
\textbf{\textcolor{blue}{CoT-4 Generation:}}
I will identify the bird in the image and describe its key features to determine its species.  The image shows a large, predominantly black bird with a white patch on its chest and a red beak. It is perched on a wooden structure and has a distinctive pattern of white feathers on its head and neck.  The bird's appearance, particularly the white patch on its chest and the red beak, is characteristic of a Barbary Ape (African Grey Parrot). The combination of these features strongly suggests it is a Barbary Ape.  African Grey Parrot, Barbary Ape
&
\textbf{\textcolor{blue}{CoT-4 Generation:}}
I will identify the main objects in the image and determine which one is the most likely subject of the photograph.  The image shows a collection of items on a table. There are several bottles of beer, including a large bottle of Budweiser and smaller bottles of Bud Light. There are also two glasses, one filled with a dark liquid and the other empty. A red plastic bottle cap is visible on the table.  The presence of multiple beer bottles and glasses strongly suggests that the subject of the photograph is beer. The specific brands and types of beer are also visible.  beer
\\[4pt]
\textbf{\textcolor{blue}{CoT-2 Generation:}}
I will identify the animal in the image and determine its classification based on its physical characteristics.  The image shows a small, fluffy, brown animal with a long, curved tail. It has large, dark eyes and a prominent, rounded head.  The animal's features, including its fluffy fur, large eyes, and rounded head, are characteristic of a domestic cat. The presence of a long, curved tail is also consistent with a cat's anatomy.  cat
&
\textbf{\textcolor{blue}{CoT-2 Generation:}}
I will identify the main object in the image and determine its classification based on its appearance and context.  The image shows a clear glass bottle filled with a dark liquid, placed on a wooden surface. The bottle has a label with the word "VODKA" visible.  The object in the image is a glass bottle containing a dark liquid. The label on the bottle clearly reads "VODKA", which is a type of alcoholic beverage. This classification is based on the visual evidence of the bottle's material, shape, and the text on the label.  
\\
\bottomrule
\end{tabular}

\end{table}

\begin{table}[ht]
\centering
\small
\begin{tabular}{@{}p{0.47\textwidth}@{\hspace{0.04\textwidth}}p{0.47\textwidth}@{}}
\toprule
\multicolumn{2}{@{}l@{}}{\textbf{Dataset: IMAGENET--1K}} \\
\midrule
{\centering\includegraphics[width=0.5\linewidth]{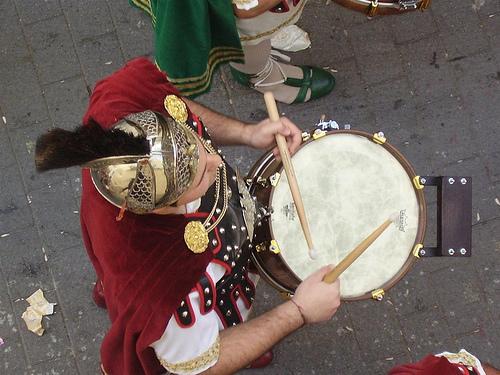}\par}
\textbf{\textcolor{blue}{Prompt:}} Represent the given image for classification \newline
\textbf{\textcolor{blue}{Retrieval Target:}} drum, membranophone, tympan
&
{\centering\includegraphics[width=0.6\linewidth]{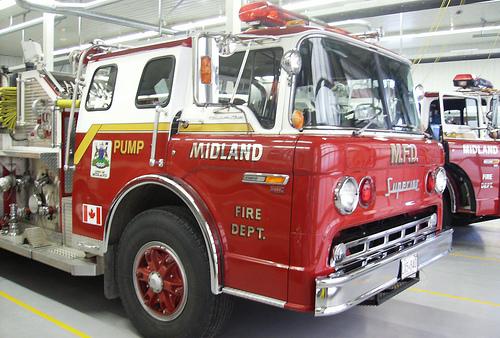}\par}
\textbf{\textcolor{blue}{Prompt:}} Represent the given image for classification \newline
\textbf{\textcolor{blue}{Retrieval Target:}} fire engine, fire truck
\\
\midrule
\textbf{\textcolor{blue}{CoT-16 Generation:}}
I will identify the main object being played by the person in the image and determine its classification based on its appearance and function.  The image shows a person wearing a red and gold helmet and armor, holding a large drumstick. The drumstick is being used to strike a large, circular drum with a red and white pattern.  The object being played is a large drum, which is a type of percussion instrument. The drum is characterized by its size, shape, and the way it is played with a drumstick.  drum
&
\textbf{\textcolor{blue}{CoT-16 Generation:}}
I will identify the main body of the vehicle and its primary function based on the visual evidence provided in the image.  The image shows a large, red vehicle with the words "MIDLAND" and "FIRE DEPT" clearly visible on its side. It has a large front grille, a hood, and a cab area. The vehicle is equipped with a large water tank and hoses, which are typical components of a fire engine.  The vehicle's design, including the cab, the large water tank, and the hose connections, are characteristic features of a fire engine. The text "FIRE DEPT" further confirms its purpose.  fire engine  The image shows a large, red vehicle with the words "MIDLAND" and "FIRE DEPT" clearly visible on its side. It has a large front grille, a hood, and a cab area. The vehicle is equipped with a large water tank and hoses, which are typical components of a fire engine. The text "FIRE DEPT" further confirms its purpose.
\\[4pt]
\textbf{\textcolor{blue}{CoT-8 Generation:}}
I will identify the main object being held by the person in the image and determine its classification.  The image shows a person wearing a helmet and armor, holding a long, curved instrument with a handle. The instrument has a distinct shape and appears to be made of metal.  The object is a long, curved instrument with a handle, which is characteristic of a drumstick or mallet used in playing a musical instrument.  drumstick, mallet
&
\textbf{\textcolor{blue}{CoT-8 Generation:}}
I will identify the main object in the image and describe its characteristics to determine the correct classification.  The image shows a red fire truck with a large white stripe on the side, parked in front of a red brick wall. There are two children standing near the front of the vehicle.  The vehicle is clearly a fire truck, identifiable by its red color, the white stripe, and the presence of a ladder on the side. The children are standing near the front of the truck, which is a common sight for fire trucks.  fire truck
\\[4pt]
\textbf{\textcolor{blue}{CoT-4 Generation:}}
I will identify the main components of the drum set shown in the image and determine which instrument is the bass drum.  The image displays a complete drum kit on a stand, including various drums and cymbals.  The drum set consists of several drums of different sizes and shapes, including a large bass drum at the back, a snare drum in the center, and multiple toms. The bass drum is the largest and lowest-pitched drum in the set, located at the back.  bass drum  The image shows a complete drum set on a stand, including various drums and cymbals.
&
\textbf{\textcolor{blue}{CoT-4 Generation:}}
I will identify the main components of the vehicle shown in the image and match them to the provided classification options.  The image displays a large, red fire engine with visible features such as a cab, a ladder, and various compartments. The text on the side reads "FIRE DEPT" and "FIRE DEPT 100".  The vehicle is clearly a fire engine, which is a type of emergency vehicle used for firefighting. The presence of a ladder and the "FIRE DEPT" label confirms its function. Therefore, the correct classification is "fire engine".  fire engine  The image shows a red fire engine with a ladder and the text "FIRE DEPT" on its side.
\\[4pt]
\textbf{\textcolor{blue}{CoT-2 Generation:}}
I will identify the main object in the image and determine its classification based on its appearance and context.  The image shows a large, black, circular object with a handle, positioned on a wooden floor. It has a distinct shape and appears to be made of metal.  The object in the image is a drum, specifically a bass drum, which is a type of percussion instrument. The presence of a large drumhead and a frame is characteristic of this instrument.  bass drum  The object in the image is a large, black, circular instrument with a handle, which is clearly identifiable as a bass drum.
&
\textbf{\textcolor{blue}{CoT-2 Generation:}}
I will identify the main features of the vehicle in the image to determine its classification.  The image shows a large, red, boxy vehicle with a prominent front grille and headlights. It has a long, flat roof and a large, open cargo area at the back. The vehicle is parked on a paved surface.  The vehicle has a distinct, boxy shape with a large cargo area, which is characteristic of a truck. The presence of a large front grille and headlights further supports this classification.  truck  The vehicle has a large, boxy shape with a prominent front grille and headlights, and a long, flat roof.
\\
\bottomrule
\end{tabular}
\end{table}


\begin{table}[ht]
\centering
\small
\begin{tabular}{@{}p{0.47\textwidth}@{\hspace{0.04\textwidth}}p{0.47\textwidth}@{}}
\toprule
\multicolumn{2}{@{}l@{}}{\textbf{Dataset: MSCOCO}} \\
\midrule
{\centering\includegraphics[width=0.7\linewidth]{images/mmeb_images/MSCOCO/000000003845.jpg}\par}
\textbf{\textcolor{blue}{Prompt:}} Crop the image to to isolate the object labeled as "broccoli"
&
{\centering\includegraphics[width=0.5\linewidth]{images/mmeb_images/MSCOCO/000000458992.png}\par}
\textbf{\textcolor{blue}{Prompt:}} Crop the image to to isolate the object labeled as "pizza"
\\
\midrule
\textbf{\textcolor{blue}{CoT-16 Generation:}}
I will identify the key visual features of the broccoli in the image to match it with a similar example.  The image shows a plate of food with a large piece of broccoli, a small white bowl, and a piece of meat.  The broccoli is the most prominent vegetable on the plate, and its green color and floret structure are key identifying features.  Look for an image with a similar green vegetable, such as broccoli or cauliflower, on a plate.
&
\textbf{\textcolor{blue}{CoT-16 Generation:}}
I will identify the key visual features of the pizza and the person eating it to determine a matching image.  A large, round pizza with a thick crust, topped with melted cheese, pepperoni slices, and mushrooms, is served on a wooden table. A woman with dark hair is sitting at the table, holding a knife and fork, and appears to be in the process of eating the pizza.  The most prominent feature is the pizza itself, which is the central subject of the image. The woman eating it is secondary.
\\[4pt]
\textbf{\textcolor{blue}{CoT-8 Generation:}}
I will identify the key visual features of the food items in the image to match a similar dish.  A plate of food with a white rice base, topped with pieces of chicken, vegetables, and a sauce.  The dominant feature is the white rice, which is a common component in many Asian dishes. A matching image should also feature rice as a primary element.  Look for an image containing rice as a main component.
&
\textbf{\textcolor{blue}{CoT-8 Generation:}}
I will identify the key visual features of the pizza to match with a similar image.  A large, round pizza with a golden-brown crust and a red sauce base, topped with slices of pepperoni and melted cheese.  The dominant feature is the pizza itself, which is the central subject.  To find a similar image, I need to look for another image that contains a pizza, with similar characteristics such as the type of toppings and the overall appearance.  The matching image should feature a pizza with comparable toppings and presentation style.
\\[4pt]
\textbf{\textcolor{blue}{CoT-4 Generation:}}
I will identify the key visual features to match.  A large pot of green vegetables on a stovetop.  The dominant feature is the large pot of green vegetables, likely spinach or broccoli, on the stovetop.  Look for an image containing a large pot of green vegetables on a stovetop.
&
\textbf{\textcolor{blue}{CoT-4 Generation:}}
I will identify the key visual features of the pizza to match.  A large, round, golden-brown pizza with melted cheese and red tomato sauce, served on a dark wooden table with a white cloth.  The dominant feature is the pizza itself, which is the main subject. The key matching criteria would be its size, color, and the presence of cheese and sauce.  Look for a similar-sized, golden-brown pizza with melted cheese and tomato sauce.  A large, round, golden-brown pizza with melted cheese and red tomato sauce, served on a dark wooden table with a white cloth..
\\[4pt]
\textbf{\textcolor{blue}{CoT-2 Generation:}}
I will identify the key visual features of the dog in the image to match a similar object.  The image shows a dog lying on a bed, with a white blanket and a red blanket visible. The dog has a black and white coat, and its body is positioned on the bed.  The key features are the dog's body, its black and white fur, and its position on the bed. A matching image should show a similar dog with the same coloration and pose.  Look for an image of a dog lying on a bed with a black and white coat.  A dog lying on a bed with a white blanket and a red blanket. The dog has a black and white coat and is positioned on the bed.
&
\textbf{\textcolor{blue}{CoT-2 Generation:}}
I will identify the key visual features of the coffee cup to match it with a similar image.  A close-up of a coffee cup with a dark liquid, a white lid, and a brown handle. The cup has a textured surface and a visible handle.  The key features are the dark liquid, the white lid, and the brown handle. A matching image should show a coffee cup with a similar dark liquid, white lid, and brown handle.  Look for an image of a coffee cup with a dark liquid, a white lid, and a brown handle.
\\
\bottomrule
\end{tabular}
\end{table}

\begin{table}[ht]
\centering
\small
\begin{tabular}{@{}p{0.47\textwidth}@{\hspace{0.04\textwidth}}p{0.47\textwidth}@{}}
\toprule
\textbf{Dataset: MSCOCO}
&
\textbf{Dataset: MSCOCO\_I2T} \\
\midrule
{\centering\includegraphics[width=0.5\linewidth]{images/mmeb_images/MSCOCO/000000243204.jpg}\par}
\textbf{\textcolor{blue}{Prompt:}} Crop the image to to isolate the object labeled as "sandwich"
&
{\centering\includegraphics[width=0.75\linewidth]{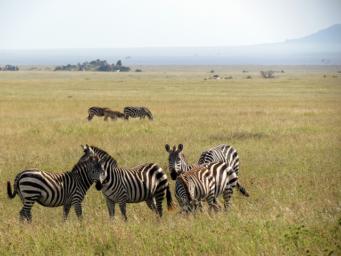}\par}
\textbf{\textcolor{blue}{Prompt:}} Find an image caption describing the given everyday image. \newline
\textbf{\textcolor{blue}{Retrieval Target:}} A group of zebra's playing and grazing in a field.
\\
\midrule
\textbf{\textcolor{blue}{CoT-16 Generation:}}
I will identify the key visual features of the sandwich and the person preparing it.  A person is making a sandwich on a black griddle. The sandwich has a toasted bun, melted cheese, and what appears to be ham or turkey. The person is wearing a dark jacket and is holding a spatula.  The dominant visual elements are the sandwich itself, its ingredients, and the cooking process. A matching image should also feature a sandwich being prepared on a griddle or similar cooking surface.  Look for an image of a sandwich being prepared on a griddle or similar cooking surface.
&
\textbf{\textcolor{blue}{CoT-16 Generation:}}
I will analyze the image to identify the main subjects and their environment, then construct a descriptive caption based on what is visible.  The image shows a group of zebras standing in a grassy field. In the background, there are several large, dark shapes that appear to be trees or bushes.  The primary subjects are clearly zebras. They are located in an open, grassy area, which is characteristic of a savanna or grassland. The presence of trees or bushes in the distance further supports this conclusion.  A group of zebras standing in a grassy field with trees in the background.
\\[4pt]
\textbf{\textcolor{blue}{CoT-8 Generation:}}
I will identify the key visual features of the image to match.  A person is cooking on a stovetop, using a spatula to flip a piece of meat.  The main subject is the cooking process itself, specifically the act of flipping food on a grill or pan. The key elements are the person, the cooking utensil (spatula), and the food being prepared.  To find a similar image, I need to look for an image that captures a comparable cooking scene, focusing on the same action of flipping food on a grill or pan.
&
\textbf{\textcolor{blue}{CoT-8 Generation:}}
I will describe the main subjects and their environment in the image to create an accurate caption.  A group of zebras is standing in a grassy field with a backdrop of trees and a clear sky.  The image clearly shows a herd of zebras in a natural, open landscape. The zebras are positioned on the ground, and the background consists of trees and a wide expanse of sky. This visual evidence supports the description of a herd of zebras in a field.  A herd of zebras in a field.
\\[4pt]
\textbf{\textcolor{blue}{CoT-4 Generation:}}
I will identify the key visual elements to match.  A person is cooking a sandwich on a griddle over a gas flame.  The primary subject is the sandwich cooking on a griddle. The key matching criteria would be the presence of a sandwich on a griddle, which is a common cooking scene.  Look for an image featuring a sandwich cooking on a griddle.
&
\textbf{\textcolor{blue}{CoT-4 Generation:}}
I will identify the main subjects in the image and describe their actions to create a caption.  The image shows a herd of zebras standing on a grassy plain. In the background, there are several other animals, including what appear to be wildebeests and a few elephants.  The central focus is a group of zebras. They are standing still on the ground. The background contains other animals, but the zebras are the primary subject.  A herd of zebras standing on the grass.
\\[4pt]
\textbf{\textcolor{blue}{CoT-2 Generation:}}
I will identify the key visual features of the image to match with a similar scene.  The image shows a person in a red shirt standing in front of a white wall with a large, colorful mural on it. The mural has a prominent red and yellow design, possibly a stylized face or abstract pattern.  To find a similar image, I need to look for an image that contains a person in a red shirt in front of a wall with a large, colorful mural. The key elements are the person, the red shirt, and the mural.  The matching image should feature a person in a red shirt in front of a wall with a large, colorful mural.  A person in a red shirt standing in front of a white wall with a large, colorful mural on it.
&
\textbf{\textcolor{blue}{CoT-2 Generation:}}
I will describe the main subjects and their environment to create an accurate caption.  The image shows a herd of elephants standing in a dry, grassy field with scattered trees in the background. The elephants are the central focus of the scene.  The image clearly depicts a group of elephants in a natural, open landscape. This aligns perfectly with the description provided in the correct answer.  A herd of elephants in a dry grassland.
\\
\bottomrule
\end{tabular}
\end{table}


\begin{table}[ht]
\centering
\small
\begin{tabular}{@{}p{0.47\textwidth}@{\hspace{0.04\textwidth}}p{0.47\textwidth}@{}}
\toprule
\multicolumn{2}{@{}l@{}}{\textbf{Dataset: MSCOCO\_I2T}} \\
\midrule
{\centering\includegraphics[width=0.6\linewidth]{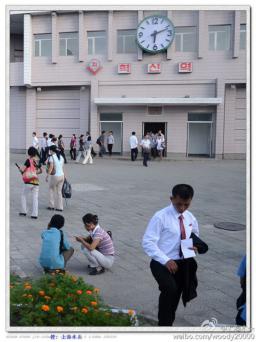}\par}
\textbf{\textcolor{blue}{Prompt:}} Find an image caption describing the given everyday image. \newline
\textbf{\textcolor{blue}{Retrieval Target:}} A bunch of people sit in an open court yard.
&
{\centering\includegraphics[width=0.7\linewidth]{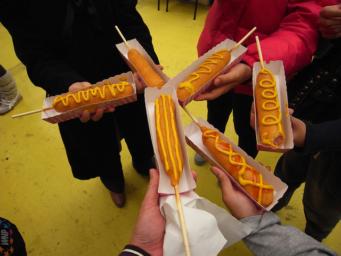}\par}
\textbf{\textcolor{blue}{Prompt:}} Find an image caption describing the given everyday image. \newline
\textbf{\textcolor{blue}{Retrieval Target:}} A group of people holding some hot dogs in his hand.
\\
\midrule
\textbf{\textcolor{blue}{CoT-16 Generation:}}
I will identify the main subject and the setting in the image to create a descriptive caption.  The image shows a large group of people gathered in front of a school building. The building has a clock tower and a sign that reads "School" in large letters. The people are standing in a line, looking towards the building.  The scene depicts a school environment with students or staff standing in front of a school building. The presence of a clock tower and the word "School" confirms the location.  A group of students standing in front of a school building.
&
\textbf{\textcolor{blue}{CoT-16 Generation:}}
I will describe the main subjects and their actions in the image to create a caption.  The image shows several hot dogs on sticks, held by people's hands. The hot dogs are placed on a yellow surface, possibly a table or tray.  The central focus is the hot dogs being held up by hands. The people are actively holding them, which is the key action in the scene.  A group of people holding hot dogs.
\\[4pt]
\textbf{\textcolor{blue}{CoT-8 Generation:}}
I will describe the main subjects and their actions in the image to create an accurate caption.  A group of people is gathered near a large clock tower. Some individuals are standing and talking, while others are seated on benches. A few people are walking around the area.  The image shows a crowd of people, some standing and others sitting, near a clock tower. This matches the description of a group of people standing and talking near a clock tower.  A group of people standing and talking near a clock tower.
&
\textbf{\textcolor{blue}{CoT-8 Generation:}}
I will describe the main subjects and their actions in the image to create an accurate caption.  A group of people are gathered around a table, holding and eating hot dogs and buns.  The image shows several individuals, likely friends or family, sitting together and actively eating food. The food consists of hot dogs and buns, which are being held and consumed by the people.  A group of friends or family eating hot dogs and buns.
\\[4pt]
\textbf{\textcolor{blue}{CoT-4 Generation:}}
I will identify the main subjects in the image and describe their actions to create a caption.  The image shows a large group of people, primarily children and teenagers, gathered in front of a school building. They are all looking towards the camera, suggesting they are posing for a photograph.  The central focus is a group of students, many of whom are holding up their phones to take pictures. This indicates they are actively participating in a photo session.  A group of students are taking pictures in front of a school.
&
\textbf{\textcolor{blue}{CoT-4 Generation:}}
I will describe the main subject and action in the image to create an accurate caption.  The image shows a person holding a long, thin, black object with a white tip. The object is being used to dip into a container of food, which appears to be a hot dog or sausage. The person is wearing a dark-colored shirt.  The central focus is the black object being dipped into the food. The food itself is a long, cylindrical item, consistent with a hot dog or sausage. The person's hand is holding the object, indicating the action of dipping.  A person dipping a long black object into a hot dog.
\\[4pt]
\textbf{\textcolor{blue}{CoT-2 Generation:}}
I will describe the main subjects and their actions in the image to create an accurate caption.  The image shows a group of people standing on a sidewalk next to a street. One person is holding a camera, another is holding a phone, and a third person is looking at a tablet. The background includes buildings and trees.  The people are gathered on the sidewalk, which is a common place for pedestrians. Their actions of holding devices suggest they are using them for navigation or capturing images.
&
\textbf{\textcolor{blue}{CoT-2 Generation:}}
I will describe the main subject and the action taking place in the image to create an accurate caption.  The image shows a person holding a large, colorful, and decorated cake. The cake is placed on a table, and the person is wearing a red shirt. The background is a simple white wall.  The central focus of the image is a cake that has been elaborately decorated with various colors and patterns. The person holding it is wearing a red shirt, which matches the color of the cake's frosting.
\\
\bottomrule
\end{tabular}
\end{table}

\begin{table}[ht]
\centering
\small
\begin{tabular}{@{}p{0.47\textwidth}@{\hspace{0.04\textwidth}}p{0.47\textwidth}@{}}
\toprule
\multicolumn{2}{@{}l@{}}{\textbf{Dataset: MSCOCO\_I2T}} \\
\midrule
{\centering\includegraphics[width=0.5\linewidth]{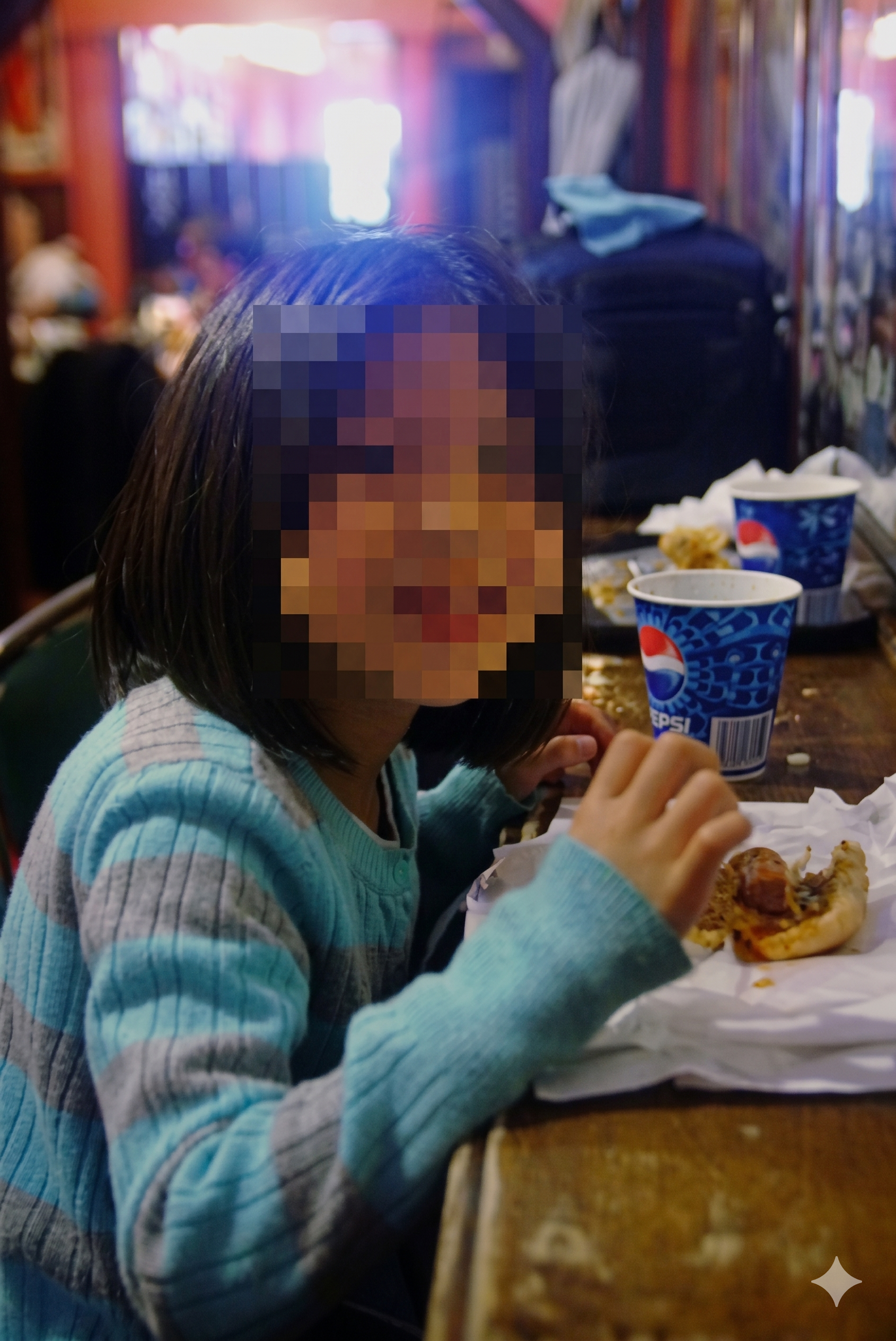}\par}
\textbf{\textcolor{blue}{Prompt:}} Find an image caption describing the given everyday image. \newline
\textbf{\textcolor{blue}{Retrieval Target:}} A little girl posing for a picture while eating food.
&
{\centering\includegraphics[width=0.7\linewidth]{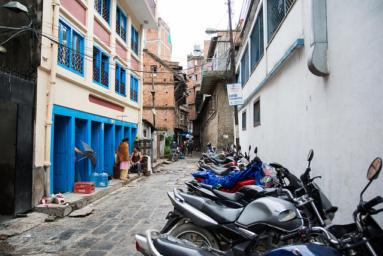}\par}
\textbf{\textcolor{blue}{Prompt:}} Find an image caption describing the given everyday image. \newline
\textbf{\textcolor{blue}{Retrieval Target:}} A row of motorcycles parked in front of a building.
\\
\midrule
\textbf{\textcolor{blue}{CoT-16 Generation:}}
I will describe the main subject and their immediate surroundings to create a caption.  A young girl with dark hair is sitting at a table. She is wearing a blue and white striped shirt and is holding a small, light-colored pastry in her hands. In front of her is a plate with a sandwich and a drink.  The image clearly shows a young girl as the central figure. She is positioned at a table, which is the primary setting. The food items on the table, including a sandwich and a drink, are directly in front of her. This combination of elements forms the core of the caption.  A young girl sitting at a table with a sandwich and a drink.
&
\textbf{\textcolor{blue}{CoT-16 Generation:}}
I will describe the main subject and setting of the image to create a caption.  The image shows a narrow street lined with rows of small, colorful buildings. In the foreground, there are several motorcycles parked on the sidewalk. The buildings have blue and white doors and windows, and the street is paved with cobblestones.  The central focus is the street itself, which is filled with parked motorcycles. The buildings are tightly packed along the sides of the street, creating a dense urban environment.  A street full of parked motorcycles and buildings.
\\[4pt]
\textbf{\textcolor{blue}{CoT-8 Generation:}}
I will describe the main subject and their action in the image to create an accurate caption.  A young girl with dark hair is sitting at a table, holding a blue cup. She is looking down at the table, which has a plate of food and a glass of water in front of her.  The central figure is a young girl. She is actively engaged with the food on the table, specifically holding a cup and looking down at it. This action suggests she is eating or drinking from the cup.  A young girl eating from a cup.
&
\textbf{\textcolor{blue}{CoT-8 Generation:}}
I will describe the main subjects and setting of the image to create an accurate caption.  The image shows a street scene with several bicycles parked along the left side. In the background, there are buildings with blue and white facades, and a person is visible on a bicycle in the distance.  The key elements are the bicycles parked on the sidewalk and the buildings in the background. The presence of a person on a bicycle adds context to the scene.  A row of bicycles parked on the side of a street with buildings in the background.
\\[4pt]
\textbf{\textcolor{blue}{CoT-4 Generation:}}
I will identify the main subject and their activity in the image to create a descriptive caption.  A young girl with dark hair is sitting at a table. She is holding a small, round, pink object, possibly a toy or a piece of food. In front of her is a glass of water and a plate with a sandwich.  The image clearly shows a child engaged in eating. The child is holding a pink item, which appears to be a toy or a snack, and is positioned over a sandwich on a plate. This directly matches the description of a child eating a sandwich.  A young girl eating a sandwich.
&
\textbf{\textcolor{blue}{CoT-4 Generation:}}
I will identify the main subject of the image, which is a street scene, and then describe its key features to create an accurate caption.  The image shows a narrow street lined with tall, dark-colored buildings. The street is paved and has a few bicycles parked along the side. The buildings have a distinct architectural style, and there are people visible in the background.  The central focus is the street itself, which is the primary subject. The surrounding buildings and the presence of people suggest a city environment. The overall scene is characterized by its narrowness and the architectural style of the buildings. 
\\[4pt]
\textbf{\textcolor{blue}{CoT-2 Generation:}}
I will identify the main subject, their clothing, and the setting to construct a descriptive caption.  A young girl with dark hair is sitting on a wooden bench. She is wearing a red shirt with a white collar and a blue skirt. In front of her is a large, colorful toy train set.  The image clearly shows a child in a red shirt and blue skirt. The child is positioned in front of a toy train set, which is a prominent feature of the scene. The background consists of a wooden bench and a wall.  A young girl in a red shirt and blue skirt sits in front of a toy train set.
&
\textbf{\textcolor{blue}{CoT-2 Generation:}}
I will describe the main elements visible in the image to create a caption.  The image shows a street scene with a large, multi-story building on the left and a smaller, single-story building on the right. A paved road runs between them, with a few cars and pedestrians. The sky above is clear and blue.  The most prominent feature is the large, multi-story building, which is clearly a commercial or office building. The smaller building is also visible, but it appears to be a residential or smaller business structure. The presence of a road and vehicles further supports the idea of a city street.
\\
\bottomrule
\end{tabular}
\end{table}

\begin{table}[ht]
\centering
\small
\begin{tabular}{@{}p{0.47\textwidth}@{\hspace{0.04\textwidth}}p{0.47\textwidth}@{}}
\toprule
\multicolumn{2}{@{}l@{}}{\textbf{Dataset: MSCOCO\_I2T}} \\
\midrule
{\centering\includegraphics[width=0.7\linewidth]{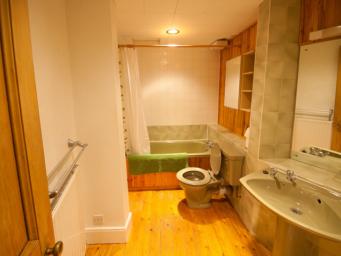}\par}
\textbf{\textcolor{blue}{Prompt:}} Find an image caption describing the given everyday image. \newline
\textbf{\textcolor{blue}{Retrieval Target:}} A bathroom with a sink, toilet, tub and mirror.
&
{\centering\includegraphics[width=0.7\linewidth]{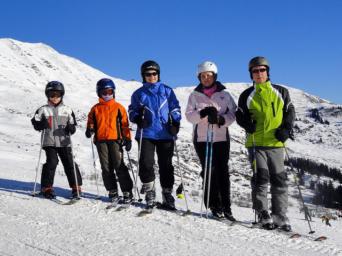}\par}
\textbf{\textcolor{blue}{Prompt:}} Find an image caption describing the given everyday image. \newline
\textbf{\textcolor{blue}{Retrieval Target:}} A group of young and old are skiing on the snow.
\\
\midrule
\textbf{\textcolor{blue}{CoT-16 Generation:}}
I will describe the main features of the bathroom shown in the image to create an accurate caption.  The image displays a bathroom with a wooden floor, a bathtub, a toilet, and a shower stall. The walls are painted a light color, and there is a wooden door leading to another room.  The key elements visible in the image are the bathtub, toilet, and shower stall, which are all standard fixtures in a bathroom. The wooden floor and walls contribute to the room's appearance.  A bathroom with a bathtub, toilet and shower stall.
&
\textbf{\textcolor{blue}{CoT-16 Generation:}}
I will describe the main subjects and their actions in the image to create a caption.  Five people are standing on a snowy slope, wearing winter gear and helmets. They are holding ski poles and looking towards the camera.  The image shows five individuals, likely skiers or snowboarders, on a snow-covered hill. They are all wearing jackets, pants, and helmets, which are typical winter sports attire. Their posture and the equipment they are holding suggest they are actively engaged in skiing or snowboarding.  Five people in winter gear skiing on a snowy hill.
\\[4pt]
\textbf{\textcolor{blue}{CoT-8 Generation:}}
I will describe the main features of the bathroom shown in the image to create an accurate caption.  The image displays a bathroom with a white toilet, a bathtub with a shower head, and a sink with a mirror above it. The walls are tiled, and there is a wooden door leading to another room.  The key elements visible in the image are the toilet, bathtub, sink, mirror, and tiled walls. These are the primary fixtures and features of the bathroom.  A bathroom with a toilet, bathtub, sink, mirror, and tiled walls.
&
\textbf{\textcolor{blue}{CoT-8 Generation:}}
I will describe the main subjects and their actions in the image to create an accurate caption.  A group of four people, including two adults and two children, are standing on a snowy slope. They are wearing winter gear and holding ski poles. The background shows a mountainous landscape under a cloudy sky.  The image clearly shows a group of four people on a snowy slope, which is the primary setting. They are dressed in winter clothing and holding ski poles, indicating they are skiing. The presence of multiple individuals and the snowy environment supports the description of a group of people skiing.  A group of people skiing on a snowy slope.
\\[4pt]
\textbf{\textcolor{blue}{CoT-4 Generation:}}
I will identify the main features of the room, such as the furniture, flooring, and fixtures, to create a descriptive caption.  The image shows a bathroom with a wooden floor, a white bathtub, a wooden vanity with a sink, and a toilet. There is also a wooden door leading to another room.  The key elements are the bathtub, the vanity, and the toilet, which are all common fixtures in a bathroom. The presence of a wooden floor and a wooden door suggests a traditional or rustic style.  A bathroom with a bathtub, a vanity, and a toilet.
&
\textbf{\textcolor{blue}{CoT-4 Generation:}}
I will identify the main subject and their activity in the image to create an accurate caption.  A group of four people is standing on a snowy slope, holding ski poles and wearing winter gear. They are positioned in front of a ski lift station.  The individuals are clearly engaged in skiing, which is the primary action depicted. The setting is a snowy mountain environment, consistent with a ski resort.  A group of people skiing on a snowy mountain.
\\[4pt]
\textbf{\textcolor{blue}{CoT-2 Generation:}}
I will identify the main features of the bathroom shown in the image to create an accurate caption.  The image shows a bathroom with a white bathtub, a white toilet, and a white sink. The walls are tiled in a light color, and there is a large mirror above the sink. The floor is covered with dark tiles.  The image displays a bathroom with a bathtub, toilet, and sink. The walls are tiled, and the floor is also tiled. The presence of a mirror above the sink is also notable.  A bathroom with a bathtub, toilet, and sink.  The image shows a bathroom with a bathtub, toilet, and sink.
&
\textbf{\textcolor{blue}{CoT-2 Generation:}}
I will describe the main subjects and their activity in the image to create an accurate caption.  The image shows a group of people on a snowy hill, with one person in a red jacket and another in a blue jacket. They are all wearing winter clothing and appear to be skiing or snowboarding down the slope.  The image clearly depicts multiple individuals engaged in winter sports on a snow-covered terrain. The presence of skis and snowboards confirms the activity.  A group of people skiing down a snowy hill.
\\
\bottomrule
\end{tabular}
\end{table}


\begin{table}[ht]
\centering
\small
\begin{tabular}{@{}p{0.47\textwidth}@{\hspace{0.04\textwidth}}p{0.47\textwidth}@{}}
\toprule
\multicolumn{2}{@{}l@{}}{\textbf{Dataset: OK\_VQA}} \\
\midrule
{\centering\includegraphics[width=0.7\linewidth]{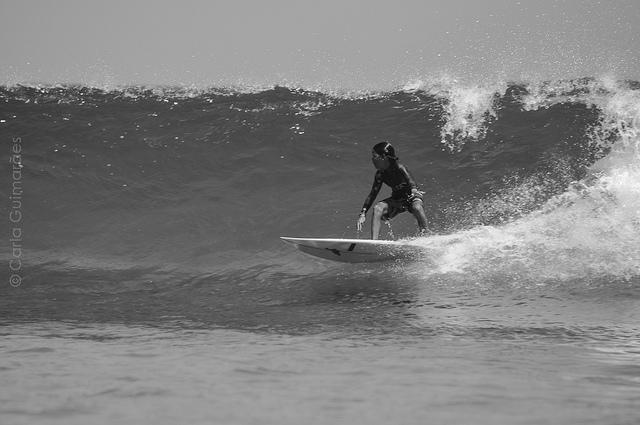}\par}
\textbf{\textcolor{blue}{Prompt:}} Represent the given image with the following question: Can you guess the sea name shown in this picture? \newline
\textbf{\textcolor{blue}{Retrieval Target:}} pacific ocean
&
{\centering\includegraphics[width=0.7\linewidth]{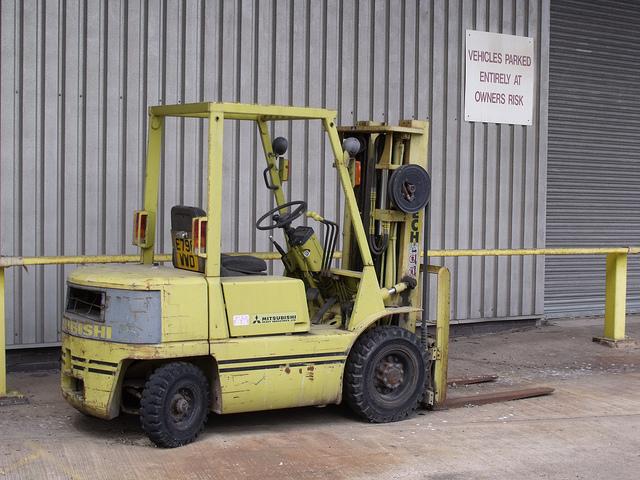}\par}
\textbf{\textcolor{blue}{Prompt:}} Represent the given image with the following question: In what sort of building would this vehicle be found? \newline
\textbf{\textcolor{blue}{Retrieval Target:}} warehouse
\\
\midrule
\textbf{\textcolor{blue}{CoT-16 Generation:}}
I will analyze the visual elements in the image to determine the likely location of the sea.  The image shows a person surfing on a wave. The water appears dark blue, and the sky is overcast with grey clouds.  The presence of a surfer on a wave is a strong indicator of a coastal area. The specific type of wave, which is a large, breaking wave, is characteristic of certain oceanic regions. The overall scene suggests a large body of water with waves suitable for surfing.  The sea is likely in the Pacific Ocean.
&
\textbf{\textcolor{blue}{CoT-16 Generation:}}
I will identify the type of vehicle shown and determine its typical location based on the surrounding environment.  A yellow forklift is parked inside a large, industrial-looking building with a high ceiling and metal walls. The building has a sign that reads "Forklifts" and "Equipment" on the wall.  The presence of a forklift, a large warehouse-like structure, and the sign indicating equipment storage strongly suggests that this is a commercial or industrial facility where such equipment is stored and used.  warehouse
\\[4pt]
\textbf{\textcolor{blue}{CoT-8 Generation:}}
I will identify the location of the wave and determine the country it is in.  A surfer is riding a wave on a surfboard. The wave is large and white, typical of a breaking wave. The background is a blue ocean.  The wave is a large, white crest that is breaking. This type of wave is characteristic of a surf spot in a coastal region. The location of the wave is in the ocean.  The wave is in the ocean.  ocean
&
\textbf{\textcolor{blue}{CoT-8 Generation:}}
I will analyze the visual cues in the image to determine the type of business or activity taking place.  The image shows a large yellow forklift truck parked in a large, open-sided metal structure. The structure has a high ceiling and appears to be made of corrugated metal sheets. There are no visible signs of a retail store, office, or residential building.  The presence of a forklift truck, which is used for lifting and moving heavy objects, strongly suggests that the space is used for industrial or commercial purposes. The large, open-sided structure is typical of a warehouse or storage facility designed for such operations.  warehouse
\\[4pt]
\textbf{\textcolor{blue}{CoT-4 Generation:}}
I will analyze the image to identify the type of watercraft shown and determine its classification.  The image displays a person on a surfboard riding waves in the ocean. The surfboard is a long, narrow board with a pointed nose, which is characteristic of a surfboard.  The object in the image is a surfboard, which is a type of watercraft specifically designed for riding waves. Therefore, it is classified as a surfboard.  surfboard  The image shows a person on a surfboard riding waves in the ocean. The surfboard is a long, narrow board with a pointed nose, which is characteristic of a surfboard.
&
\textbf{\textcolor{blue}{CoT-4 Generation:}}
I will identify the primary object in the image and determine its function based on its surroundings and the visible equipment.  A large, industrial forklift is parked inside a warehouse or storage facility. The forklift has a high mast and a long boom, which are characteristic of a reach stacker or a similar type of material handling equipment. The warehouse has a concrete floor and metal shelving units.  The presence of a forklift, a common piece of industrial equipment used for lifting and moving heavy loads, strongly suggests that the primary function of this space is to store and manage goods. The warehouse is designed to hold and organize materials, which aligns with the purpose of a storage facility.  storage
\\[4pt]
\textbf{\textcolor{blue}{CoT-2 Generation:}}
I will analyze the visual elements in the image to determine the type of watercraft being used.  The image shows a person on a large, white, inflatable boat with a red and yellow stripe. The boat is moving through blue water, and the background features a distant shoreline with trees and buildings.  The boat's design, with its large inflatable structure and the way it's being used for recreation, is characteristic of a surfboard. The person is sitting on the board, which is a common way to ride waves.  surfboard
&
\textbf{\textcolor{blue}{CoT-2 Generation:}}
I will analyze the image to identify the type of vehicle and its primary function, which will help determine the correct category.  The image shows a large, white, boxy vehicle with a flatbed and a crane-like arm extending from the back. This vehicle is used for moving heavy materials, such as concrete or asphalt, from one location to another.  The vehicle's design, including its large size, flatbed, and crane arm, is characteristic of a dump truck.
\\
\bottomrule
\end{tabular}
\end{table}

\begin{table}[ht]
\centering
\small
\begin{tabular}{@{}p{0.47\textwidth}@{\hspace{0.04\textwidth}}p{0.47\textwidth}@{}}
\toprule
\multicolumn{2}{@{}l@{}}{\textbf{Dataset: OK\_VQA}} \\
\midrule
{\centering\includegraphics[width=0.7\linewidth]{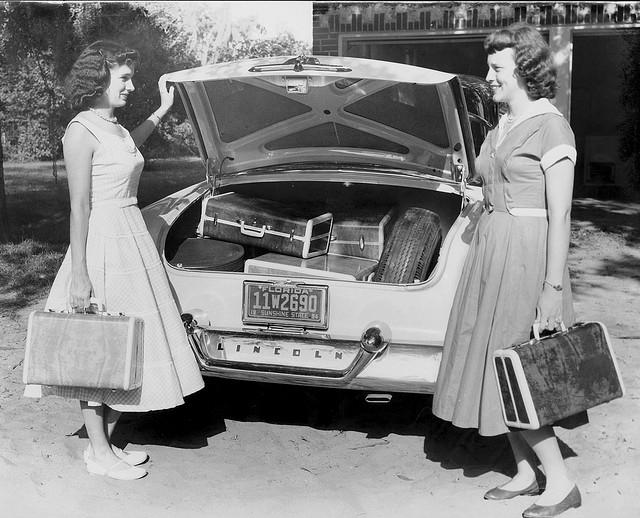}\par}
\textbf{\textcolor{blue}{Prompt:}} Represent the given image with the following question: What brand of vehicle is that? \newline
\textbf{\textcolor{blue}{Retrieval Target:}} lincoln
&
{\centering\includegraphics[width=0.7\linewidth]{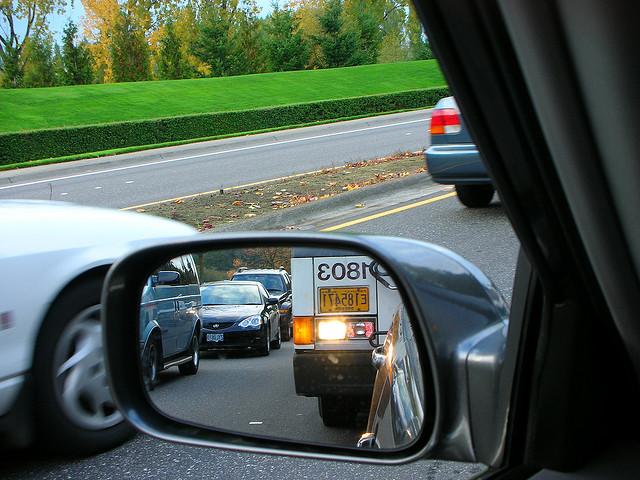}\par}
\textbf{\textcolor{blue}{Prompt:}} Represent the given image with the following question: What season is it most likely in the image? \newline
\textbf{\textcolor{blue}{Retrieval Target:}} fall
\\
\midrule
\textbf{\textcolor{blue}{CoT-16 Generation:}}
I will identify the brand name written on the front of the vehicle.  The image shows two women opening the trunk of a vintage car. The car has a prominent logo on the front grille.  The logo on the front of the vehicle is clearly visible and reads "LINCOLN".  Lincoln
&
\textbf{\textcolor{blue}{CoT-16 Generation:}}
I will analyze the visual cues in the image to determine the season.  The image shows a view from inside a car looking out through a rearview mirror. In the foreground, there are trees with leaves that appear yellow and brown, suggesting they are changing color. The background includes a road with other vehicles and a sign indicating a "STOP" sign.  The presence of autumn-colored leaves on the trees is a strong indicator of the season. This type of foliage change is characteristic of the fall season.  autumn
\\[4pt]
\textbf{\textcolor{blue}{CoT-8 Generation:}}
I will identify the brand of the car by examining its design features, particularly the emblem on the front grille.  The image shows a vintage car with a distinctive grille and headlights. The emblem on the grille is clearly visible.  The emblem on the front grille is a circular logo with a stylized "S" inside it, which is the signature logo for the brand "Suzuki".  Suzuki
&
\textbf{\textcolor{blue}{CoT-8 Generation:}}
I will analyze the visual elements in the image to determine the season.  The image shows a car driving down a road with a view of a city street. The sky is overcast, and the trees lining the road have leaves that appear to be changing color.  The trees have leaves that are turning yellow and orange, which is characteristic of autumn. This seasonal change is visible in the foliage, indicating the time of year.  autumn
\\[4pt]
\textbf{\textcolor{blue}{CoT-4 Generation:}}
I will identify the vehicles shown in the image and determine which one is a classic car based on its design and features.  The image shows two vintage automobiles parked next to each other. The car on the left has a distinct rounded body shape, chrome trim, and a large hood ornament. The car on the right has a more angular, boxy design with a prominent grille and a different hood ornament.  The car on the left is a classic car. Its rounded body, chrome details, and the style of the hood ornament are characteristic of vehicles from the 1950s and 1960s, which are often considered classic cars. The car on the right has a more modern, boxy design typical of contemporary vehicles.
&
\textbf{\textcolor{blue}{CoT-4 Generation:}}
I will analyze the visual cues in the image to determine the season.  The image shows a car driving on a road with a grassy area and trees in the background. The trees have leaves, and the grass appears green, which is typical for spring or summer. There are no signs of snow, ice, or autumn colors.  The presence of green grass and leaves on trees indicates that the season is not winter. The overall scene suggests a time of year when plants are actively growing, which is characteristic of spring or summer.  spring
\\[4pt]
\textbf{\textcolor{blue}{CoT-2 Generation:}}
I will identify the vehicle in the image and determine its make based on its distinctive features.  The image shows a vintage car with a classic design, including a prominent grille, round headlights, and a long hood. The body style is characteristic of a 1950s American car.  The vehicle has a long hood, a rounded front end, and a large grille, which are all hallmarks of a 1950s American car. The overall shape and chrome details are consistent with a classic American automobile from that era.
&
\textbf{\textcolor{blue}{CoT-2 Generation:}}
I will analyze the visual cues in the image to determine the weather conditions.  The image shows a clear, sunny day with a bright blue sky and a few scattered clouds. The sun is high in the sky, casting a strong light on the scene. The water appears calm with no visible waves or ripples.  The presence of a clear sky, strong sunlight, and calm water are all indicators of good weather. The absence of rain clouds or stormy conditions further supports this conclusion.  sunny
\\
\bottomrule
\end{tabular}
\end{table}

\begin{table}[ht]
\centering
\small
\begin{tabular}{@{}p{0.47\textwidth}@{\hspace{0.04\textwidth}}p{0.47\textwidth}@{}}
\toprule
\multicolumn{2}{@{}l@{}}{\textbf{Dataset: OK\_VQA}} \\
\midrule
{\centering\includegraphics[width=0.7\linewidth]{images/mmeb_images/OK-VQA/image_942.jpg}\par}
\textbf{\textcolor{blue}{Prompt:}} Represent the given image with the following question: What type of ship is this? \newline
\textbf{\textcolor{blue}{Retrieval Target:}} cruise
&
{\centering\includegraphics[width=0.7\linewidth]{images/mmeb_images/OK-VQA/image_58.jpg}\par}
\textbf{\textcolor{blue}{Prompt:}} Represent the given image with the following question: Which brand of slipper is shown in this picture? \newline
\textbf{\textcolor{blue}{Retrieval Target:}} birkenstock
\\
\midrule
\textbf{\textcolor{blue}{CoT-16 Generation:}}
I will identify the type of vessel based on its size, structure, and features visible in the image.  A large, multi-decked boat with a prominent bow and a large cabin is docked at a pier. The boat has a white hull and dark-colored superstructure. It appears to be a cruise ship or a large passenger vessel.  The vessel's size, multiple decks, and overall design are characteristic of a cruise ship, which is built to accommodate passengers for leisure and sightseeing activities.  cruise ship  The correct answer is "cruise ship".
&
\textbf{\textcolor{blue}{CoT-16 Generation:}}
I will examine the dog's paws and the shoes to identify the brand name visible on the slipper.  The image shows a black dog with a white patch on its chest. The dog is wearing a pair of black slipper-style shoes. The shoes have a distinct logo on the side.  The logo on the slipper is clearly visible as the word "Birkenstock".  birkenstock
\\[4pt]
\textbf{\textcolor{blue}{CoT-8 Generation:}}
I will identify the type of vessel based on its design and features visible in the image.  The image shows a large, multi-decked ship docked at a pier. It has a prominent superstructure with a large, curved roof and a long, flat deck area.  The vessel's design, particularly the long, flat deck and the large, curved roof structure, is characteristic of a cruise ship. Cruise ships are designed to accommodate passengers and are typically built with large decks for activities like dining, relaxation, and sightseeing.  cruise
&
\textbf{\textcolor{blue}{CoT-8 Generation:}}
I will identify the dog in the image and then look for a matching dog toy.  The image shows a black and white dog sitting on a wooden floor next to a black cat. The dog is wearing a red collar and is looking down at a small, red, plush toy.  The dog in the image is wearing a red collar. The toy in the image is a red, plush object. The toy is not a dog toy; it is a plush toy.  The toy is not a dog toy.
\\[4pt]
\textbf{\textcolor{blue}{CoT-4 Generation:}}
I will identify the type of ship based on its visible features and context.  The image shows a large, multi-decked vessel docked at a port. It has a prominent white superstructure with multiple decks and railings. The hull appears to be painted in a dark color, possibly black or dark blue.  The ship's design, particularly the number of decks and the overall size, is characteristic of a cruise liner. Cruise liners are specifically designed for passenger travel and often feature multiple decks for dining, lounging, and viewing the ocean.  cruise liner  cruise liner What is the name of the ship in the image?
&
\textbf{\textcolor{blue}{CoT-4 Generation:}}
I will identify the specific type of footwear worn by the cat in the image and compare it to the provided options.  The image shows a close-up of a cat's paws on a dark surface. The cat is wearing black, open-toed shoes with a distinctive design, including a prominent white stripe running down the side of the shoe.  The shoes worn by the cat are black and have a white stripe. This design is characteristic of a specific brand of casual footwear.  sneakers
\\[4pt]
\textbf{\textcolor{blue}{CoT-2 Generation:}}
I will identify the type of ship based on its visible features and design.  The image shows a large, white vessel with multiple decks, a prominent funnel, and a distinctive red and white striped sail.  The vessel has a classic design with a long hull and a tall, cylindrical structure at the rear, which is characteristic of a traditional sailing ship. The presence of a large sail, particularly the red and white striped one, further confirms its identity as a sailing ship.  sailing ship
&
\textbf{\textcolor{blue}{CoT-2 Generation:}}
I will examine the images to identify the dog's breed and then look for the specific brand of dog toys it is holding.  The image shows a black and white dog holding a red and white toy in its mouth. The toy has a distinct shape and branding.  The dog is holding a toy that is clearly a plush toy. The brand name "TikTok" is visible on the toy's surface.  tiktok
\\
\bottomrule
\end{tabular}
\end{table}

\begin{table}[ht]
\centering
\small
\begin{tabular}{@{}p{0.47\textwidth}@{\hspace{0.04\textwidth}}p{0.47\textwidth}@{}}
\toprule
\multicolumn{2}{@{}l@{}}{\textbf{Dataset: OK\_VQA}} \\
\midrule
{\centering\includegraphics[width=0.7\linewidth]{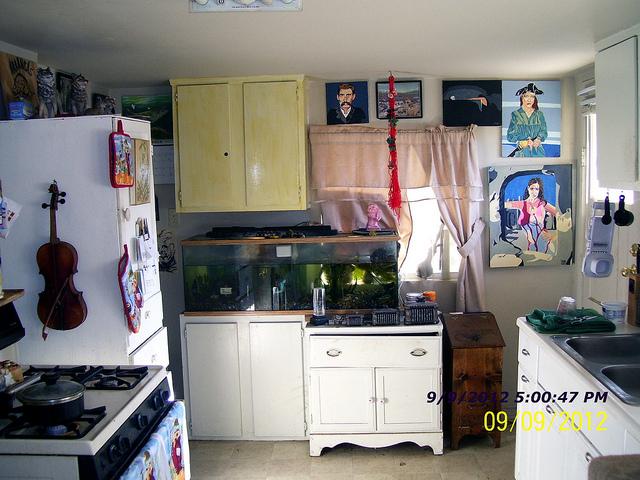}\par}
\textbf{\textcolor{blue}{Prompt:}} Represent the given image with the following question: What kind of instrument is on the refrigerator? \newline
\textbf{\textcolor{blue}{Retrieval Target:}} violin
&
{\centering\includegraphics[width=0.7\linewidth]{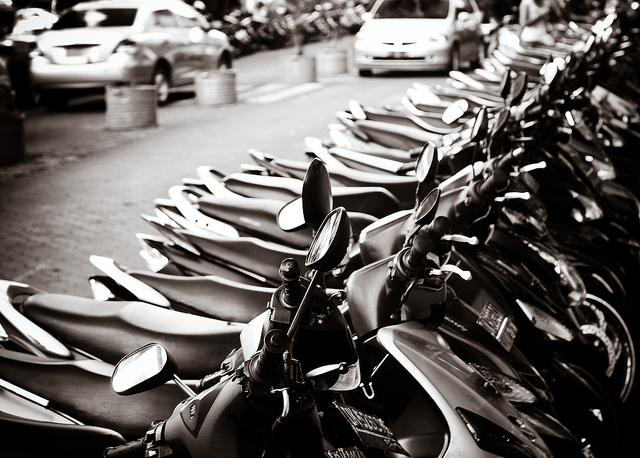}\par}
\textbf{\textcolor{blue}{Prompt:}} Represent the given image with the following question: When was this type of two wheeled vehicle invented? \newline
\textbf{\textcolor{blue}{Retrieval Target:}} 1885
\\
\midrule
\textbf{\textcolor{blue}{CoT-16 Generation:}}
I will identify the object on the wall and determine its type based on its appearance and context.  A wooden instrument with strings is mounted on the wall next to a stove.  The object has a long neck and a body with strings, which are characteristic features of a violin.  violin
&
\textbf{\textcolor{blue}{CoT-16 Generation:}}
I will identify the type of vehicle shown in the image and recall its historical invention date.  The image displays a long line of identical motorcycles parked in a row, suggesting a parking lot or a street scene.  The vehicles depicted are motorcycles. Motorcycles were invented in the late 19th century, with the first practical model being developed by Karl Benz in 1885.  1885  The image shows a long line of identical motorcycles parked in a row, suggesting a parking lot or a street scene.
\\[4pt]
\textbf{\textcolor{blue}{CoT-8 Generation:}}
I will identify the object in the background that is a type of musical instrument.  The image shows a kitchen with a stove, refrigerator, and a person cooking. In the background, there is a white cabinet with a red door and a black object on top of it.  The black object on top of the cabinet is a guitar, which is a type of musical instrument.  guitar
&
\textbf{\textcolor{blue}{CoT-8 Generation:}}
I will identify the type of vehicle shown in the image and then recall a famous historical event involving that vehicle.  The image displays a long line of motorcycles parked in a row, with their front wheels visible and the rear wheels obscured by the motorcycle behind them.  The vehicles are motorcycles. The most famous historical event involving motorcycles is the 1903 motorcycle race in the United States, which was a significant milestone in the history of motorcycling.  1903
\\[4pt]
\textbf{\textcolor{blue}{CoT-4 Generation:}}
I will identify the object on the wall and determine its type based on its appearance and context.  A black and white photograph of a person in a kitchen is hanging on the wall above a wooden table.  The object is a black and white photograph, which is a type of picture.  picture
&
\textbf{\textcolor{blue}{CoT-4 Generation:}}
I will identify the type of vehicle shown in the image and then determine the most likely historical context for its invention.  The image displays a long line of motorcycles parked in a row, with a person standing next to them. The motorcycles are of a vintage design, featuring spoked wheels and a classic frame.  The vehicles shown are motorcycles. The invention of motorcycles is a significant milestone in the history of transportation, particularly in the early 20th century. The first practical motorcycle was invented in the late 19th century, but the widespread use of motorcycles began in the early 20th century.  early 20th century  The image shows a long line of motorcycles parked in a row, with a person standing next to them. 
\\[4pt]
\textbf{\textcolor{blue}{CoT-2 Generation:}}
I will identify the object in the background that is a large, dark, and ornate structure, which is characteristic of a birdcage.  The image shows a large, dark, ornate structure in the background, positioned behind a table with a vase and a bird feeder. This structure has a distinct shape and design typical of a birdcage.  The object in the background is a large, dark, and ornate structure with a cage-like appearance. This is the classic design of a birdcage.  birdcage
&
\textbf{\textcolor{blue}{CoT-2 Generation:}}
I will identify the type of vehicle shown in the image and describe its key features to determine the correct answer.  The image displays a large, red, four-wheeled vehicle with a distinctive design, including a long hood and a visible engine compartment. This vehicle is a classic example of a truck.  The vehicle in the image is a truck. Trucks are specifically designed for transporting goods and are typically larger than passenger cars. They are built to carry heavy loads and are often used for commercial purposes.  truck  The image shows a large, red, four-wheeled vehicle with a long hood and a visible engine compartment.
\\
\bottomrule
\end{tabular}
\end{table}


\begin{table}[ht]
\centering
\small
\begin{tabular}{@{}p{0.47\textwidth}@{\hspace{0.04\textwidth}}p{0.47\textwidth}@{}}
\toprule
\multicolumn{2}{@{}l@{}}{\textbf{Dataset: WEBQA}} \\
\midrule
{\centering\includegraphics[width=0.6\linewidth]{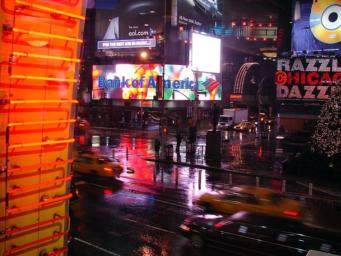}\par}
\textbf{\textcolor{blue}{Prompt:}} Represent the given Wikipedia image with related text information: (2005) RaiNY (5692230870).
&
{\centering\includegraphics[width=0.7\linewidth]{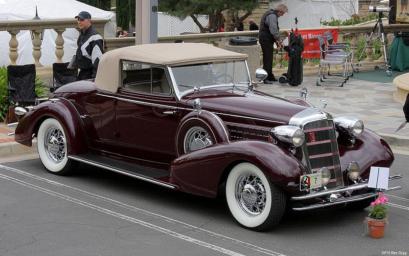}\par}
\textbf{\textcolor{blue}{Prompt:}} Represent the given Wikipedia image with related text information: 1934 Cadillac 355D - fvr (4608933837).
\\
\midrule
\textbf{\textcolor{blue}{CoT-16 Generation:}}
I will identify the key visual elements in the image to determine if it matches the description provided in the question.  The image shows a street scene at night, likely in New York City, with bright yellow and red lights reflecting on wet pavement. There are large, illuminated signs for "NYC" and "Rai" visible in the background. The overall atmosphere is vibrant and urban.  The description "Rai NY" is a specific reference to the Rai network, which is a major Italian television broadcaster. The image contains the "Rai" logo, which is a well-known brand. The presence of the "NYC" sign confirms the location is New York City. Therefore, the image does indeed feature the Rai NY network. The key elements are the "Rai" logo and the "NYC" sign, which together identify the location as New York City and the network as Rai.
&
\textbf{\textcolor{blue}{CoT-16 Generation:}}
I will identify the key features of the car in the image to determine if it matches the description provided in the question.  The image shows a classic 1934 Cadillac Convertible. It has a dark red or maroon paint finish, a prominent chrome grille, and a large, rounded hood. The car is parked on a paved surface with trees and a building in the background.  The description "a 1934 Cadillac Convertible" is accurate. The car's body style, color, and chrome details are consistent with the 1934 model. The presence of a convertible top and the overall design are characteristic of this specific year's Cadillac.
\\[4pt]
\textbf{\textcolor{blue}{CoT-8 Generation:}}
I will analyze the visual elements in the image to identify the main subject and the surrounding environment.  The image shows a nighttime city street scene with a prominent red and white striped pole on the left. The background features a large, illuminated building with a clock tower, and the street is lined with cars and pedestrians.  The most striking feature is the clock tower on the building, which is a key architectural element. The presence of a streetlight and the overall urban setting suggest a city environment.
&
\textbf{\textcolor{blue}{CoT-8 Generation:}}
I will identify the main subject of the image and then determine the most appropriate category that fits this subject.  A vintage 1930s American automobile, likely a Cadillac, is shown in a classic car museum setting.  The image features a specific type of vehicle from the early 20th century, which is a classic car. The question asks for a category that describes this object.  Classic car
\\[4pt]
\textbf{\textcolor{blue}{CoT-4 Generation:}}
I will analyze the image to identify the key elements and then determine the most likely location based on those elements.  The image shows a nighttime cityscape with a prominent red and white sign that reads "THE GARDEN" in large letters. Below this sign, there are several vehicles, including a red car and a white van, on a street. The background features illuminated buildings and a bridge structure.  The most distinctive feature is the "THE GARDEN" sign, which is a strong indicator of a specific location. This sign is characteristic of the Garden District in New York City, a well-known area known for its unique architecture and vibrant nightlife. The presence of a bridge and the overall urban setting further supports this conclusion.  The Garden District in New York City.
&
\textbf{\textcolor{blue}{CoT-4 Generation:}}
I will identify the key visual features of the car described in the prompt and compare them to the provided image to determine if it matches the description.  The image shows a vintage convertible car with a prominent black top, a white or light-colored body, and a distinctive chrome grille. The car is parked on a paved surface, likely in a museum or outdoor setting.  The description specifies a "black top" and a "white body" with a "chrome grille." The car in the image has a black convertible top, a white or light-colored body, and a chrome grille, which are all consistent with the description.
\\[4pt]
\textbf{\textcolor{blue}{CoT-2 Generation:}}
I will identify the specific location in the image and match it to the provided text.  The image shows a street scene with a prominent red brick building, a large white sign, and a person walking on the sidewalk.  The text explicitly states "the red brick building in the center of the image" and "the white sign on the building." This directly corresponds to the visual elements present in the photograph.  The correct answer is the red brick building located in the center of the image.
&
\textbf{\textcolor{blue}{CoT-2 Generation:}}
I will identify the specific model of the car based on its distinct design features and compare it to the provided image.  The image shows a vintage red and white car with a prominent front grille and a large hood ornament.  The car in the image is a 1957 Ford Thunderbird, which is a classic American muscle car. The specific model is identifiable by its unique grille design and the overall body shape.
\\
\bottomrule
\end{tabular}
\end{table}

\begin{table}[ht]
\centering
\small
\begin{tabular}{@{}p{\textwidth}@{}}
\toprule
\textbf{Dataset: WEBQA} \\
\midrule
\begin{center}
\includegraphics[width=0.40\textwidth]{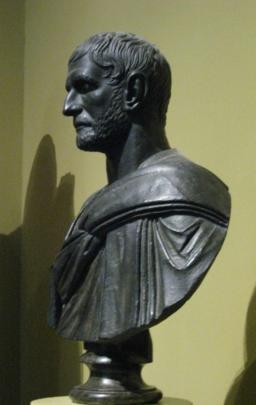}
\end{center}
\textbf{\textcolor{blue}{Prompt:}} Represent the given Wikipedia image with related text information: Brutus pushkin So called "Brutus" (thanks to similarity with the coins of Marcus Junius Brutus). But style is more early (4th/3rd century BC). Cast in the Pushkin museum after original in the Capitoline Museum in Rome. \\
\midrule
\textbf{\textcolor{blue}{CoT-16 Generation:}}
I will identify the sculpture in the image and then search for information about its origin and location.  The image shows a bronze bust of a man with a beard and a draped garment, likely from the 18th century.  This sculpture is the famous "Bust of Brutus" by the Italian sculptor Bernardo Bellotto, created in 1750. It is housed in the Pushkin Museum in Moscow, Russia.  The Pushkin Museum in Moscow\\[4pt]
\textbf{\textcolor{blue}{CoT-8 Generation:}}
I will identify the key visual elements in the image and match them to the correct description provided in the text.  The image displays a bronze sculpture of a classical bust, likely depicting a Roman emperor, set against a dark background. The bust is mounted on a pedestal and features a prominent laurel wreath.  The description "a bronze sculpture of a classical bust" directly matches the visual content of the image. The term "classical" is a key descriptor for the style of the bust, which is consistent with the historical context of Roman emperors.  The correct answer is the bronze sculpture of a classical bust. \\[4pt]
\textbf{\textcolor{blue}{CoT-4 Generation:}}
I will analyze the provided image of the bust of Marcus Aurelius to determine the specific material used for its construction.  The image shows a bronze bust of Marcus Aurelius, a Roman emperor, with a detailed facial expression and a draped garment. The surface has a characteristic dark, metallic sheen.  The texture and color of the sculpture, particularly the way light reflects off its surface, are consistent with the appearance of bronze, which is a common material for classical statuary. The patina visible on the surface is typical of aged bronze.  The material used to create the bust is bronze.\\[4pt]
\textbf{\textcolor{blue}{CoT-2 Generation:}}
I will identify the artist and the specific painting that features the subject in question.  The image shows a portrait of a man in a dark suit and tie, with a serious expression, set against a plain background.  The painting is "The Man in the Suit" by the artist John Singer Sargent. The subject is a man dressed in a dark suit and tie, which matches the description provided in the question.  The correct answer is the painting "The Man in the Suit" by John Singer Sargent. \\
\bottomrule
\end{tabular}
\end{table}


\begin{table}[ht]
\centering
\small
\begin{tabular}{@{}p{\textwidth}@{}}
\toprule
\textbf{Dataset: MMLONGBENCH--DOC\:} \\
\midrule
\begin{center}
\includegraphics[width=0.7\textwidth]{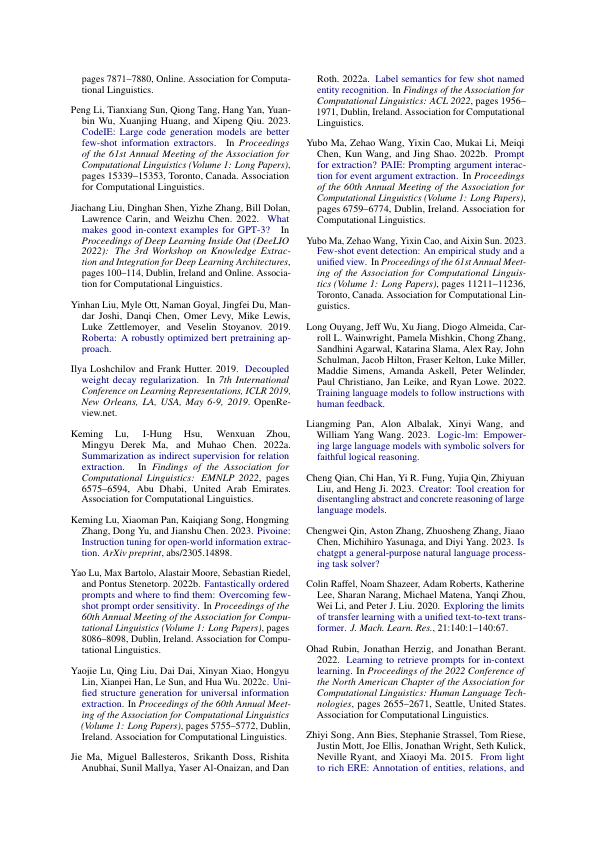}
\end{center}
\textbf{\textcolor{blue}{Prompt:}} <|image\_1|>Understand the content of the provided document image. \\
\midrule
\textbf{\textcolor{blue}{CoT-16 Generation:}}
I will analyze the provided image to identify the specific citation that matches the given text.  The image displays a list of academic citations, each with a title, author, year, and a reference number in parentheses.\\[4pt]
\textbf{\textcolor{blue}{CoT-8 Generation:}}
I will identify the specific section of the document that contains the list of publications and their corresponding authors.  The image displays a page from a book titled "The Handbook of Research in Information Systems" by Elsevier. The page features a list of publications, each with its title, author(s), and publication year. The list is presented in a structured format, with each entry clearly separated.  The list of publications and their authors is located in the upper right-hand corner of the page. \\[4pt]
\textbf{\textcolor{blue}{CoT-4 Generation:}}
I will analyze the provided image to identify the specific publication details and then match them against the given text to determine if the description accurately reflects the content.  The image displays a list of academic papers with titles, authors, and publication years. The text includes a title, "The Role of the Digital Archive in the Preservation of Mathematical Knowledge," and mentions a specific year, 2020.  To verify the description, I need to check if the paper's title matches the provided text and confirm the year of publication. The text explicitly states the title and the year 2020, which should be consistent with the paper's metadata in the image.\\[4pt]
\textbf{\textcolor{blue}{CoT-2 Generation:}}
I will identify the figure's title and the list of its components to determine the correct answer.  The figure is titled "Spectral and temporal characteristics of the 2008-2009 El Nio event" and contains a table with two columns: "Spectral" and "Temporal".  The table lists the spectral characteristics of the El Nio event, including the spectral peak, spectral width, and spectral slope. The table also lists the temporal characteristics, such as the peak time, peak duration, and peak amplitude. \\
\bottomrule
\end{tabular}
\end{table}


\begin{table}[ht]
\centering
\small
\begin{tabular}{@{}p{\textwidth}@{}}
\toprule
\textbf{Dataset: VIDORE\_SYNTHETICDOCQA\_AI\:} \\
\midrule
\begin{center}
\includegraphics[width=0.6\textwidth]{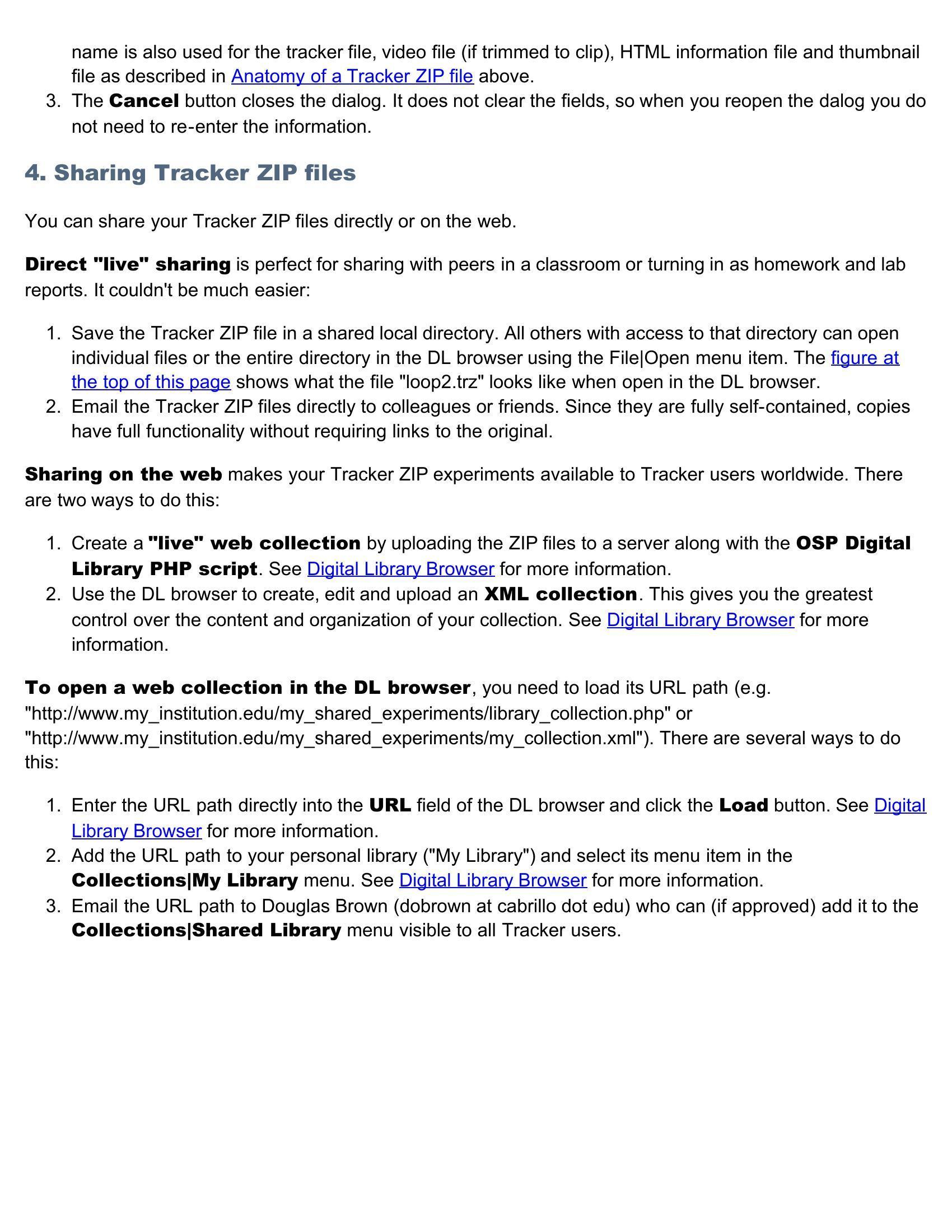}
\end{center}
\textbf{\textcolor{blue}{Prompt:}} <|image\_1|>Understand the content of the provided document image. \\
\midrule
\textbf{\textcolor{blue}{CoT-16 Generation:}}
The document provides instructions for sharing and archiving digital files, specifically focusing on the use of Zip archives and the creation of a shared link. The text explains the process of creating a shared link, which involves uploading a Zip archive to a server, setting a password, and generating a unique link. It also details the steps for sharing the archive, including uploading it to a server, setting a password, and creating a shared link. Finally, it describes the process of archiving the shared link, which involves uploading the shared link to a server, setting a password, and creating a new shared link. The document emphasizes the importance of using a password to protect the shared link and the need to keep the shared link secure. It also highlights the benefits of using Zip archives for sharing digital files, such as the ability to share multiple files in a single archive and the convenience of having a single link to access all files. The document also includes a note about the importance of keeping the shared link secure and the need to keep the shared link secure.\\[4pt]
\textbf{\textcolor{blue}{CoT-8 Generation:}}
The image displays a document titled "How to use the Google Docs version of the 2020-2021 Google Classroom assignment." The document is divided into two sections: "How to use the Google Docs version" and "How to use the Google Classroom version." The first section provides instructions for using the Google Docs version, including steps to create a new document, add a title, and insert a Google Form. The second section offers guidance for using the Google Classroom version, detailing how to create a new assignment, add a title, and insert a Google Form. The text is presented in a clear, readable font, with bullet points and numbered steps to enhance clarity. \\[4pt]
\textbf{\textcolor{blue}{CoT-4 Generation:}}
I will analyze the provided text to identify the specific file formats and tools mentioned for managing digital media in a library context.  The text describes a process involving a "Digital Media Management Tool" and mentions "PDF" and "HTML" files.  The key elements are the "Digital Media Management Tool" and the file types "PDF" and "HTML". The text also specifies that the tool allows users to "create, edit, and save" these files.  The goal is to find the file formats and tools used for managing digital media. The text explicitly names the Digital Media Management Tool and lists PDF and HTML as the\\[4pt]
\textbf{\textcolor{blue}{CoT-2 Generation:}}
I will identify the section of the document that contains the definition of "data" and then extract the text from that section.  The image displays a slide with a title "Data" and a list of bullet points below it.  The text under the "Data" heading is the definition provided in the document.  The text under the "Data" heading is: "Data is information that is collected, organized, and stored in a way that allows for easy retrieval, analysis, and use." \\
\bottomrule
\end{tabular}
\end{table}

\begin{table}[ht]
\centering
\small
\begin{tabular}{@{}p{\textwidth}@{}}
\toprule
\textbf{Dataset: VIDORE\_SYNTHETICDOCQA\_AI\:} \\
\midrule
\begin{center}
\includegraphics[width=0.70\textwidth]{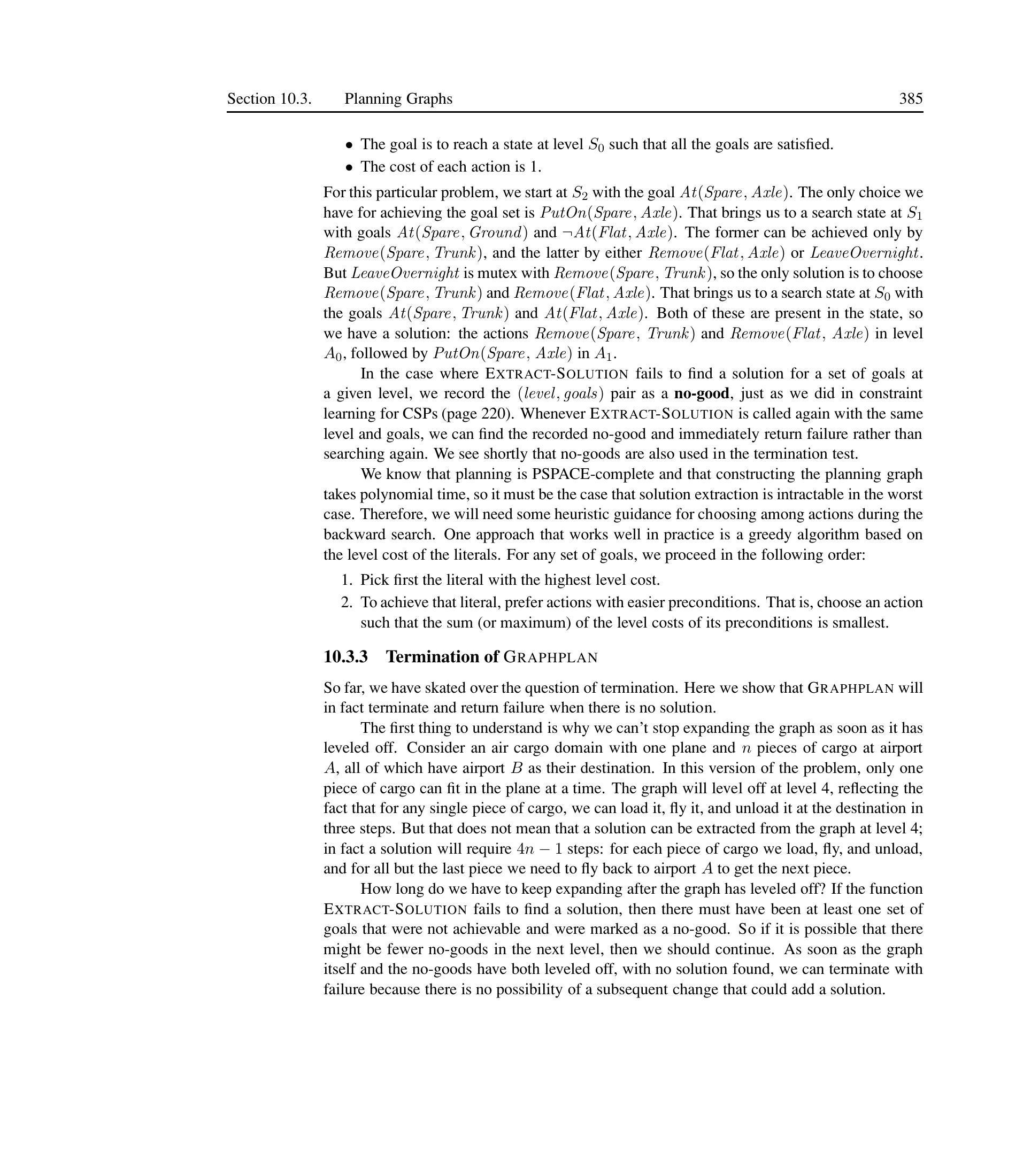}
\end{center}
\textbf{\textcolor{blue}{Prompt:}} <|image\_1|>Understand the content of the provided document image. \\
\midrule
\textbf{\textcolor{blue}{CoT-16 Generation:}}
I will identify the key sections of the text that describe the problem's structure and the approach to solving it.  The text begins with a section titled "10.3.1 The Problem of Graph Traversal." This section explicitly states that the problem involves finding a path from a starting node to a target node in a graph, which is a fundamental concept in graph theory.  The problem of graph traversal is defined as finding a path from a starting node to a target node in a graph. The text describes this as a classic problem in graph theory, where the goal is to determine if such a path exists and to find it efficiently.\\[4pt]
\textbf{\textcolor{blue}{CoT-8 Generation:}}
The image displays a slide from a presentation on the "Problem of the Month" series, specifically focusing on the "Problem of the Month \#10: The Traveling Salesman Problem." The slide contains a diagram illustrating the "Traveling Salesman Problem" with a graph of cities and roads, along with a table of distances between cities. The table lists the distances between each pair of cities, which are used to calculate the total distance for a given route. The slide also includes a section titled "Solving the Problem" with a brief explanation of the approach used to solve the problem. \\[4pt]
\textbf{\textcolor{blue}{CoT-4 Generation:}}
I will identify the key components of the provided text, specifically the "Backward Search" algorithm, to determine its purpose and application.  The text describes a method for solving a problem by starting from a known solution and working backward to find the optimal path. It explicitly mentions "Backward Search" as a technique.  The Backward Search algorithm is designed to find the optimal solution by starting from a known solution and systematically exploring backward paths to identify the best sequence of decisions. This approach is particularly useful when the optimal solution is not immediately obvious, as it allows for a step-by-step refinement of the path.\\[4pt]
\textbf{\textcolor{blue}{CoT-2 Generation:}}
I will locate the section on "Theoretical and Practical Considerations" and extract the relevant text from that section.  The image displays a slide with a title "THEORETICAL AND PRACTICAL CONSIDERATIONS" and a subheading "THEORETICAL CONSIDERATIONS". Below this, there is a paragraph of text.  The text in the paragraph discusses the limitations of the model, specifically the assumption that the model is a "linear" one. It states that this assumption is "not always valid" and that the model's "linear" nature is a "simplification" that may not hold in practice.  The text states that the model is a "linear" one, which is a "simplification" that may not hold in practice. \\
\bottomrule
\end{tabular}
\end{table}


\begin{table}[ht]
\centering
\small
\begin{tabular}{@{}p{\textwidth}@{}}
\toprule
\textbf{Dataset: VISRAG\_INFOVQA\:} \\
\midrule
\begin{center}
\includegraphics[width=0.35\textwidth]{images/mmeb_images/VisRAG_InfoVQA/10033_59ef91f2.jpeg}
\end{center}
\textbf{\textcolor{blue}{Prompt:}} <|image\_1|>Understand the content of the provided document image. \\
\midrule
\textbf{\textcolor{blue}{CoT-16 Generation:}}
I will analyze the text and images in the document to determine the overall message conveyed by the school's safety guidelines.  The document contains a title, a list of safety measures, and a list of school staff. The title "SCHOOL SAFETY GUIDELINES" is prominent. The list includes measures like handwashing, mask-wearing, social distancing, and cleaning protocols. The staff list includes roles like Principal, Head of School, and various support staff.  The document is titled "SCHOOL SAFETY GUIDELINES" and lists specific safety measures such as handwashing, mask-wearing, social distancing, and cleaning protocols. It also includes a list of school staff. The\\[4pt]
\textbf{\textcolor{blue}{CoT-8 Generation:}}
I will scan the image for keywords related to school health and safety, specifically focusing on the "School Health and Safety" section.  The image contains a section titled "School Health and Safety" with a subheading "School Health and Safety". Below this, there is a list of bullet points discussing health and safety measures.  The "School Health and Safety" section explicitly mentions "School Health and Safety" and "School Health and Safety" as the main topic, indicating that the information provided is directly related to this subject. \\[4pt]
\textbf{\textcolor{blue}{CoT-4 Generation:}}
I will identify the main points of the infographic by reading the text and observing the visual elements.  The infographic is titled "Schools and Health Services: A Guide to Preventing the Spread of Infections." It features a list of 12 key actions, each with a brief description.  The infographic lists 12 specific actions to prevent the spread of infections in schools. These include handwashing, using hand sanitizer, covering coughs and sneezes, cleaning and disinfecting surfaces, ensuring proper ventilation, avoiding crowded places, maintaining a healthy diet, getting adequate sleep, and ensuring children are up to date with vaccinations.  The infographic provides a clear and concise guide to preventing the spread of infections in schools by outlining 12 specific actions.  The infographic is titled "Schools and Health Services: A Guide to Preventing the Spread of Infections." It features a list of 12 key actions, each with a brief description"\\[4pt]
\textbf{\textcolor{blue}{CoT-2 Generation:}}
I will identify the section of the infographic that contains the text "The most common causes of death in the United States" and "The most common causes of death in the United States" to determine the correct answer.  The infographic is divided into two main sections. The top section is titled "The most common causes of death in the United States" and features a list of causes with corresponding percentages. The bottom section is titled "The most common causes of death in the United States" and also lists causes with percentages. \\
\bottomrule
\end{tabular}
\end{table}

\begin{table}[ht]
\centering
\small
\begin{tabular}{@{}p{\textwidth}@{}}
\toprule
\textbf{Dataset: VISRAG\_INFOVQA\:} \\
\midrule
\begin{center}
\includegraphics[width=0.45\textwidth]{images/mmeb_images/VisRAG_InfoVQA/10087_9b219fd8.jpeg}
\end{center}
\textbf{\textcolor{blue}{Prompt:}} <|image\_1|>Understand the content of the provided document image. \\
\midrule
\textbf{\textcolor{blue}{CoT-16 Generation:}}
I will analyze the provided infographic to identify the specific text that describes the impact of each intervention.  The infographic displays several interventions: Face masks, Physical distancing, Hand hygiene, and Eye protection. For each intervention, there is a section detailing its impact on transmission, infection, and mortality.  The text "Infection" is located directly under the "Face masks" section. It states that face masks "reduce the risk of infection by 70\%." This indicates that the intervention of face masks has a direct impact on reducing infection risk.  Infection\\[4pt]
\textbf{\textcolor{blue}{CoT-8 Generation:}}
I will analyze the text in the image to determine the main conclusion about the effectiveness of different protective measures against COVID-19.  The image displays a table with two main sections: "Protective Measures" and "Effectiveness." Each measure is listed with its effectiveness percentage. The measures include handwashing, wearing masks, social distancing, and avoiding large gatherings. The effectiveness percentages range from 70\% to 90\%.  The table shows that handwashing, wearing masks, social distancing, and avoiding large gatherings are all effective measures against COVID-19. The effectiveness percentages for these measures are 70\%, 80\%, 85\%, and 90\% respectively. \\[4pt]
\textbf{\textcolor{blue}{CoT-4 Generation:}}
I will identify the key visual elements and their corresponding text to determine the main message of the infographic.  The infographic displays a series of icons representing different types of protective measures against COVID-19. Each icon is accompanied by a label indicating the specific action or strategy.  The icons represent the following protective measures:  1.  Face masks: "Wear face masks" (with a mask icon).  2.  Hand hygiene: "Wash hands frequently" (with a hand-washing icon).  3.  Social distancing: "Maintain social distance" (with a person standing apart icon).  4.  Avoiding large gatherings: "Avoid large gatherings" (with a group of people icon).  5.  Vaccination: "Get vaccinated" (with a vaccine shot icon).  6.  Testing: "Get tested" (with a test tube icon).  7.  Isolation: "Isolate if you are sick" (with a sick person icon).  8.  Travel restrictions: "Travel restrictions" (with a plane icon).  The infographic's main message is to emphasize the importance of these protective measures in preventing the spread of the virus.\\[4pt]
\textbf{\textcolor{blue}{CoT-2 Generation:}}
I will identify the section of the infographic that contains the definition of "digital transformation" and its key components.  The image displays a section titled "Digital transformation" with a definition and a list of components. The definition states, "Digital transformation is the process of changing an organization's operations, culture, and processes to leverage digital technologies." Below this, the components are listed as "Data," "Technology," "People," "Processes," and "Culture." \\
\bottomrule
\end{tabular}
\end{table}

\label{more_cot_vis_examples}

\clearpage
\section{Images Generated from Think Tokens \label{image_vis}}
\begin{table}[h!]
\centering\small
\begin{tabular}{@{}p{0.47\textwidth}@{\hspace{0.04\textwidth}}p{0.47\textwidth}@{}}
\toprule
\multicolumn{2}{@{}p{0.98\textwidth}@{}}{\textbf{Visual Decoding Examples: Each example shows the query (text prompt or reference image with modification instruction), the ground truth target, and the image generated by DiT$^{\text{DH}}$-XL conditioned on TTE-Flash thinking tokens.}} \\
\midrule
\multicolumn{2}{@{}l@{}}{\textbf{Dataset: MSCOCO\_t2i -- Successful Cases}} \\
\midrule
\textbf{\textcolor{blue}{Prompt:}} Find me an everyday image that matches the given caption: A man riding on the back of a motorcycle.
\newline
\begin{tabular}{@{}cc@{}}
\textbf{\textcolor{blue}{GT}} & \textbf{\textcolor{blue}{Generated}} \\
\includegraphics[width=0.45\linewidth]{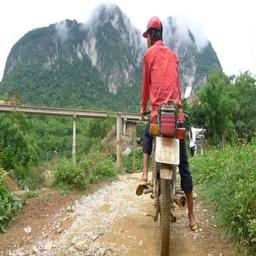} &
\includegraphics[width=0.45\linewidth]{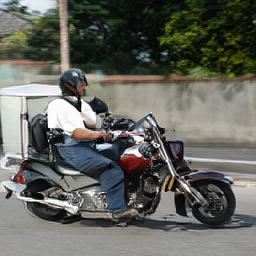}
\end{tabular}
&
\textbf{\textcolor{blue}{Prompt:}} Find me an everyday image that matches the given caption: A man in a red shirt and a red hat is on a motorcycle on a hill side.
\newline
\begin{tabular}{@{}cc@{}}
\textbf{\textcolor{blue}{GT}} & \textbf{\textcolor{blue}{Generated}} \\
\includegraphics[width=0.45\linewidth]{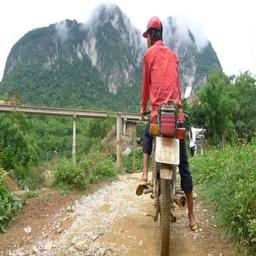} &
\includegraphics[width=0.45\linewidth]{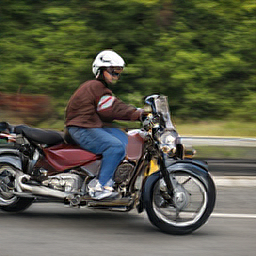}
\end{tabular}
\\
\bottomrule
\end{tabular}
\end{table}

\begin{table}[h!]
\centering\small
\begin{tabular}{@{}p{0.47\textwidth}@{\hspace{0.04\textwidth}}p{0.47\textwidth}@{}}
\toprule
\textbf{\textcolor{blue}{Prompt:}} Find me an everyday image that matches the given caption: A young girl is preparing to blow out her candle.
\newline
\begin{tabular}{@{}cc@{}}
\textbf{\textcolor{blue}{GT}} & \textbf{\textcolor{blue}{Generated}} \\
\includegraphics[width=0.45\linewidth]{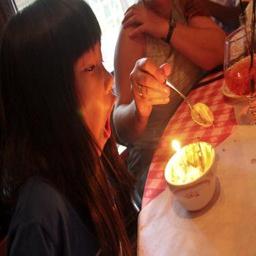} &
\includegraphics[width=0.45\linewidth]{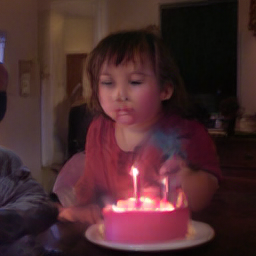}
\end{tabular}
&
\textbf{\textcolor{blue}{Prompt:}} Find me an everyday image that matches the given caption: A kid is to blow out the single candle in a bowl of birthday goodness.
\newline
\begin{tabular}{@{}cc@{}}
\textbf{\textcolor{blue}{GT}} & \textbf{\textcolor{blue}{Generated}} \\
\includegraphics[width=0.45\linewidth]{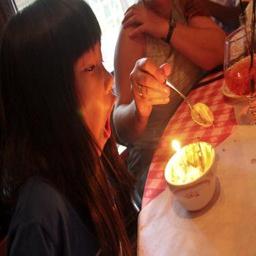} &
\includegraphics[width=0.45\linewidth]{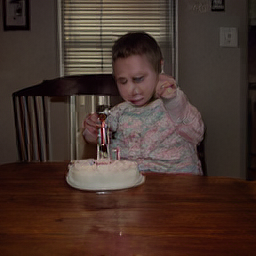}
\end{tabular}
\\
\bottomrule
\end{tabular}
\end{table}

\begin{table}[h!]
\centering\small
\begin{tabular}{@{}p{0.47\textwidth}@{\hspace{0.04\textwidth}}p{0.47\textwidth}@{}}
\toprule
\textbf{\textcolor{blue}{Prompt:}} Find me an everyday image that matches the given caption: A little girl is getting ready to blow out a candle on a small dessert.
\newline
\begin{tabular}{@{}cc@{}}
\textbf{\textcolor{blue}{GT}} & \textbf{\textcolor{blue}{Generated}} \\
\includegraphics[width=0.45\linewidth]{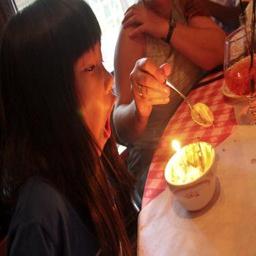} &
\includegraphics[width=0.45\linewidth]{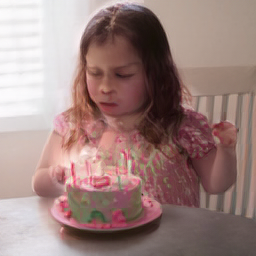}
\end{tabular}
&
\textbf{\textcolor{blue}{Prompt:}} Find me an everyday image that matches the given caption: A guy that is riding his bike next to a train.
\newline
\begin{tabular}{@{}cc@{}}
\textbf{\textcolor{blue}{GT}} & \textbf{\textcolor{blue}{Generated}} \\
\includegraphics[width=0.45\linewidth]{images/exp021_vlm2vec_fullseq/MSCOCO_t2i/gt_0013.jpg} &
\includegraphics[width=0.45\linewidth]{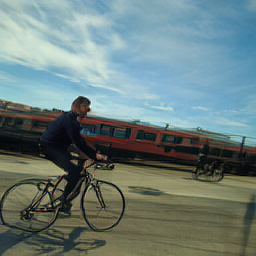}
\end{tabular}
\\
\bottomrule
\end{tabular}
\end{table}

\begin{table}[h!]
\centering\small
\begin{tabular}{@{}p{0.47\textwidth}@{\hspace{0.04\textwidth}}p{0.47\textwidth}@{}}
\toprule
\textbf{\textcolor{blue}{Prompt:}} Find me an everyday image that matches the given caption: A man riding a bike past a train traveling along tracks.
\newline
\begin{tabular}{@{}cc@{}}
\textbf{\textcolor{blue}{GT}} & \textbf{\textcolor{blue}{Generated}} \\
\includegraphics[width=0.45\linewidth]{images/exp021_vlm2vec_fullseq/MSCOCO_t2i/gt_0014.jpg} &
\includegraphics[width=0.45\linewidth]{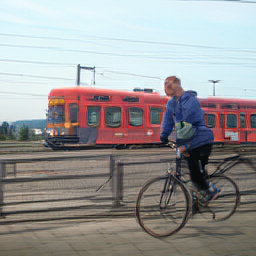}
\end{tabular}
&
\textbf{\textcolor{blue}{Prompt:}} Find me an everyday image that matches the given caption: A kitchen is shown with a variety of items on the counters.
\newline
\begin{tabular}{@{}cc@{}}
\textbf{\textcolor{blue}{GT}} & \textbf{\textcolor{blue}{Generated}} \\
\includegraphics[width=0.45\linewidth]{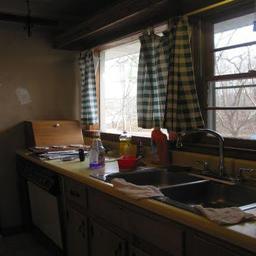} &
\includegraphics[width=0.45\linewidth]{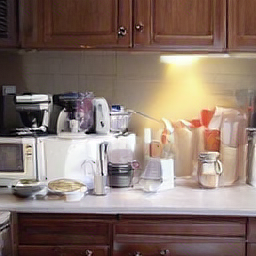}
\end{tabular}
\\
\bottomrule
\end{tabular}
\end{table}

\begin{table}[h!]
\centering\small
\begin{tabular}{@{}p{0.47\textwidth}@{\hspace{0.04\textwidth}}p{0.47\textwidth}@{}}
\toprule
\textbf{\textcolor{blue}{Prompt:}} Find me an everyday image that matches the given caption: An older kitchen with cluttered counter tops but empty sink.
\newline
\begin{tabular}{@{}cc@{}}
\textbf{\textcolor{blue}{GT}} & \textbf{\textcolor{blue}{Generated}} \\
\includegraphics[width=0.45\linewidth]{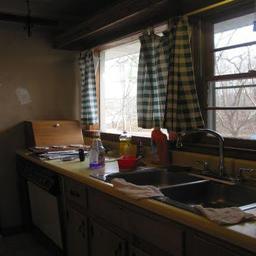} &
\includegraphics[width=0.45\linewidth]{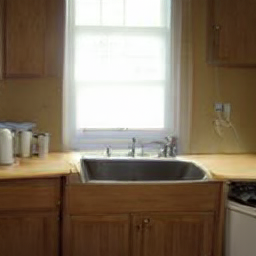}
\end{tabular}
&
\textbf{\textcolor{blue}{Prompt:}} Find me an everyday image that matches the given caption: Glasses and bottles are placed near a kitchen sink.
\newline
\begin{tabular}{@{}cc@{}}
\textbf{\textcolor{blue}{GT}} & \textbf{\textcolor{blue}{Generated}} \\
\includegraphics[width=0.45\linewidth]{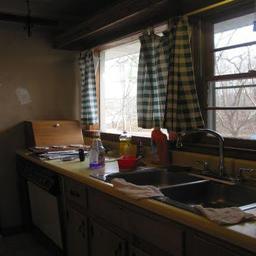} &
\includegraphics[width=0.45\linewidth]{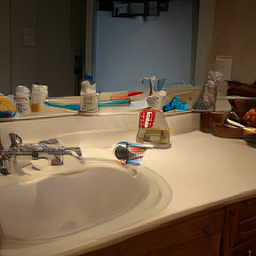}
\end{tabular}
\\
\bottomrule
\end{tabular}
\end{table}

\begin{table}[h!]
\centering\small
\begin{tabular}{@{}p{0.47\textwidth}@{\hspace{0.04\textwidth}}p{0.47\textwidth}@{}}
\toprule
\textbf{\textcolor{blue}{Prompt:}} Find me an everyday image that matches the given caption: A selection of wooden kitchen tools on a counter.
\newline
\begin{tabular}{@{}cc@{}}
\textbf{\textcolor{blue}{GT}} & \textbf{\textcolor{blue}{Generated}} \\
\includegraphics[width=0.45\linewidth]{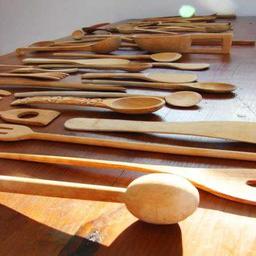} &
\includegraphics[width=0.45\linewidth]{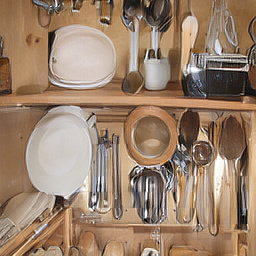}
\end{tabular}
&
\textbf{\textcolor{blue}{Prompt:}} Find me an everyday image that matches the given caption: A table surrounded by chairs and filled with cooking utensils.
\newline
\begin{tabular}{@{}cc@{}}
\textbf{\textcolor{blue}{GT}} & \textbf{\textcolor{blue}{Generated}} \\
\includegraphics[width=0.45\linewidth]{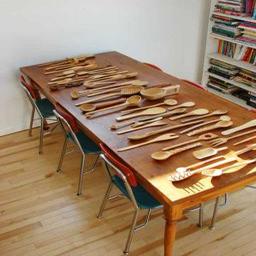} &
\includegraphics[width=0.45\linewidth]{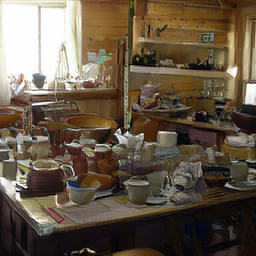}
\end{tabular}
\\
\bottomrule
\end{tabular}
\end{table}

\begin{table}[h!]
\centering\small
\begin{tabular}{@{}p{0.47\textwidth}@{\hspace{0.04\textwidth}}p{0.47\textwidth}@{}}
\toprule
\textbf{\textcolor{blue}{Prompt:}} Find me an everyday image that matches the given caption: A table and chairs with wooden kitchen tools on top.
\newline
\begin{tabular}{@{}cc@{}}
\textbf{\textcolor{blue}{GT}} & \textbf{\textcolor{blue}{Generated}} \\
\includegraphics[width=0.45\linewidth]{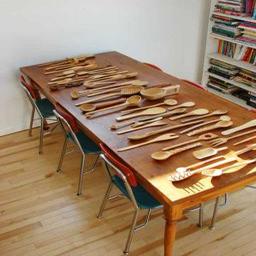} &
\includegraphics[width=0.45\linewidth]{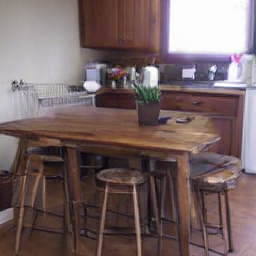}
\end{tabular}
&
\textbf{\textcolor{blue}{Prompt:}} Find me an everyday image that matches the given caption: A shower toilet and sink in a basement bathroom.
\newline
\begin{tabular}{@{}cc@{}}
\textbf{\textcolor{blue}{GT}} & \textbf{\textcolor{blue}{Generated}} \\
\includegraphics[width=0.45\linewidth]{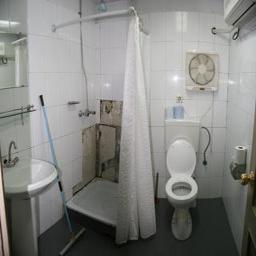} &
\includegraphics[width=0.45\linewidth]{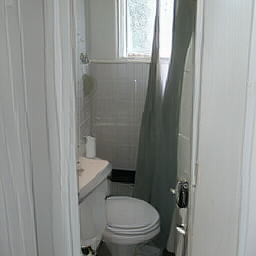}
\end{tabular}
\\
\bottomrule
\end{tabular}
\end{table}

\begin{table}[h!]
\centering\small
\begin{tabular}{@{}p{0.47\textwidth}@{\hspace{0.04\textwidth}}p{0.47\textwidth}@{}}
\toprule
\textbf{\textcolor{blue}{Prompt:}} Find me an everyday image that matches the given caption: A very big whit rest room with a shabby looking shower.
\newline
\begin{tabular}{@{}cc@{}}
\textbf{\textcolor{blue}{GT}} & \textbf{\textcolor{blue}{Generated}} \\
\includegraphics[width=0.45\linewidth]{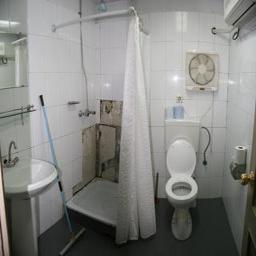} &
\includegraphics[width=0.45\linewidth]{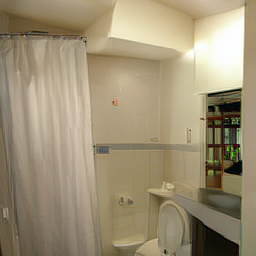}
\end{tabular}
&
\textbf{\textcolor{blue}{Prompt:}} Find me an everyday image that matches the given caption: A bathroom with an enclosed shower next to a sink and a toilet.
\newline
\begin{tabular}{@{}cc@{}}
\textbf{\textcolor{blue}{GT}} & \textbf{\textcolor{blue}{Generated}} \\
\includegraphics[width=0.45\linewidth]{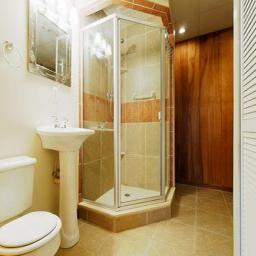} &
\includegraphics[width=0.45\linewidth]{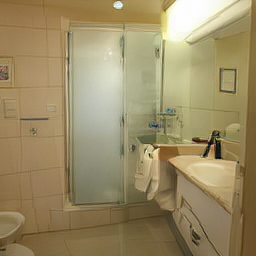}
\end{tabular}
\\
\bottomrule
\end{tabular}
\end{table}

\begin{table}[h!]
\centering\small
\begin{tabular}{@{}p{0.47\textwidth}@{\hspace{0.04\textwidth}}p{0.47\textwidth}@{}}
\toprule
\textbf{\textcolor{blue}{Prompt:}} Find me an everyday image that matches the given caption: A clean, spacious bathroom with a large shower stall.
\newline
\begin{tabular}{@{}cc@{}}
\textbf{\textcolor{blue}{GT}} & \textbf{\textcolor{blue}{Generated}} \\
\includegraphics[width=0.45\linewidth]{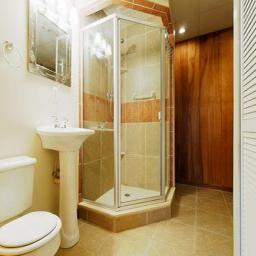} &
\includegraphics[width=0.45\linewidth]{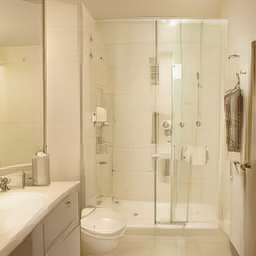}
\end{tabular}
&
\textbf{\textcolor{blue}{Prompt:}} Find me an everyday image that matches the given caption: A bathroom featuring a walk in shower, mirror, sink and toilet.
\newline
\begin{tabular}{@{}cc@{}}
\textbf{\textcolor{blue}{GT}} & \textbf{\textcolor{blue}{Generated}} \\
\includegraphics[width=0.45\linewidth]{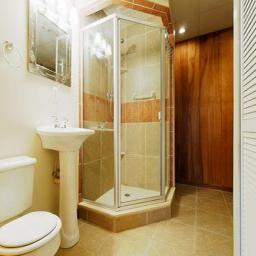} &
\includegraphics[width=0.45\linewidth]{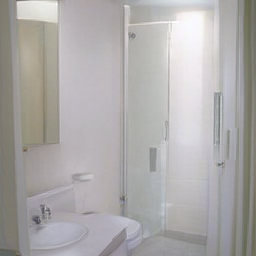}
\end{tabular}
\\
\bottomrule
\end{tabular}
\end{table}

\begin{table}[h!]
\centering\small
\begin{tabular}{@{}p{0.47\textwidth}@{\hspace{0.04\textwidth}}p{0.47\textwidth}@{}}
\toprule
\textbf{\textcolor{blue}{Prompt:}} Find me an everyday image that matches the given caption: Bathroom with a shower, sink, and toilet in it.
\newline
\begin{tabular}{@{}cc@{}}
\textbf{\textcolor{blue}{GT}} & \textbf{\textcolor{blue}{Generated}} \\
\includegraphics[width=0.45\linewidth]{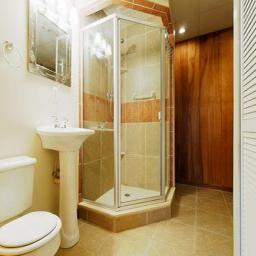} &
\includegraphics[width=0.45\linewidth]{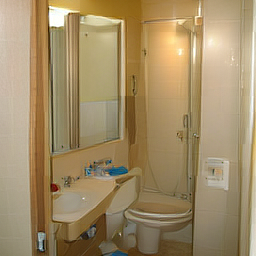}
\end{tabular}
&
\textbf{\textcolor{blue}{Prompt:}} Find me an everyday image that matches the given caption: People on bicycles ride down a busy street.
\newline
\begin{tabular}{@{}cc@{}}
\textbf{\textcolor{blue}{GT}} & \textbf{\textcolor{blue}{Generated}} \\
\includegraphics[width=0.45\linewidth]{images/exp021_vlm2vec_fullseq/MSCOCO_t2i/gt_0040.jpg} &
\includegraphics[width=0.45\linewidth]{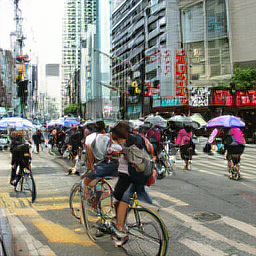}
\end{tabular}
\\
\bottomrule
\end{tabular}
\end{table}

\begin{table}[h!]
\centering\small
\begin{tabular}{@{}p{0.47\textwidth}@{\hspace{0.04\textwidth}}p{0.47\textwidth}@{}}
\toprule
\textbf{\textcolor{blue}{Prompt:}} Find me an everyday image that matches the given caption: A group of bicyclists are riding in the bike lane.
\newline
\begin{tabular}{@{}cc@{}}
\textbf{\textcolor{blue}{GT}} & \textbf{\textcolor{blue}{Generated}} \\
\includegraphics[width=0.45\linewidth]{images/exp021_vlm2vec_fullseq/MSCOCO_t2i/gt_0043.jpg} &
\includegraphics[width=0.45\linewidth]{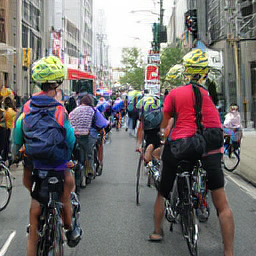}
\end{tabular}
&
\textbf{\textcolor{blue}{Prompt:}} Find me an everyday image that matches the given caption: Bicyclists on a city street, most not using the bike lane.
\newline
\begin{tabular}{@{}cc@{}}
\textbf{\textcolor{blue}{GT}} & \textbf{\textcolor{blue}{Generated}} \\
\includegraphics[width=0.45\linewidth]{images/exp021_vlm2vec_fullseq/MSCOCO_t2i/gt_0044.jpg} &
\includegraphics[width=0.45\linewidth]{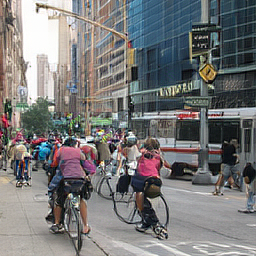}
\end{tabular}
\\
\bottomrule
\end{tabular}
\end{table}

\begin{table}[h!]
\centering\small
\begin{tabular}{@{}p{0.47\textwidth}@{\hspace{0.04\textwidth}}p{0.47\textwidth}@{}}
\toprule
\textbf{\textcolor{blue}{Prompt:}} Find me an everyday image that matches the given caption: The bathroom is clean and ready to be used.
\newline
\begin{tabular}{@{}cc@{}}
\textbf{\textcolor{blue}{GT}} & \textbf{\textcolor{blue}{Generated}} \\
\includegraphics[width=0.45\linewidth]{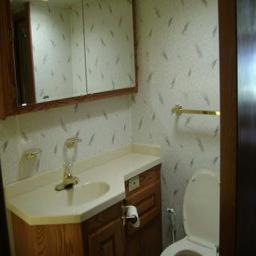} &
\includegraphics[width=0.45\linewidth]{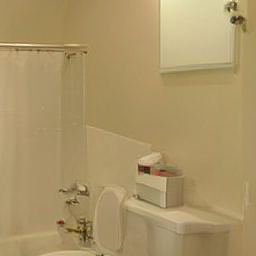}
\end{tabular}
&
\textbf{\textcolor{blue}{Prompt:}} Find me an everyday image that matches the given caption: A small bathroom with a sink and a toilet.
\newline
\begin{tabular}{@{}cc@{}}
\textbf{\textcolor{blue}{GT}} & \textbf{\textcolor{blue}{Generated}} \\
\includegraphics[width=0.45\linewidth]{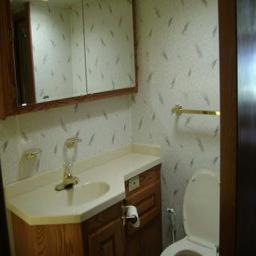} &
\includegraphics[width=0.45\linewidth]{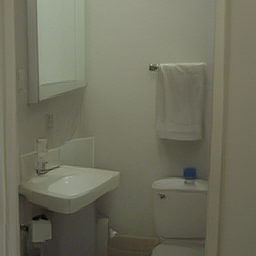}
\end{tabular}
\\
\bottomrule
\end{tabular}
\end{table}

\begin{table}[h!]
\centering\small
\begin{tabular}{@{}p{0.47\textwidth}@{\hspace{0.04\textwidth}}p{0.47\textwidth}@{}}
\toprule
\textbf{\textcolor{blue}{Prompt:}} Find me an everyday image that matches the given caption: A toilet, sink and mirror in the bathroom.
\newline
\begin{tabular}{@{}cc@{}}
\textbf{\textcolor{blue}{GT}} & \textbf{\textcolor{blue}{Generated}} \\
\includegraphics[width=0.45\linewidth]{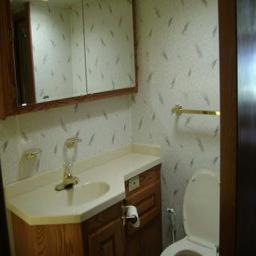} &
\includegraphics[width=0.45\linewidth]{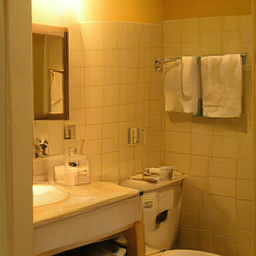}
\end{tabular}
&
\textbf{\textcolor{blue}{Prompt:}} Find me an everyday image that matches the given caption: Doorway view into bathroom with a sink and toilet.
\newline
\begin{tabular}{@{}cc@{}}
\textbf{\textcolor{blue}{GT}} & \textbf{\textcolor{blue}{Generated}} \\
\includegraphics[width=0.45\linewidth]{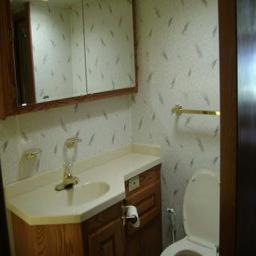} &
\includegraphics[width=0.45\linewidth]{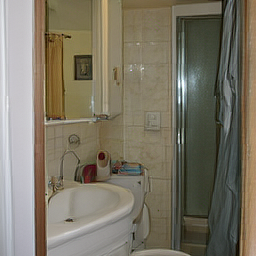}
\end{tabular}
\\
\bottomrule
\end{tabular}
\end{table}

\begin{table}[h!]
\centering\small
\begin{tabular}{@{}p{0.47\textwidth}@{\hspace{0.04\textwidth}}p{0.47\textwidth}@{}}
\toprule
\multicolumn{2}{@{}l@{}}{\textbf{Dataset: VisualNews\_t2i -- Successful Cases}} \\
\midrule
\textbf{\textcolor{blue}{Prompt:}} Retrieve an image of this news caption. A Maryland State Police cruiser sits at a blocked southbound entrance on the BaltimoreWashington Parkway that accesses Fort Meade.
\newline
\begin{tabular}{@{}cc@{}}
\textbf{\textcolor{blue}{GT}} & \textbf{\textcolor{blue}{Generated}} \\
\includegraphics[width=0.45\linewidth]{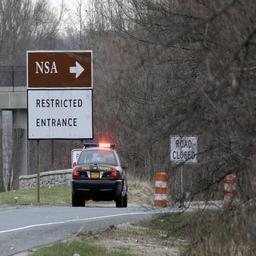} &
\includegraphics[width=0.45\linewidth]{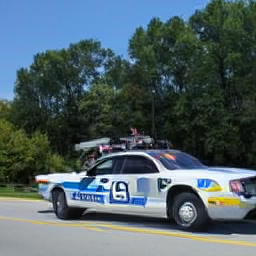}
\end{tabular}
&
\textbf{\textcolor{blue}{Prompt:}} Retrieve an image of this news caption. Ashton Eaton of the US competes in the decathlon 400m run.
\newline
\begin{tabular}{@{}cc@{}}
\textbf{\textcolor{blue}{GT}} & \textbf{\textcolor{blue}{Generated}} \\
\includegraphics[width=0.45\linewidth]{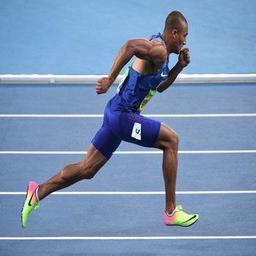} &
\includegraphics[width=0.45\linewidth]{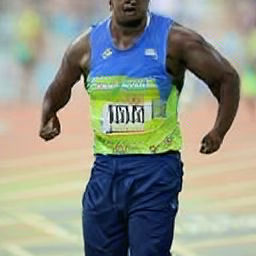}
\end{tabular}
\\
\bottomrule
\end{tabular}
\end{table}

\begin{table}[h!]
\centering\small
\begin{tabular}{@{}p{0.47\textwidth}@{\hspace{0.04\textwidth}}p{0.47\textwidth}@{}}
\toprule
\textbf{\textcolor{blue}{Prompt:}} Retrieve an image of this news caption. Marc Jacobs FallWinter 2016 collection.
\newline
\begin{tabular}{@{}cc@{}}
\textbf{\textcolor{blue}{GT}} & \textbf{\textcolor{blue}{Generated}} \\
\includegraphics[width=0.45\linewidth]{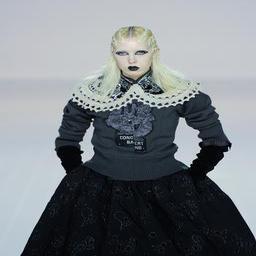} &
\includegraphics[width=0.45\linewidth]{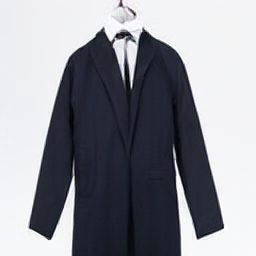}
\end{tabular}
&
\textbf{\textcolor{blue}{Prompt:}} Retrieve an image of this news caption. St Lawrence University on a snowy day.
\newline
\begin{tabular}{@{}cc@{}}
\textbf{\textcolor{blue}{GT}} & \textbf{\textcolor{blue}{Generated}} \\
\includegraphics[width=0.45\linewidth]{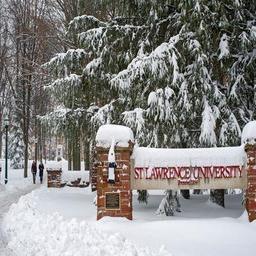} &
\includegraphics[width=0.45\linewidth]{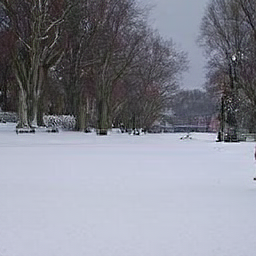}
\end{tabular}
\\
\bottomrule
\end{tabular}
\end{table}

\begin{table}[h!]
\centering\small
\begin{tabular}{@{}p{0.47\textwidth}@{\hspace{0.04\textwidth}}p{0.47\textwidth}@{}}
\toprule
\textbf{\textcolor{blue}{Prompt:}} Retrieve an image of this news caption. Malaysia Airlines Flight 17 in Amsterdam.
\newline
\begin{tabular}{@{}cc@{}}
\textbf{\textcolor{blue}{GT}} & \textbf{\textcolor{blue}{Generated}} \\
\includegraphics[width=0.45\linewidth]{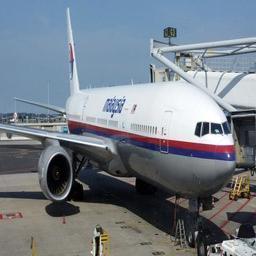} &
\includegraphics[width=0.45\linewidth]{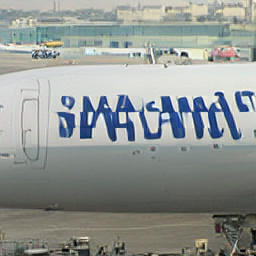}
\end{tabular}
&
\textbf{\textcolor{blue}{Prompt:}} Retrieve an image of this news caption. Obama and Kenyan President Uhuru Kenyatta take part in a roundtable in Nairobi with young businesspeople at the Global Entrepreneurship Summit at the United Nations compound.
\newline
\begin{tabular}{@{}cc@{}}
\textbf{\textcolor{blue}{GT}} & \textbf{\textcolor{blue}{Generated}} \\
\includegraphics[width=0.45\linewidth]{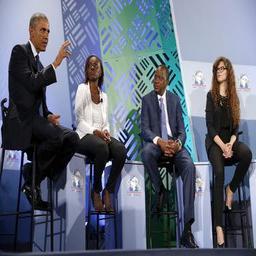} &
\includegraphics[width=0.45\linewidth]{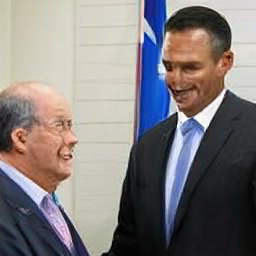}
\end{tabular}
\\
\bottomrule
\end{tabular}
\end{table}

\begin{table}[h!]
\centering\small
\begin{tabular}{@{}p{0.47\textwidth}@{\hspace{0.04\textwidth}}p{0.47\textwidth}@{}}
\toprule
\textbf{\textcolor{blue}{Prompt:}} Retrieve an image of this news caption. A vehicle burns in front of the home of a follower of Shiite cleric Mahmoud alSarkhi after clashes with his supporters in Karbala Iraq.
\newline
\begin{tabular}{@{}cc@{}}
\textbf{\textcolor{blue}{GT}} & \textbf{\textcolor{blue}{Generated}} \\
\includegraphics[width=0.45\linewidth]{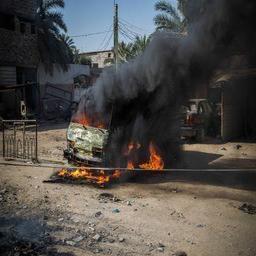} &
\includegraphics[width=0.45\linewidth]{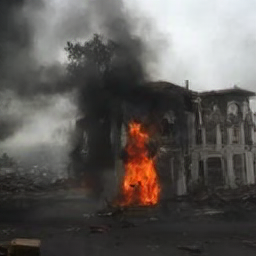}
\end{tabular}
\\
\bottomrule
\end{tabular}
\end{table}

\begin{table}[h!]
\centering\small
\begin{tabular}{@{}p{0.47\textwidth}@{\hspace{0.04\textwidth}}p{0.47\textwidth}@{}}
\toprule
\multicolumn{2}{@{}l@{}}{\textbf{Dataset: CIRR -- Successful Cases}} \\
\midrule
\raisebox{-\height}{\includegraphics[height=1.5cm]{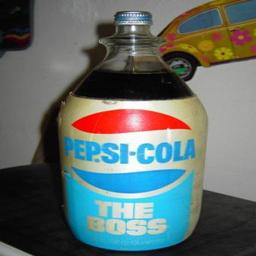}}\hspace{0.3em}\raisebox{-\baselineskip}{\parbox[t]{0.65\linewidth}{\textbf{\textcolor{blue}{Prompt:}} <|image\_1|>
Given an image, find a similar everyday image with the described changes: Show three bottles of soft drink.}}
\newline
\begin{tabular}{@{}cc@{}}
\textbf{\textcolor{blue}{GT}} & \textbf{\textcolor{blue}{Generated}} \\
\includegraphics[width=0.45\linewidth]{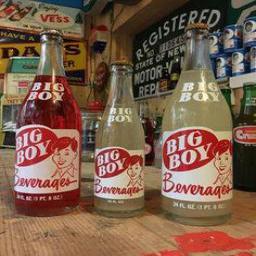} &
\includegraphics[width=0.45\linewidth]{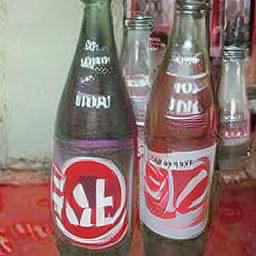}
\end{tabular}
&
\raisebox{-\height}{\includegraphics[height=1.5cm]{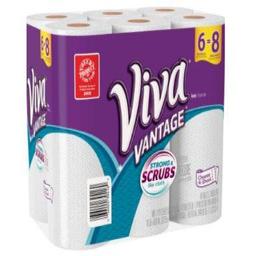}}\hspace{0.3em}\raisebox{-\baselineskip}{\parbox[t]{0.65\linewidth}{\textbf{\textcolor{blue}{Prompt:}} <|image\_1|>
Given an image, find a similar everyday image with the described changes: Fewer paper towels per pack.}}
\newline
\begin{tabular}{@{}cc@{}}
\textbf{\textcolor{blue}{GT}} & \textbf{\textcolor{blue}{Generated}} \\
\includegraphics[width=0.45\linewidth]{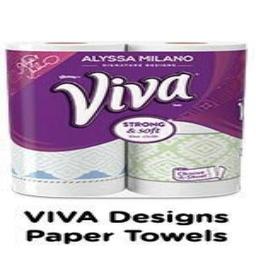} &
\includegraphics[width=0.45\linewidth]{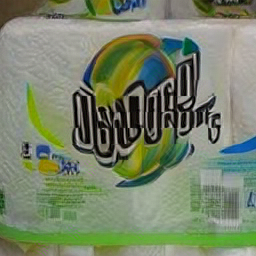}
\end{tabular}
\\
\bottomrule
\end{tabular}
\end{table}

\begin{table}[h!]
\centering\small
\begin{tabular}{@{}p{0.47\textwidth}@{\hspace{0.04\textwidth}}p{0.47\textwidth}@{}}
\toprule
\raisebox{-\height}{\includegraphics[height=1.5cm]{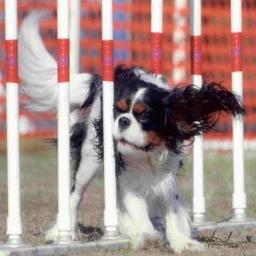}}\hspace{0.3em}\raisebox{-\baselineskip}{\parbox[t]{0.65\linewidth}{\textbf{\textcolor{blue}{Prompt:}} <|image\_1|>
Given an image, find a similar everyday image with the described changes: Same breed dog, focus on its head.}}
\newline
\begin{tabular}{@{}cc@{}}
\textbf{\textcolor{blue}{GT}} & \textbf{\textcolor{blue}{Generated}} \\
\includegraphics[width=0.45\linewidth]{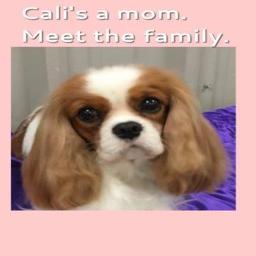} &
\includegraphics[width=0.45\linewidth]{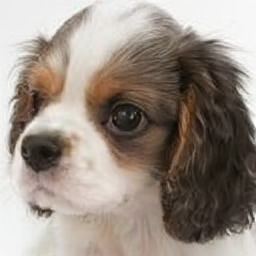}
\end{tabular}
&
\raisebox{-\height}{\includegraphics[height=1.5cm]{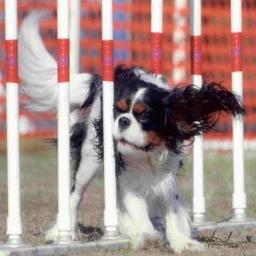}}\hspace{0.3em}\raisebox{-\baselineskip}{\parbox[t]{0.65\linewidth}{\textbf{\textcolor{blue}{Prompt:}} <|image\_1|>
Given an image, find a similar everyday image with the described changes: Two dogs of the same breed on the floor.}}
\newline
\begin{tabular}{@{}cc@{}}
\textbf{\textcolor{blue}{GT}} & \textbf{\textcolor{blue}{Generated}} \\
\includegraphics[width=0.45\linewidth]{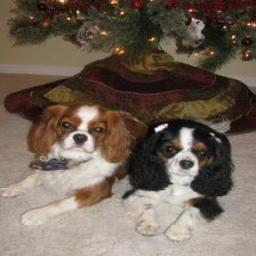} &
\includegraphics[width=0.45\linewidth]{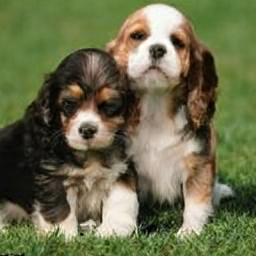}
\end{tabular}
\\
\bottomrule
\end{tabular}
\end{table}

\begin{table}[h!]
\centering\small
\begin{tabular}{@{}p{0.47\textwidth}@{\hspace{0.04\textwidth}}p{0.47\textwidth}@{}}
\toprule
\raisebox{-\height}{\includegraphics[height=1.5cm]{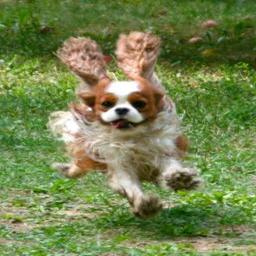}}\hspace{0.3em}\raisebox{-\baselineskip}{\parbox[t]{0.65\linewidth}{\textbf{\textcolor{blue}{Prompt:}} <|image\_1|>
Given an image, find a similar everyday image with the described changes: Be a same breed dog with his puppy running.}}
\newline
\begin{tabular}{@{}cc@{}}
\textbf{\textcolor{blue}{GT}} & \textbf{\textcolor{blue}{Generated}} \\
\includegraphics[width=0.45\linewidth]{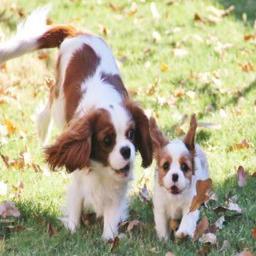} &
\includegraphics[width=0.45\linewidth]{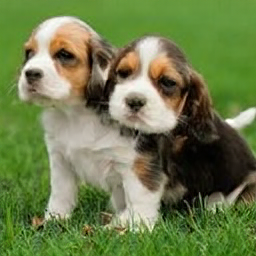}
\end{tabular}
&
\raisebox{-\height}{\includegraphics[height=1.5cm]{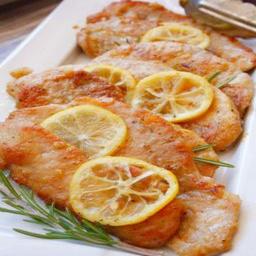}}\hspace{0.3em}\raisebox{-\baselineskip}{\parbox[t]{0.65\linewidth}{\textbf{\textcolor{blue}{Prompt:}} <|image\_1|>
Given an image, find a similar everyday image with the described changes: Remove the lemon.}}
\newline
\begin{tabular}{@{}cc@{}}
\textbf{\textcolor{blue}{GT}} & \textbf{\textcolor{blue}{Generated}} \\
\includegraphics[width=0.45\linewidth]{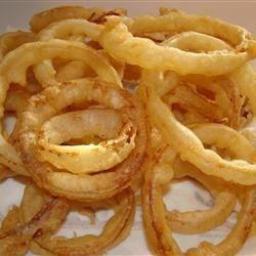} &
\includegraphics[width=0.45\linewidth]{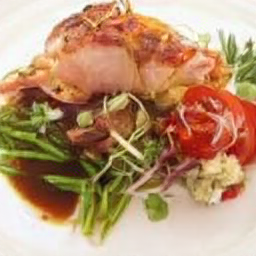}
\end{tabular}
\\
\bottomrule
\end{tabular}
\end{table}

\begin{table}[h!]
\centering\small
\begin{tabular}{@{}p{0.47\textwidth}@{\hspace{0.04\textwidth}}p{0.47\textwidth}@{}}
\toprule
\raisebox{-\height}{\includegraphics[height=1.5cm]{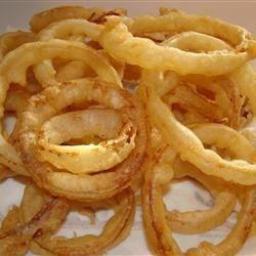}}\hspace{0.3em}\raisebox{-\baselineskip}{\parbox[t]{0.65\linewidth}{\textbf{\textcolor{blue}{Prompt:}} <|image\_1|>
Given an image, find a similar everyday image with the described changes: Put the fries in a white plate with white background, clean.}}
\newline
\begin{tabular}{@{}cc@{}}
\textbf{\textcolor{blue}{GT}} & \textbf{\textcolor{blue}{Generated}} \\
\includegraphics[width=0.45\linewidth]{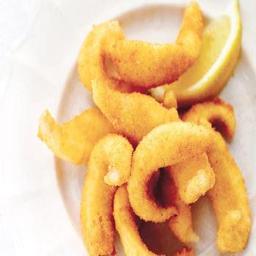} &
\includegraphics[width=0.45\linewidth]{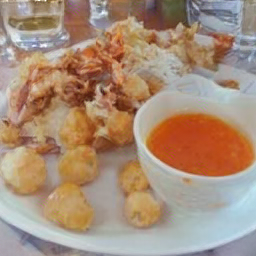}
\end{tabular}
&
\raisebox{-\height}{\includegraphics[height=1.5cm]{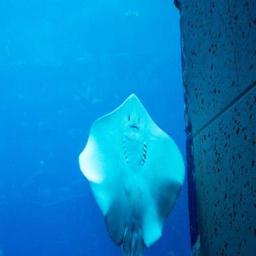}}\hspace{0.3em}\raisebox{-\baselineskip}{\parbox[t]{0.65\linewidth}{\textbf{\textcolor{blue}{Prompt:}} <|image\_1|>
Given an image, find a similar everyday image with the described changes: Remove the concret to the right.}}
\newline
\begin{tabular}{@{}cc@{}}
\textbf{\textcolor{blue}{GT}} & \textbf{\textcolor{blue}{Generated}} \\
\includegraphics[width=0.45\linewidth]{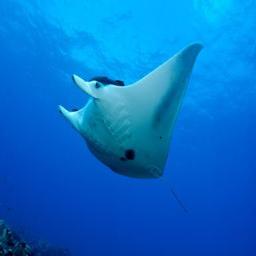} &
\includegraphics[width=0.45\linewidth]{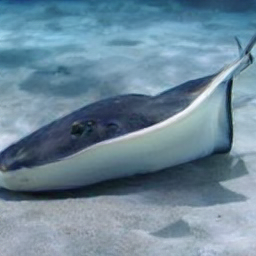}
\end{tabular}
\\
\bottomrule
\end{tabular}
\end{table}

\begin{table}[h!]
\centering\small
\begin{tabular}{@{}p{0.47\textwidth}@{\hspace{0.04\textwidth}}p{0.47\textwidth}@{}}
\toprule
\raisebox{-\height}{\includegraphics[height=1.5cm]{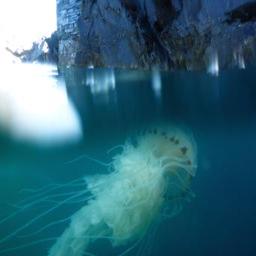}}\hspace{0.3em}\raisebox{-\baselineskip}{\parbox[t]{0.65\linewidth}{\textbf{\textcolor{blue}{Prompt:}} <|image\_1|>
Given an image, find a similar everyday image with the described changes: Shot from an angle so that the surface is visible.}}
\newline
\begin{tabular}{@{}cc@{}}
\textbf{\textcolor{blue}{GT}} & \textbf{\textcolor{blue}{Generated}} \\
\includegraphics[width=0.45\linewidth]{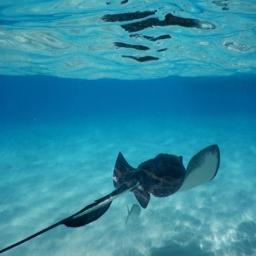} &
\includegraphics[width=0.45\linewidth]{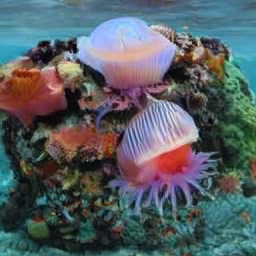}
\end{tabular}
&
\raisebox{-\height}{\includegraphics[height=1.5cm]{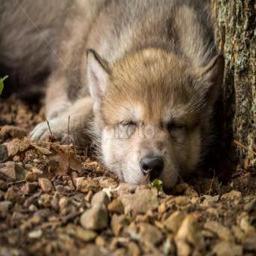}}\hspace{0.3em}\raisebox{-\baselineskip}{\parbox[t]{0.65\linewidth}{\textbf{\textcolor{blue}{Prompt:}} <|image\_1|>
Given an image, find a similar everyday image with the described changes: More focused on its head.}}
\newline
\begin{tabular}{@{}cc@{}}
\textbf{\textcolor{blue}{GT}} & \textbf{\textcolor{blue}{Generated}} \\
\includegraphics[width=0.45\linewidth]{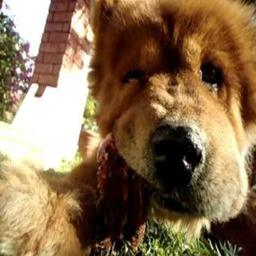} &
\includegraphics[width=0.45\linewidth]{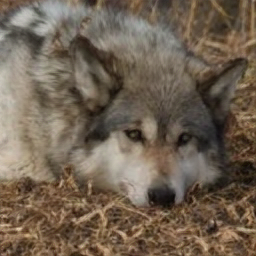}
\end{tabular}
\\
\bottomrule
\end{tabular}
\end{table}

\begin{table}[h!]
\centering\small
\begin{tabular}{@{}p{0.47\textwidth}@{\hspace{0.04\textwidth}}p{0.47\textwidth}@{}}
\toprule
\raisebox{-\height}{\includegraphics[height=1.5cm]{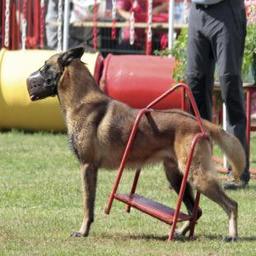}}\hspace{0.3em}\raisebox{-\baselineskip}{\parbox[t]{0.65\linewidth}{\textbf{\textcolor{blue}{Prompt:}} <|image\_1|>
Given an image, find a similar everyday image with the described changes: Dog is barking, nothing blocking his mouth.}}
\newline
\begin{tabular}{@{}cc@{}}
\textbf{\textcolor{blue}{GT}} & \textbf{\textcolor{blue}{Generated}} \\
\includegraphics[width=0.45\linewidth]{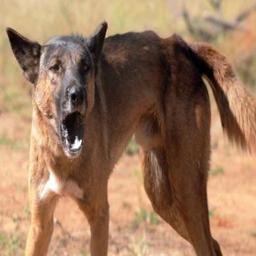} &
\includegraphics[width=0.45\linewidth]{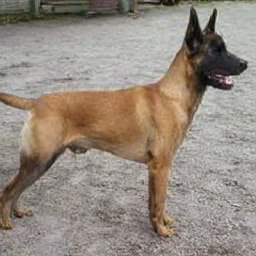}
\end{tabular}
&
\raisebox{-\height}{\includegraphics[height=1.5cm]{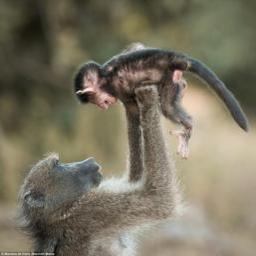}}\hspace{0.3em}\raisebox{-\baselineskip}{\parbox[t]{0.65\linewidth}{\textbf{\textcolor{blue}{Prompt:}} <|image\_1|>
Given an image, find a similar everyday image with the described changes: Add a wooden log.}}
\newline
\begin{tabular}{@{}cc@{}}
\textbf{\textcolor{blue}{GT}} & \textbf{\textcolor{blue}{Generated}} \\
\includegraphics[width=0.45\linewidth]{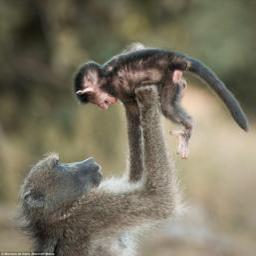} &
\includegraphics[width=0.45\linewidth]{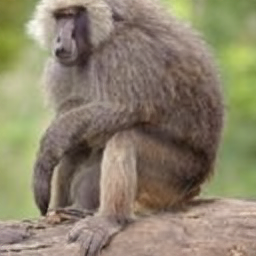}
\end{tabular}
\\
\bottomrule
\end{tabular}
\end{table}

\begin{table}[h!]
\centering\small
\begin{tabular}{@{}p{0.47\textwidth}@{\hspace{0.04\textwidth}}p{0.47\textwidth}@{}}
\toprule
\raisebox{-\height}{\includegraphics[height=1.5cm]{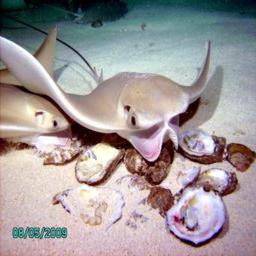}}\hspace{0.3em}\raisebox{-\baselineskip}{\parbox[t]{0.65\linewidth}{\textbf{\textcolor{blue}{Prompt:}} <|image\_1|>
Given an image, find a similar everyday image with the described changes: Remove the seashelles and make the water green.}}
\newline
\begin{tabular}{@{}cc@{}}
\textbf{\textcolor{blue}{GT}} & \textbf{\textcolor{blue}{Generated}} \\
\includegraphics[width=0.45\linewidth]{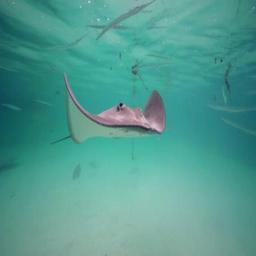} &
\includegraphics[width=0.45\linewidth]{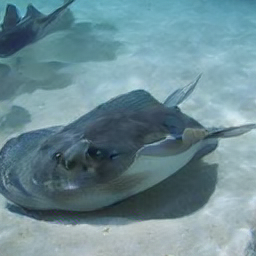}
\end{tabular}
&
\raisebox{-\height}{\includegraphics[height=1.5cm]{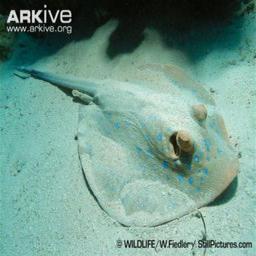}}\hspace{0.3em}\raisebox{-\baselineskip}{\parbox[t]{0.65\linewidth}{\textbf{\textcolor{blue}{Prompt:}} <|image\_1|>
Given an image, find a similar everyday image with the described changes: Change to blue water.}}
\newline
\begin{tabular}{@{}cc@{}}
\textbf{\textcolor{blue}{GT}} & \textbf{\textcolor{blue}{Generated}} \\
\includegraphics[width=0.45\linewidth]{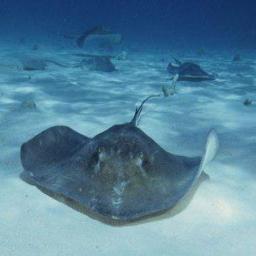} &
\includegraphics[width=0.45\linewidth]{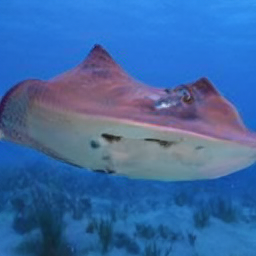}
\end{tabular}
\\
\bottomrule
\end{tabular}
\end{table}

\begin{table}[h!]
\centering\small
\begin{tabular}{@{}p{0.47\textwidth}@{\hspace{0.04\textwidth}}p{0.47\textwidth}@{}}
\toprule
\raisebox{-\height}{\includegraphics[height=1.5cm]{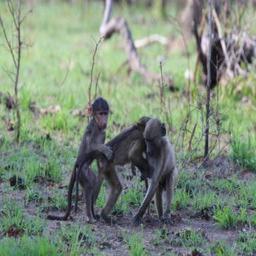}}\hspace{0.3em}\raisebox{-\baselineskip}{\parbox[t]{0.65\linewidth}{\textbf{\textcolor{blue}{Prompt:}} <|image\_1|>
Given an image, find a similar everyday image with the described changes: Brighter sunshine.}}
\newline
\begin{tabular}{@{}cc@{}}
\textbf{\textcolor{blue}{GT}} & \textbf{\textcolor{blue}{Generated}} \\
\includegraphics[width=0.45\linewidth]{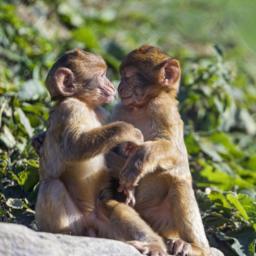} &
\includegraphics[width=0.45\linewidth]{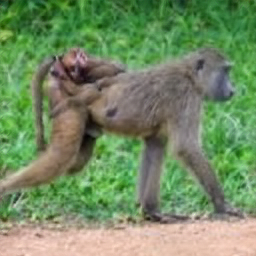}
\end{tabular}
&
\raisebox{-\height}{\includegraphics[height=1.5cm]{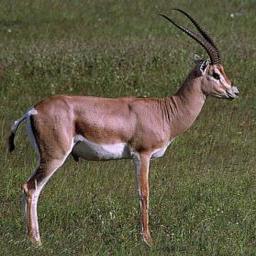}}\hspace{0.3em}\raisebox{-\baselineskip}{\parbox[t]{0.65\linewidth}{\textbf{\textcolor{blue}{Prompt:}} <|image\_1|>
Given an image, find a similar everyday image with the described changes: Add one more deer and add some sunlight.}}
\newline
\begin{tabular}{@{}cc@{}}
\textbf{\textcolor{blue}{GT}} & \textbf{\textcolor{blue}{Generated}} \\
\includegraphics[width=0.45\linewidth]{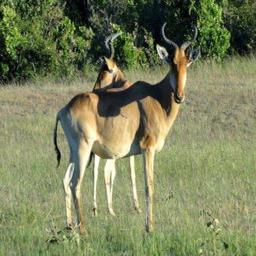} &
\includegraphics[width=0.45\linewidth]{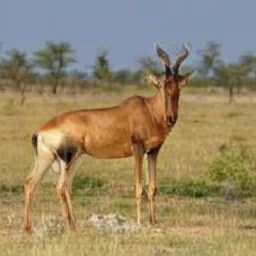}
\end{tabular}
\\
\bottomrule
\end{tabular}
\end{table}

\begin{table}[h!]
\centering\small
\begin{tabular}{@{}p{0.47\textwidth}@{\hspace{0.04\textwidth}}p{0.47\textwidth}@{}}
\toprule
\raisebox{-\height}{\includegraphics[height=1.5cm]{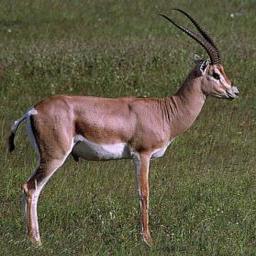}}\hspace{0.3em}\raisebox{-\baselineskip}{\parbox[t]{0.65\linewidth}{\textbf{\textcolor{blue}{Prompt:}} <|image\_1|>
Given an image, find a similar everyday image with the described changes: Show the back of the deer, and he is looking to the camera.}}
\newline
\begin{tabular}{@{}cc@{}}
\textbf{\textcolor{blue}{GT}} & \textbf{\textcolor{blue}{Generated}} \\
\includegraphics[width=0.45\linewidth]{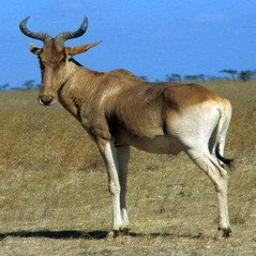} &
\includegraphics[width=0.45\linewidth]{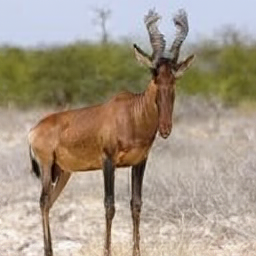}
\end{tabular}
&
\raisebox{-\height}{\includegraphics[height=1.5cm]{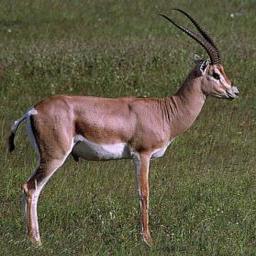}}\hspace{0.3em}\raisebox{-\baselineskip}{\parbox[t]{0.65\linewidth}{\textbf{\textcolor{blue}{Prompt:}} <|image\_1|>
Given an image, find a similar everyday image with the described changes: The deer looks to the camera, the back is a forest.}}
\newline
\begin{tabular}{@{}cc@{}}
\textbf{\textcolor{blue}{GT}} & \textbf{\textcolor{blue}{Generated}} \\
\includegraphics[width=0.45\linewidth]{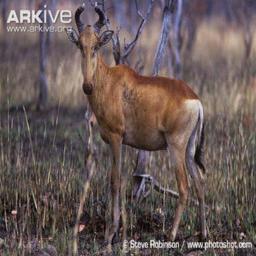} &
\includegraphics[width=0.45\linewidth]{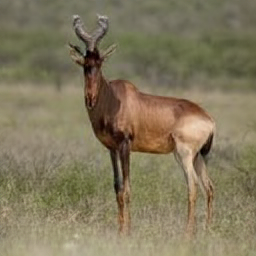}
\end{tabular}
\\
\bottomrule
\end{tabular}
\end{table}

\begin{table}[h!]
\centering\small
\begin{tabular}{@{}p{0.47\textwidth}@{\hspace{0.04\textwidth}}p{0.47\textwidth}@{}}
\toprule
\raisebox{-\height}{\includegraphics[height=1.5cm]{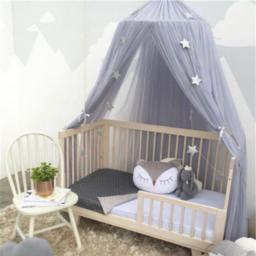}}\hspace{0.3em}\raisebox{-\baselineskip}{\parbox[t]{0.65\linewidth}{\textbf{\textcolor{blue}{Prompt:}} <|image\_1|>
Given an image, find a similar everyday image with the described changes: Show a living room with white tent.}}
\newline
\begin{tabular}{@{}cc@{}}
\textbf{\textcolor{blue}{GT}} & \textbf{\textcolor{blue}{Generated}} \\
\includegraphics[width=0.45\linewidth]{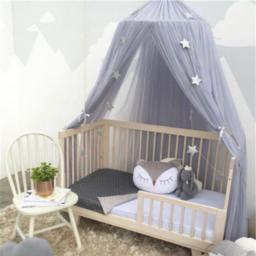} &
\includegraphics[width=0.45\linewidth]{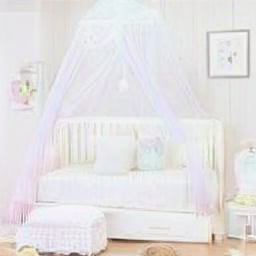}
\end{tabular}
&
\raisebox{-\height}{\includegraphics[height=1.5cm]{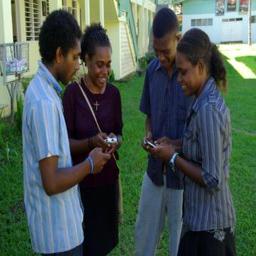}}\hspace{0.3em}\raisebox{-\baselineskip}{\parbox[t]{0.65\linewidth}{\textbf{\textcolor{blue}{Prompt:}} <|image\_1|>
Given an image, find a similar everyday image with the described changes: Better lighting, make them look at phones.}}
\newline
\begin{tabular}{@{}cc@{}}
\textbf{\textcolor{blue}{GT}} & \textbf{\textcolor{blue}{Generated}} \\
\includegraphics[width=0.45\linewidth]{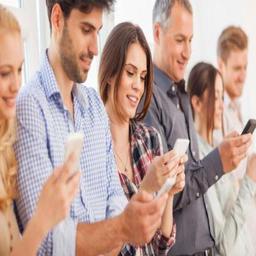} &
\includegraphics[width=0.45\linewidth]{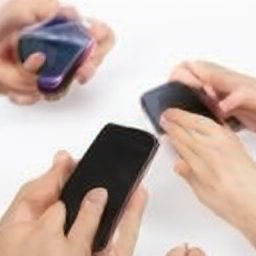}
\end{tabular}
\\
\bottomrule
\end{tabular}
\end{table}

\begin{table}[h!]
\centering\small
\begin{tabular}{@{}p{0.47\textwidth}@{\hspace{0.04\textwidth}}p{0.47\textwidth}@{}}
\toprule
\raisebox{-\height}{\includegraphics[height=1.5cm]{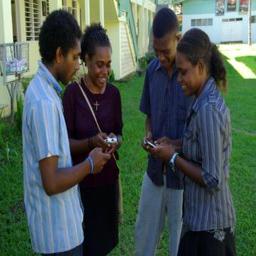}}\hspace{0.3em}\raisebox{-\baselineskip}{\parbox[t]{0.65\linewidth}{\textbf{\textcolor{blue}{Prompt:}} <|image\_1|>
Given an image, find a similar everyday image with the described changes: Inside, with food.}}
\newline
\begin{tabular}{@{}cc@{}}
\textbf{\textcolor{blue}{GT}} & \textbf{\textcolor{blue}{Generated}} \\
\includegraphics[width=0.45\linewidth]{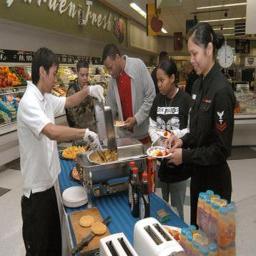} &
\includegraphics[width=0.45\linewidth]{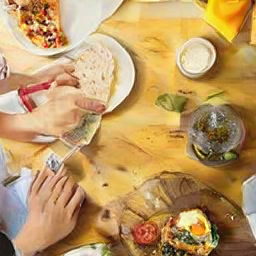}
\end{tabular}
&
\raisebox{-\height}{\includegraphics[height=1.5cm]{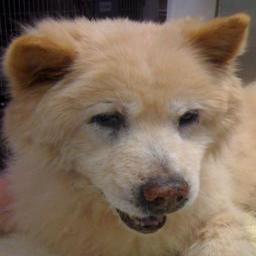}}\hspace{0.3em}\raisebox{-\baselineskip}{\parbox[t]{0.65\linewidth}{\textbf{\textcolor{blue}{Prompt:}} <|image\_1|>
Given an image, find a similar everyday image with the described changes: Show the whole dog standing on a dirt floor.}}
\newline
\begin{tabular}{@{}cc@{}}
\textbf{\textcolor{blue}{GT}} & \textbf{\textcolor{blue}{Generated}} \\
\includegraphics[width=0.45\linewidth]{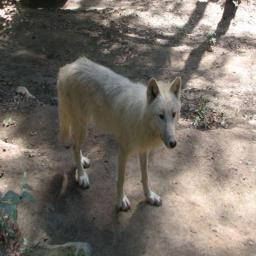} &
\includegraphics[width=0.45\linewidth]{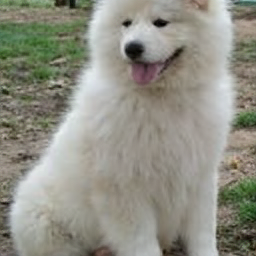}
\end{tabular}
\\
\bottomrule
\end{tabular}
\end{table}

\begin{table}[h!]
\centering\small
\begin{tabular}{@{}p{0.47\textwidth}@{\hspace{0.04\textwidth}}p{0.47\textwidth}@{}}
\toprule
\multicolumn{2}{@{}l@{}}{\textbf{Dataset: CIRR -- Failure Cases}} \\
\midrule
\raisebox{-\height}{\includegraphics[height=1.5cm]{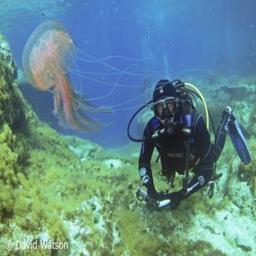}}\hspace{0.3em}\raisebox{-\baselineskip}{\parbox[t]{0.65\linewidth}{\textbf{\textcolor{blue}{Prompt:}} <|image\_1|>
Given an image, find a similar everyday image with the described changes: Two animals that are of a different species from the first one, and no human appearing.}}
\newline
\begin{tabular}{@{}cc@{}}
\textbf{\textcolor{blue}{GT}} & \textbf{\textcolor{blue}{Generated}} \\
\includegraphics[width=0.45\linewidth]{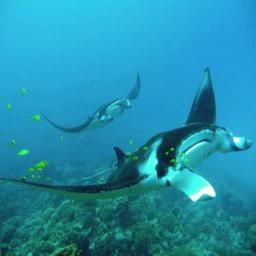} &
\includegraphics[width=0.45\linewidth]{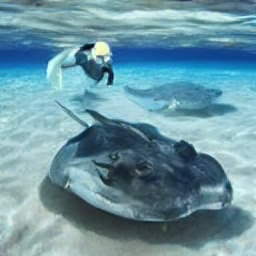}
\end{tabular}
&
\raisebox{-\height}{\includegraphics[height=1.5cm]{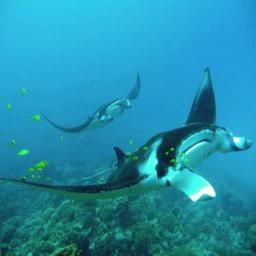}}\hspace{0.3em}\raisebox{-\baselineskip}{\parbox[t]{0.65\linewidth}{\textbf{\textcolor{blue}{Prompt:}} <|image\_1|>
Given an image, find a similar everyday image with the described changes: Human and one animal from a different species from the last two.}}
\newline
\begin{tabular}{@{}cc@{}}
\textbf{\textcolor{blue}{GT}} & \textbf{\textcolor{blue}{Generated}} \\
\includegraphics[width=0.45\linewidth]{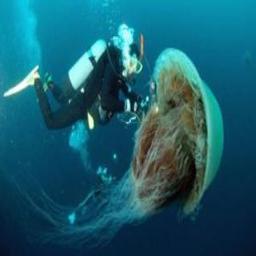} &
\includegraphics[width=0.45\linewidth]{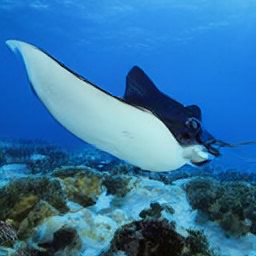}
\end{tabular}
\\
\bottomrule
\end{tabular}
\end{table}

\begin{table}[h!]
\centering\small
\begin{tabular}{@{}p{0.47\textwidth}@{\hspace{0.04\textwidth}}p{0.47\textwidth}@{}}
\toprule
\raisebox{-\height}{\includegraphics[height=1.5cm]{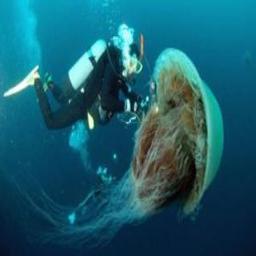}}\hspace{0.3em}\raisebox{-\baselineskip}{\parbox[t]{0.65\linewidth}{\textbf{\textcolor{blue}{Prompt:}} <|image\_1|>
Given an image, find a similar everyday image with the described changes: Mae the human facing the camera with less gear.}}
\newline
\begin{tabular}{@{}cc@{}}
\textbf{\textcolor{blue}{GT}} & \textbf{\textcolor{blue}{Generated}} \\
\includegraphics[width=0.45\linewidth]{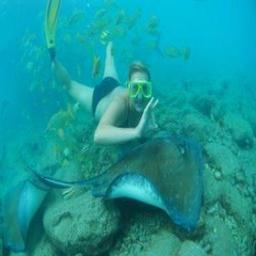} &
\includegraphics[width=0.45\linewidth]{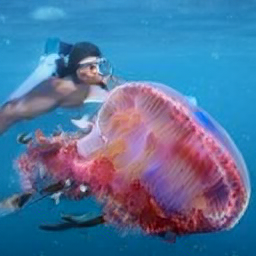}
\end{tabular}
&
\raisebox{-\height}{\includegraphics[height=1.5cm]{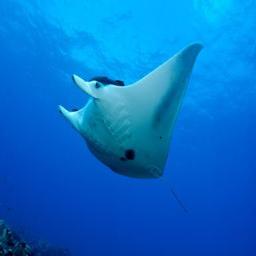}}\hspace{0.3em}\raisebox{-\baselineskip}{\parbox[t]{0.65\linewidth}{\textbf{\textcolor{blue}{Prompt:}} <|image\_1|>
Given an image, find a similar everyday image with the described changes: Change the anmial type but preserve the environment.}}
\newline
\begin{tabular}{@{}cc@{}}
\textbf{\textcolor{blue}{GT}} & \textbf{\textcolor{blue}{Generated}} \\
\includegraphics[width=0.45\linewidth]{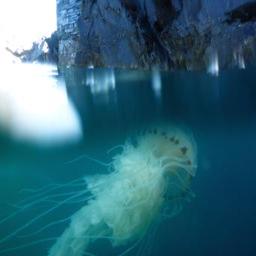} &
\includegraphics[width=0.45\linewidth]{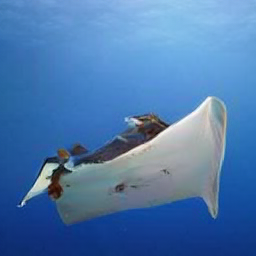}
\end{tabular}
\\
\bottomrule
\end{tabular}
\end{table}

\begin{table}[h!]
\centering\small
\begin{tabular}{@{}p{0.47\textwidth}@{\hspace{0.04\textwidth}}p{0.47\textwidth}@{}}
\toprule
\raisebox{-\height}{\includegraphics[height=1.5cm]{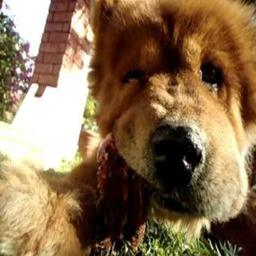}}\hspace{0.3em}\raisebox{-\baselineskip}{\parbox[t]{0.65\linewidth}{\textbf{\textcolor{blue}{Prompt:}} <|image\_1|>
Given an image, find a similar everyday image with the described changes: Show the animal standing on dirt.}}
\newline
\begin{tabular}{@{}cc@{}}
\textbf{\textcolor{blue}{GT}} & \textbf{\textcolor{blue}{Generated}} \\
\includegraphics[width=0.45\linewidth]{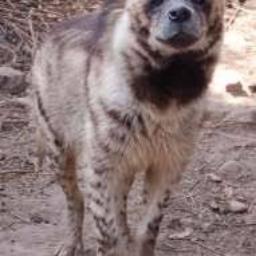} &
\includegraphics[width=0.45\linewidth]{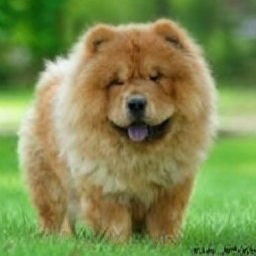}
\end{tabular}
\\
\bottomrule
\end{tabular}
\end{table}

\end{document}